# Historical patterns of rice farming explain modern-day language use in China and Japan more than modernization and urbanization

Sharath Chandra Guntuku[1,2]*, Thomas Talhelm[3]*, Garrick Sherman[1,2], Angel Fan[2], Salvatore Giorgi[1], Liuqing Wei[4], Lyle H. Ungar[1,2]

[1]Department of Computer and Information Science, University of Pennsylvania, Philadelphia, PA, USA
[2]Positive Psychology Center, University of Pennsylvania, Philadelphia, PA, USA
[3]Booth School of Business, University of Chicago, Chicago, IL, USA
[4]Department of Psychology, Hubei University, Wuhan, Hubei, China

*Corresponding authors: sharathg@cis.upenn.edu, thomas.talhelm@chicagobooth.edu

**Abstract**
We used natural language processing to analyze a billion words to study cultural differences on Weibo, one of China's largest social media platforms. We compared predictions from two common explanations about cultural differences in China (economic development and urban-rural differences) against the less-obvious legacy of rice versus wheat farming. Rice farmers had to coordinate shared irrigation networks and exchange labor to cope with higher labor requirements. In contrast, wheat relied on rainfall and required half as much labor. We test whether this legacy made southern China more interdependent. Across all word categories, rice explained twice as much variance as economic development and urbanization. Rice areas used more words reflecting tight social ties, holistic thought, and a cautious, prevention orientation. We then used Twitter data comparing prefectures in Japan, which largely replicated the results from China. This provides crucial evidence of the rice theory in a different nation, language, and platform.

**Introduction**
    Social psychologists have discovered that the words people use can give insight into their thought and behavior[1]. For example, people's word use reflects their personalities[2]. Word use can also predict future behavior. Studies have found that depressed college students and poets who later went on to commit suicide used more self-focused language than non-depressed people[3,4].
    Beyond differences between individuals, researchers have also used language to explore differences between regions[5]. For example, researchers analyzed language use on Twitter and found that people in areas that expressed more negative emotions—particularly anger—had higher rates of heart attacks[6]. In another study, sentiment toward the Affordable Care Act ("Obamacare") on Twitter predicted differences in enrollment across states[7].
    In sum, these studies suggest that language use can give insight into people's psychology and regional differences. In this study, we analyze over a billion words from Weibo (which is similar to Twitter) to gain insight into regional differences across China. To frame our search, we test categories and constructs that cultural psychology has linked to individualism and collectivism. We also use bottom-up machine learning to discover word-use differences that might not map onto predicted differences. This could allow us to discover new, unanticipated differences. For each category, we test the societal factors that psychologists have argued are causes of collectivism.

**Modernization**

One theory we test is modernization theory. Modernization theory is perhaps the most widely researched theory of culture[8,9]. It is the idea that, as cultures become more wealthy, modernized, or urbanized, they become more individualistic[10]. This narrative is particularly strong among papers researching differences in China[11]. Researchers have argued that modernization has made people in China more narcissistic, more individualistic, and more self-indulgent[11–13].

The attention on modernization in China makes sense, as China is at a unique place in history to test modernization theory due to its rapid economic growth in the last two decades. Economic development varies dramatically around China. GDP per capita goes from US$3,859 in northwest Gansu province to US$14,600 in Beijing (based on 2014 GDP per capita, converted to US dollars). That's roughly the difference between the Republic of Congo and Argentina[14]. The broad pattern of development is low in the west and interior of China and high along the eastern coast (Figure 1).

**The Urban-Rural Divide**

Closely linked to modernization is the urban-rural divide. Cultural psychologists have found some evidence that people in cities are more individualistic than people in rural areas[15]. In China, a lot of the discussion of the urban-rural divide focuses on the wealth gap[16]. If urban-rural differences are mostly wealth differences, urbanization predictions would be mostly redundant with the economic development predictions.

However, urbanization isn't exactly the same as wealth. Some smaller towns and rural areas are quite wealthy. Plus, cities have distinctive features beyond wealth. Cities are hubs of diversity, museums, art, and universities. Thus, we test for urbanization separately from modernization. In China, urbanization mostly falls along the eastern coast, but there are important interior large cities, such as Chongqing (Figure 1).

**Figure 1:** *The Geographic Distribution of Rice (Upper Left), Modernization (Upper Right), and Urbanization (Lower Left)*

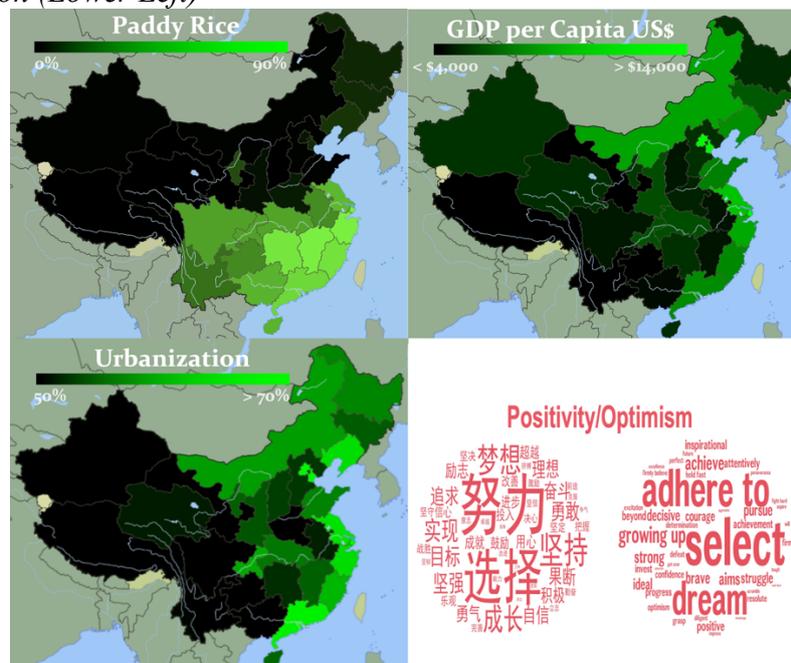

Note: The lower right displays the most common words for the positivity/optimism category. In the word cloud, larger words appear more frequently in the Weibo data. Urbanization is the percentage of urban residents in 2016. GDP per capita is from 2014. Rice data is the

earliest available (from the 1996 *Statistical Yearbook*), although this data correlates strongly with limited data available from 100 years ago.

**The Rice Theory**

Besides modernization, we test whether China's history of farming rice has left a lasting influence on the culture. For generations, people around the Yangtze River and further south have farmed paddy rice. Farther north from the Yangtze River, people have farmed wheat, millet, and other dryland crops.

Why would rice and wheat be important for culture? Paddy rice is unlike any other major grain[17]. For one, rice grows best in standing water. If farmers can flood their fields, they can reap 4-5 more tons per hectare than dryland rice[18]. That encourages rice farmers to build irrigation systems to control water levels.

However, those irrigation systems create classic commons dilemmas. All of the farmers can produce more rice with irrigation systems, but no single farmer wants to be responsible for building, dredging, and repairing the irrigation networks. In response, traditional rice villages in China created rotating task schedules and enforced punishments for people who did not show up[19].

Once farmers controlled the water, it meant they now had to coordinate their water use. When water was scarce, farmers had to coordinate which fields get flooded and which did not. In some irrigation networks, farmers had to flood and drain their fields at the same time[20]. That made it difficult to be a rogue rice farmer.

Paddy rice also comes with a huge labor burden. Anthropologists observing pre-modern rice farmers found that rice required about twice the number of hours per hectare as crops like wheat or barley[21]. This was true even when the same farmer planted a plot with rice one year and other crops the year after[21].

Part of the labor burden comes from managing irrigation, but another part also comes from the process of transplanting rice from seedbeds into the main field (which is not done with wheat). Wet, muddy fields also made work more difficult[22]. The labor burden is important for culture because it led rice farmers from Japan to West Africa to form cooperative labor exchanges[20,21,23]. In sum, rice farming was more interdependent, with tighter social ties than wheat farming.

In line with this theory, people in southern China score higher on measures of implicit interdependence and loyalty/nepotism[24]. In contrast, people in the wheat-farming north of China are more likely to spend time alone[25]. Around the world, countries with a history of rice farming tend to have tighter social norms[26] and smaller, more binding social ties[27].

**Opposite Predictions**

Importantly, the history of rice in China leads to opposite predictions from modernization. Rice happens to be distributed conveniently for researchers interested in the causes of culture (Figure 1). In the 1980s, Deng Xiaoping started the Reform and Opening policy, creating special economic zones. Perhaps to insulate the central government from these risky reforms, Deng put these zones in southern China. With the runaway success of foreign trade, the rice area of China is now wealthier on average than the wheat area[24].

That accident of history puts the modernization theory and the rice theory in direct contrast. If modernization is a strong force on culture, the rice areas of China should be more individualistic than the wheat areas. But if China's agricultural legacy continues to influence culture, we should see more markers of interdependent culture in southern China.

That said, rice and wealth are not so highly correlated as to be confounded. The rice region includes wealthy areas and some of China's poorest provinces. For example, Shanghai and Zhejiang have more than double the GDP per capita as provinces like Guangxi and Jiangxi. China's wheat areas also include wealthy areas like Beijing and poorer provinces like Henan and Shanxi.

**Other Theories of Culture**

Of course, modernization and subsistence style are not the only influences on culture. We also test a thorough set of other theories on the causes of culture based on disease[28], climate[29], herding[30], education[9], and ethnic diversity[31]. Table S1 lays out all data sources, measures, and theoretical rationales for regional differences.

**Linguistic Categories**

To categorize words into psychological constructs, we started with the Linguistic Inquiry and Word Count tool (LIWC). We used the 2015 version of the Chinese dictionary, except for the "humans" category, which is only available in the 2007 version. LIWC has psychometrically validated categories, such as positive emotion, cognitive processes, and achievement words. Researchers have used the LIWC dictionary in many studies, analyzing everything from blogs to poems[3]. We used the simplified Chinese version of LIWC, which predicts personality traits, depression, suicidal tendencies, and mental health in individuals and communities[32].

**Theory-Driven Categories.** We created five new categories based on theories of collectivism in cultural psychology that were not represented in the LIWC categories: in-group/out-group, universalism, positivity-optimism, and fashion/trends. We provide details on the theoretical founding of these predictions in Supplemental Section 6. In the main text, we focus results on categories that revealed meaningful regional differences, but we report all results in the supplemental materials. The supplemental materials include a full list of the words in the newly created word categories. Next, we outline our main word categories of interest.

**1. Cognition and Discourse**

**Cognitive Process.** First, we analyzed differences related to thought style. Decades of research have documented differences in thought styles between East and West[33]. For example, people in North America and Western Europe are more likely than people in China and Japan to rely on rules of formal logic, such as logical non-contradiction[34]. Participants from East Asia are more likely to think dialectically, which accepts the possibility that an idea and a contradictory idea can both be true[35]. Researchers have used the term "holistic" to describe the dominant thought style in East Asia and "analytic" to describe the thought style of the West[33].

Cultural psychologists have theorized that differences in social style cause these differences in thought style[36]. This theory is based on several observations:

  1. Thought style and social style are correlated across nations. Interdependent cultures tend to think more holistically than individualistic cultures[37].
  2. Within national cultures, more interdependent groups tend to think more holistically. For example, women and people from working-class backgrounds tend to think more holistically than men and people from white-collar backgrounds[38,39].
  3. Researchers have experimentally put people in an interdependent mindset using tasks like reading stories about characters who take other people into account or who act independently from others[40]. A meta-analysis of different priming techniques found that people tend to think more holistically after interdependent priming[41].

Based on this data, the rice theory and modernization theory make two different predictions:

**Rice:** If rice cultures are more interdependent, then the rice areas of southern China should think more holistically.

**Modernization:** If modernization makes cultures more individualistic, the wealthier southern provinces should think *less* holistically.

We tested this idea using the "cognitive processes" category of LIWC. Cognitive process words are related to thought and logic, such as *therefore* (所以), *suppose* (假如), and *analyze* (分析). Are cognitive process words tapping into cultural thought style? Table 2 finds that provinces' use of cognitive process words on Weibo are significantly correlated with findings from an earlier study measuring analytic thought style among students across China[24].

**Causation.** The cognitive processes category also includes two more specific sub-categories: causation and certainty. "Cause" words relate to causality, such as *effect* (作用), *cause-and-effect* (因果), and *due to* (由于).

**Certainty.** Certainty includes words such as *certain* (确定), *definitely* (肯定), and *confident* (自信). Holistic thinkers may use fewer words expressing confidence and certainty because holistic thought emphasizes frequent change and humility about what we can know. For example, people in Korea were less surprised than Americans when their predictions failed to come true[34]. Greater certainty may also be related to the tendency to take action in individualistic, promotion-focused cultures (which we discuss below).

**Possibility and Openness.** The cognitive process sub-category of possibility and openness[a] expresses a willingness to explore. The category includes words like *suppose/hypothesize* (假设), *try out* (尝试), and *conjecture* (推测). People who use more of these words tend to score higher on the personality trait of openness to experience[2], express greater individuality[42], and participate more in class[2]. These words are also more common among people with a promotion focus[43] (discussed below).

**Non-Fluencies.** The LIWC non-fluencies category includes words like "uh" and "um" (such as 呃). There is some evidence that people use non-fluencies when they are uncertain and hesitant—the opposite of the certainty category[44].

**Assent.** The assent category includes words where the speaker expresses agreement, such as *yeah* (嗯) and *OK* (OK, 好吧). These words might reflect the speaker's desire to get along with the listener and avoid confrontation. Cultural psychologists and historians have argued that Western culture has traditionally encouraged more debate[33], while thinkers in China more often conceded "that other opinions had something to be said for them"[45].

## 2. Promotion Orientation and Emotion

**Achievement.** The machine learning created a new category of achievement words. There is some evidence linking interdependence to what researchers call "prevention focus"[46]. People with a prevention focus tend to see the world as a dangerous place. They focus on avoiding bad outcomes and feel relief when they prevent bad outcomes. In contrast, people with a promotion focus worry less about risks and instead focus on exploring and attaining good, new things.

Researchers originally thought of prevention focus as a personality trait, but later researchers found that cultures vary in their prevention focus[47,48]. For example, people in interdependent cultures respond more favorably to advertisements that focus on preventing bad outcomes, such as framing grape juice as containing antioxidants that lower the risk of heart disease[47]. In contrast, people in independent cultures respond with more favorable framing about attaining positive outcomes, such as framing the grape juice as giving people energy. Similarly, when researchers asked participants to think about being interdependent with other people, they became more prevention-oriented[48]. Thus, there is evidence linking interdependence with prevention focus.

---

[a] The LIWC dictionary titles this category "tentative." However, we think this title gives readers a misleading picture of the behaviors linked to it. Self-expression and participation are the opposite of "tentative." We submit that the title "possibility and openness" sticks more closely to behaviors this category correlates with.

Several LIWC categories are related to prevention and promotion. For example, achievement words focus on approaching and obtaining new things, such as *overcome* (克服), *triumph* (战胜), and *obtain* (获取).

**Positivity/Optimism.** Machine learning also created another category of words centered around positivity and optimism. These optimism and goal words include words such as *ideal* (理想), *goal* (目标), and *positive* (积极). They seem to reflect striving and positivity, which fit with the idea of promotion focus.

**Affect.** We ran analyses of the LIWC category of affect words, such as *sad* (伤心), *happy* (高兴), and *lose face* (丢脸). We analyzed affect for two reasons. First, affect words might be a sort of counterweight to cognitive process words. If rice-farming regions use fewer cognitive words, they might use more affect words instead. Second, testing affect words allows us to pull apart differences in emotion words in general versus specific emotion categories like positivity/optimism.

### 3. Self and Groups

**Self, I, and We.** Self words might be more common in individualistic cultures. For example, one study found that people in interdependent sub-cultures within the US use "we," "us," and "our" more, whereas people in independent sub-cultures use "I," "me," and "mine" more[49]. However, another study found mixed results[50]. We tested whether people in rice areas used more "we" and less "I". We also created a broader list of "self words," including basic words like "self" (自己,自我) and "personal" (个人).

**Humans and Universalism.** We argue that one common misunderstanding of collectivistic cultures is to think they are more social *in general*[17]. This assumption is apparent in self-report scales designed to measure interdependence. These scales often include items that ask about "other people," without specifying who those people are and whether they have a relationship with the respondent.

For example, one classic collectivism scale item reads, "To me, pleasure is spending time with others"[51]. This item makes sense if people generally don't distinguish between people of different relationships. However, we argue that interdependent cultures like rice cultures focus intensely on the type of relationship. The sorts of behaviors people associate with collectivism are concentrated in known, trusted relationships—family, close friends, and trusted co-workers[17].

If the other person is outside that circle, the behaviors sometimes flip. Counter-intuitively, it is individualistic cultures that care more about strangers and people in general. For example, trust toward strangers is *lower* in collectivistic cultures[52]. It is people in individualistic cultures that agree more with abstract statements like, "I feel good when I cooperate with others"[53].

In a similar line of research, studies have found that historically rice-farming cultures have lower relational mobility[27]. In cultures with low relational mobility, relationships tend to be more stable and longer lasting, although people have less freedom and choice over who they interact with. Across 39 societies, people in cultures with low relational mobility reported meeting fewer new acquaintances in the last month and having dated fewer people[27].

These prior findings led us to test whether people in rice areas would use more in-group words, whereas people in wheat areas would use more words to describe people in general.

LIWC has the category humans, which has some words that fit with an emphasis on universalism (such as *the people*, 人民), but others that do not (such as *self*, 自己). Therefore, we created categories that were more precise in terms of the size of the social network. The universalism category includes words about broad groups, such as *humanity* (人类), *the people* (人民), and *worldwide* (全球).

**In-Group/Out-Group Connecting.** We created two categories of in-group/out-group words: connecting and dividing. Both categories draw a distinction between in-group members versus people who are outside the group. The "connecting" category contains words that people often use when they want to connect with other people, such as *collective* (集体) and *compatriot* (同胞).

**In-Group/Out-Group Dividing.** In contrast, we also created a category for in-group/out-group words that identify near and far people in a dividing way, such as *non-local* (外地人) and *outsider* (外人).

**Fashion and Trends.** Previous research found that people in interdependent cultures are more likely to use shared social standards for traits and success, whereas people in individualistic cultures are more likely to use personally defined standards[54]. We speculated that the social focus of rice areas might mean they pay more attention to social trends. To test this, we created a category of words about fashion and trends, such as *hot* (to describe ideas and trends, 热门), *out-of-style* (过时), and *celebrities* (名流).

**Theoretical Contributions**

**Causes of Culture.** Having big data down to the prefecture level gives fine-grained data to test theories of the causes of cultural differences in China. Previous studies have tested for cultural differences across China[24,26,29]. However, the question of regional differences in China is far from settled, let alone a unified theory of *why* cultures differ. Furthermore, the sheer scale of this study surpasses prior studies in terms of the sample size, the number of fine-grained geographic units, and the number of psychological outcomes.

**New Dimensions of Cultural Differences.** One contribution of using natural language is that it allows us to explore a wider range of outcomes than previous research. Lab studies have been limited to a single attitude scale[26,29], a handful of lab tests[24,55], or a particular Census indicator such as patents[24,56]. Using natural language opens up many new possibilities for outcomes. For example, no prior research on rice farming has tested whether there are differences in interest in fashion/trends, emotion words, or universalism. Bottom-up, machine-learning categories can unearth new categories of cultural differences that are (a) not obvious from prior research or (b) not contained in existing LIWC categories.

**Legacy of Farming Culture in the Face of Modernization.** The large sample size allows us to test a theoretical question that has not been tested in prior studies with less diverse samples[24,29,55]: whether these historically rooted cultural differences are disappearing in more modernized areas of China. As China races ahead into modernization, it has regions firmly in the developed world and other regions still rooted in subsistence economies and poverty. For example, Shanghai has a GDP per capita on par with countries in Europe, whereas prefectures like Bijie are on par with developing countries like Algeria and El Salvador. This large and diverse dataset can allow us to test whether cultural differences are different in these two types of regions. If rice-wheat differences are disappearing in the face of modernization, we should find smaller differences—or even no differences—in China's modernized areas.

**Replication Outside of China.** Japan offers an important test of the rice theory. Although there is evidence for rice-wheat differences in China[24,55], the question of whether rice farming influenced culture is far from settled. Although we can statistically control for potential confounds in China, a stronger "stress test" of the theory is to check whether it applies in different contexts.

Although Japan and China have had much cultural exchange, Japan is different in important ways. For example, Japanese is from an entirely different language family from Chinese. Its Shintoism religion is distinct from religions in China. Historically, the central government had a stronger role in Japan, with higher taxes and more public goods per capita[57]. Its geography as an island nation has shielded it from historical forces, leaving Japan free from Genghis Khan's Mongolian Empire, for example. Testing whether rice farming influences culture in a different context provides an important empirical check on the theory.

**Results**

**1. Validation Tests for New Categories**

**Reliability.** We tested the validity of the newly created categories in three ways. First, we tested the internal consistency using KR20, a statistic similar to Cronbach's alpha but better suited to text analysis[58]. Because our constructs are culture-level constructs rather than individual-level constructs, we analyzed them at the group level (prefectures). Previous research has found that constructs that are reliable at the culture level do not always show up at the individual level[37].

All of the categories had reliabilities above the common cutoff of 0.70 (Table 1). This result suggests that words that we theorized are connected actually tend to occur together. The one exception was the self category, which was borderline at 0.68. This is probably because the self category only has eight words, and reliability scores "punish" measures with fewer items. Given that the reliability fell close to the cutoff despite having few items, we kept it in the main analyses.

**Discriminant Validity.** Next, we asked whether the newly created categories are different from previous categories. We tested this by checking whether the new categories are not highly correlated with the LIWC categories. Researchers have suggested correlations above 0.90 are clear signs of redundancy, 0.80 is a warning sign, and below 0.80 is acceptable[59]. All correlations were below 0.60 (Table S21). This result suggests that the new categories are measuring topics that are not already represented in the LIWC categories.

**Convergent Validity.** Finally, we tested whether the word categories that we expected to reflect collectivism and holistic thought actually correlated with behavioral markers of collectivism around China. As an external index benchmark of collectivism, we calculated collectivism indexes for provinces and prefectures. We followed prior studies by combining Census statistics on divorce rates (less collectivistic), percentage of people living alone (less collectivistic), and three-generation families (more collectivistic)[60,61]. For further validity criteria, we also used data on holistic thought across China[24], norm tightness[62] (which tends to be higher in collectivistic cultures), and implicit individualism on the sociogram task[24], which measures how large people draw the self versus how large they draw their friends.

If the word categories are really tapping into collectivism, they should correlate with these markers of collectivism. Some researchers have suggested external correlates should be above $r = 0.20$[63]. However, the limited number of provinces only gave us the statistical power to detect significant correlations above $r = 0.56$ (90% statistical power). Therefore, we focus on the overall pattern of correlations, rather than using a binary cutoff.

Table 2 shows the correlations for the 12 word categories that showed rice-wheat differences. Correlations are highlighted in green if they were in the theoretically expected direction and in red if they were in the wrong direction. All correlations were in the expected direction with the external markers of collectivism across China, except for the in-group/out-group dividing category. The dividing category may need to be considered in tandem with the connecting category (connecting minus dividing words), rather than on its own. Except for this category, the 11 word categories showed convergent validity with other markers of collectivism.

Of the four word categories that **did not** show consistent rice-wheat differences, the tests of convergent validity mostly failed (Table S20). For example, provinces that scored high on collectivism in behavioral indexes and psychological tests tended to use less "we." "I" and "we" continued to fail convergent validity checks after controlling for general pronoun use or calculating the ratio of "I" to "we." These results suggest that using "I" versus "we" in China is not tapping into collectivism, at least as measured by other markers of collectivism. In summary, 12 of the newly created word categories showed high internal consistency (reliability), clear discriminant validity from LIWC word categories, and acceptable convergent validity with external markers of collectivism and holistic thought.

Finally, we tested whether the 11 word categories that passed the external validity checks in Table 2 tap into an underlying dimension of collectivism. The word categories showed acceptable reliability for provinces (Cronbach's alpha = .69) and prefectures (alpha = .73).

**Table 1**

*Reliability of Newly Created Word Categories (Prefecture Level)*

| Word Category | KR20 |
|---|---|
| Achievement | 0.99 |
| Universalism | 0.89 |
| In/Outgroup: Connecting | 0.72 |
| In/Outgroup: Dividing | 0.78 |
| Self | 0.68 |
| Positivity/Optimism | 0.96 |
| Fashion and Trends | 0.95 |

Note: KR20 values are a measure of reliability similar to Cronbach's alpha but better suited to word frequencies[58]. Values above 0.70 are generally considered acceptable. The word categories are conceptualized at the culture level, so reliabilities are calculated at the prefecture level.

**Table 2**

*Convergent Validity Tests with Provincial Collectivism Index, Prefecture Collectivism Index, Norm Tightness, Holistic Thought, and Self-Inflation*

| | Word Category | Markers That Are Higher in Collectivistic Cultures | | | | | | | | Lower Self-Inflation | |
|---|---|---|---|---|---|---|---|---|---|---|---|
| | | Province Collectivism | | Pref. Collectivism Index | | Norm Tightness | | Holistic Thought | | | |
| | | r | P | r | P | r | P | r | P | r | P |
| Individualistic/Analytic | Cognitive Processes | -0.27 | 0.139 | -0.31 | 0.028 | -0.38 | 0.034 | -0.37 | 0.047 | 0.24 | 0.220 |
| | Causation | -0.41 | 0.023 | -0.35 | 0.012 | -0.07 | 0.727 | -0.36 | 0.049 | 0.22 | 0.264 |
| | Certainty | -0.23 | 0.211 | -0.30 | 0.035 | -0.51 | 0.003 | -0.42 | 0.020 | 0.23 | 0.240 |
| | Possibility/Openness | -0.24 | 0.189 | -0.27 | 0.054 | -0.33 | 0.075 | -0.47 | 0.010 | 0.25 | 0.199 |
| | Positivity/Optimism | -0.24 | 0.195 | -0.21 | 0.141 | -0.36 | 0.049 | -0.45 | 0.013 | 0.28 | 0.148 |
| | Achievement | -0.32 | 0.080 | -0.26 | 0.071 | -0.001 | 0.997 | -0.40 | 0.027 | 0.32 | 0.098 |
| | Universalism | -0.34 | 0.060 | -0.26 | 0.070 | -0.23 | 0.217 | -0.50 | 0.005 | 0.09 | 0.640 |
| | Humans | -0.45 | 0.011 | -0.29 | 0.041 | -0.18 | 0.337 | -0.26 | 0.162 | 0.00 | 0.995 |
| | In/Outgroup: Connecting | -0.27 | 0.150 | -0.19 | 0.189 | -0.56 | 0.001 | -0.46 | 0.010 | 0.19 | 0.327 |
| Coll./Holistic | In/Outgroup: Dividing | 0.04 | 0.825 | -0.14 | 0.318 | 0.16 | 0.383 | -0.15 | 0.429 | 0.06 | 0.780 |
| | Non-Fluencies | 0.24 | 0.191 | 0.24 | 0.092 | 0.01 | 0.939 | 0.22 | 0.251 | -0.28 | 0.157 |
| | Assent | 0.22 | 0.225 | 0.07 | 0.634 | 0.03 | 0.874 | 0.38 | 0.041 | -0.07 | 0.736 |

Note: Green shaded rows correlate in the theoretically consistent direction. Red shaded rows correlate in the inconsistent direction. The province and prefectural collectivism indexes are Z scores of (% 3-generation households - % living alone - % nuclear families – divorce to marriage ratio) based on prior indexes in the US and China[60,61]. The tightness of social norms comes from a survey of 11,662 people from 31 provinces[62]. Holistic thought comes from tests using the triad categorization task with 1,019 students from 30 provinces[24]. Self-inflation data comes from 515 college students from 28 provinces who completed the sociogram task[24]. In the sociogram task, participants draw circles to represent the self and their friends. People in individualistic cultures draw the self much larger than friends on average.

## 2. Cognition and Discourse

**Cognitive Words**

People in wheat areas consistently used more cognitive words (Figure 2). People in wheat-farming provinces used more cognitive process words than people in rice-farming provinces ($ß$ = -0.08, $P$ < 0.001, $r_{prov}$ = -0.44, Table 3). Zooming into the prefecture level revealed the same pattern ($ß$ = -0.08, $P$ < 0.001, $r_{pref}$ = -0.11, Table 4).

Differences were similar in the sub-categories. People in rice prefectures used fewer words related to causality ($ß$ = -0.08, $P$ < 0.001, $r_{pref}$ = -0.24), possibility/openness ($ß$ = -0.04, $P$ < 0.001, $r_{pref}$ = -0.07), and certainty ($ß$ = -0.06, $P$ < 0.001, $r_{pref}$ = -0.14). Rice-wheat differences were independent from education (Tables S3A-S3B). This makes sense with the idea that holistic and analytic thought are cultural thought styles, rather than cognitive abilities. Studies of students from top-ranked colleges in the US and China still find cultural differences in thought style[24]. Rice continued to predict fewer cognitive process words after accounting for gender, age, GDP, temperature, and all other variables listed in Table S1.

**Non-Fluencies**

People in rice-farming provinces used more non-fluencies such as "uh" and "um" ($ß$ = 0.06, $P$ < 0.001, $r_{prov}$ = 0.69). "Uh's" and "um's" may reflect the hesitation and circumspection related to the rice area's less frequent use of certainty words. This hesitation might also reflect the prevention focus of rice cultures (discussed below).

**Assent**

Similarly, people in rice provinces used more assent words ($ß$ = 0.09, $P$ < 0.001, $r_{prov}$ = 0.78). These words could reflect a hesitance toward debate and a desire to avoid conflict.

**Figure 2**

*People in Rice Provinces Use Fewer Universalism, Cognitive Process, and Positivity/Optimism Words, But More Assent Words*

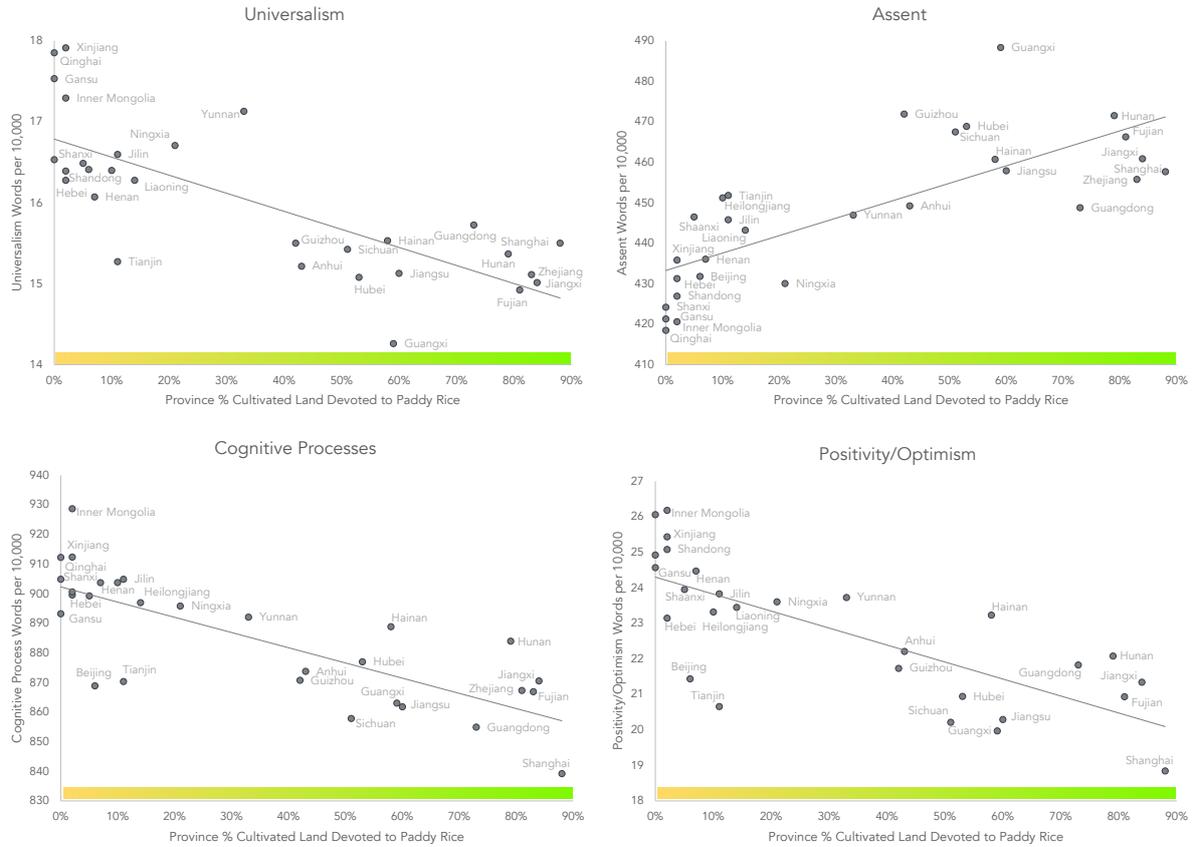

### 3. Promotion Orientation and Emotion

**Achievement**

In line with the idea that rice farming might cause a focus on prevention, people in rice-farming provinces used fewer achievement words ($\beta$ = -0.08, $P$ = 0.011, $r_{prov}$ = -0.75), such as *determination* (决心).

**Positivity/Optimism**

People in rice provinces also used fewer positivity/optimism words ($\beta$ = -0.09, $P$ < 0.001, $r_{prov}$ = -0.76).

**Affect**

Rice-wheat differences in cognitive words and positivity words seemed to be independent of differences in emotion words in general. People in rice and wheat provinces did not differ significantly in their use of affect words ($\beta$ = -0.02, $P$ = 0.148, $r_{prov}$ = -0.31). Given the large samples, the mixed results suggest there are not consistent rice-wheat differences in emotion words in general.

An interesting pattern emerged when we looked at differences in the sub-categories of affect words. Although there were no overall rice-wheat differences in affect, that seemed to be because rice areas used more positive emotion ($\beta$ = 0.03, $P$ < 0.001) and less negative emotion ($\beta$ = -0.02, $P$ = 0.014). The differences extended to the negative emotion subcategories of anger and sadness but not anxiety ($P$ = 0.457).

At first glance, this might seem to contradict the finding that rice areas use fewer positivity/optimism words. However, this highlights the distinction between the two categories. The positive emotion category includes a broad range of positive words, such as *thank you* (谢谢) and *kiss* (亲亲). In contrast, the positivity/optimism category focuses more narrowly on words about goals, striving, and achieving, such as *goal* (目标), *triumph* (克服), and *optimistic* (乐观).

**Table 3**

*Provinces' History of Rice Farming Predicts Word Use on Weibo*

| | Word Category | | β | t | P | | Word Category | | β | t | P | | Word Category | | β | t | P |
|---|---|---|---|---|---|---|---|---|---|---|---|---|---|---|---|---|---|
| Cognition and Discourse | Cognitive Processes | Female | 0.08 | 40.98 | < 0.001 | Self and Groups | Universalism | Female | -0.08 | -36.85 | < 0.001 | Promotion Orientation and Emotion | Positivity/ Optimism | Female | 0.001 | 0.66 | 0.507 |
| | | GDP | 0.02 | 0.51 | 0.614 | | | GDP | -0.01 | -0.18 | 0.856 | | | GDP | 0.001 | 0.03 | 0.975 |
| | | % Urban | -0.06 | -1.32 | 0.200 | | | % Urban | 0.002 | 0.05 | 0.962 | | | % Urban | -0.04 | -0.82 | 0.420 |
| | | Rice % | -0.08 | -5.81 | < 0.001 | | | Rice % | -0.06 | -5.26 | < 0.001 | | | Rice % | -0.09 | -5.56 | < 0.001 |
| | Causation | Female | -0.12 | -58.55 | < 0.001 | | Humans | Female | 0.11 | 55.51 | < 0.001 | | Achievement | Female | -0.21 | -109.01 | < 0.001 |
| | | GDP | 0.03 | 0.68 | 0.503 | | | GDP | 0.01 | 0.18 | 0.860 | | | GDP | 0.02 | 0.39 | 0.697 |
| | | % Urban | -0.02 | -0.53 | 0.604 | | | % Urban | -0.06 | -1.02 | 0.318 | | | % Urban | -0.004 | -0.08 | 0.934 |
| | | Rice % | -0.08 | -5.92 | < 0.001 | | | Rice % | -0.09 | -5.36 | < 0.001 | | | Rice % | -0.08 | -5.40 | < 0.001 |
| | Certainty | Female | 0.09 | 45.00 | < 0.001 | | In/Outgroup: Connecting | Female | 0.0003 | 0.12 | 0.902 | | Fashion and Trends | Female | 0.04 | 15.57 | < 0.001 |
| | | GDP | -0.01 | -0.34 | 0.735 | | | GDP | 0.02 | 0.59 | 0.560 | | | GDP | -0.003 | -0.11 | 0.916 |
| | | % Urban | -0.03 | -0.69 | 0.497 | | | % Urban | -0.07 | -1.80 | 0.086 | | | % Urban | 0.02 | 0.51 | 0.617 |
| | | Rice % | -0.06 | -5.68 | < 0.001 | | | Rice % | -0.03 | -2.80 | 0.011 | | | Rice % | 0.002 | 0.26 | 0.793 |
| | Possibility/ Openness | Female | 0.08 | 37.81 | < 0.001 | | In/Outgroup: Dividing | Female | -0.01 | -2.94 | 0.003 | | Affect | Female | 0.26 | 137.07 | < 0.001 |
| | | GDP | 0.02 | 0.52 | 0.607 | | | GDP | -0.09 | -1.95 | 0.062 | | | GDP | -0.01 | -0.28 | 0.785 |
| | | % Urban | -0.05 | -1.26 | 0.219 | | | % Urban | 0.12 | 2.27 | 0.033 | | | % Urban | -0.06 | -1.27 | 0.216 |
| | | Rice % | -0.04 | -3.85 | < 0.001 | | | Rice % | 0.01 | 0.36 | 0.724 | | | Rice % | -0.02 | -1.49 | 0.148 |
| | Assent | Female | 0.18 | 93.52 | < 0.001 | | I | Female | 0.21 | 106.47 | < 0.001 | | We | Female | 0.03 | 12.32 | < 0.001 |
| | | GDP | -0.01 | -0.20 | 0.840 | | | GDP | -0.08 | -1.75 | 0.092 | | | GDP | 0.02 | 0.76 | 0.458 |
| | | % Urban | -0.01 | -0.29 | 0.772 | | | % Urban | -0.01 | -0.14 | 0.887 | | | % Urban | -0.07 | -1.93 | 0.068 |
| | | Rice % | 0.09 | 6.09 | < 0.001 | | | Rice % | -0.01 | -0.71 | 0.487 | | | Rice % | -0.03 | -2.94 | 0.009 |
| | Non-Fluencies | Female | 0.18 | 90.23 | < 0.001 | | Self | Female | -0.05 | -17.62 | < 0.001 | | | | Province | | Prefecture | |
| | | GDP | -0.02 | -0.37 | 0.716 | | | GDP | -0.02 | -0.49 | 0.629 | | Geogr. Units: | | 29 | | 421 | |
| | | % Urban | -0.002 | -0.04 | 0.966 | | | % Urban | 0.02 | 0.66 | 0.516 | | Located Users: | | 249,361 | | 80,825 | |
| | | Rice % | 0.06 | 4.70 | < 0.001 | | | Rice % | -0.01 | -0.51 | 0.616 | | Distinct Terms[a]: | | 4,955,629 | | 2,711,765 | |
| | | | | | | | | | | | | | Total Terms[b]: | | 1,002,453,505 | | 345,423,348 | |

Note: Analyses are hierarchical linear models nested in provinces. Urbanization is indexed by the percent of urban residents per province in 2016. Provincial GDP per capita statistics are from 2014 in RMB. Province rice is the percentage of cultivated land devoted to rice paddies. Summary statistics in the lower right differ depending on the availability of control variables such as age in different analyses. [a]Distinct terms are the number of *different* words in the dataset. [b]Total terms is the total number of words in the dataset, allowing for counting of each appearance.

**Table 4**

*Prefectures' History of Rice Farming Predicts Word Use on Weibo*

| | Word Category | | β | t | P | | Word Category | | β | t | P | | Word Category | | β | t | P |
|---|---|---|---|---|---|---|---|---|---|---|---|---|---|---|---|---|---|
| | | Female | 0.08 | 40.63 | < 0.001 | | | Female | -0.08 | -37.13 | < 0.001 | | | Female | 0.001 | 0.33 | 0.739 |
| | Cognitive Processes | GDP | -0.04 | -3.36 | 0.002 | | Universalism | GDP | -0.007 | -0.71 | 0.482 | | Positivity/ Optimism | GDP | -0.04 | -3.22 | 0.003 |
| | | Rice % | -0.08 | -6.77 | < 0.001 | | | Rice % | -0.06 | -5.74 | < 0.001 | | | Rice % | -0.08 | -6.34 | < 0.001 |
| | | Female | -0.12 | -59.27 | < 0.001 | | | Female | 0.11 | 55.41 | < 0.001 | | | Female | -0.21 | -110.09 | < 0.001 |
| | Causation | GDP | 0.004 | 0.36 | 0.719 | | Humans | GDP | -0.05 | -3.30 | 0.003 | | Achievement | GDP | 0.01 | 1.13 | 0.271 |
| | | Rice % | -0.08 | -6.41 | < 0.001 | | | Rice % | -0.09 | -5.92 | < 0.001 | | | Rice % | -0.08 | -5.53 | < 0.001 |
| Cognition and Discourse | | Female | 0.09 | 44.87 | < 0.001 | | | Female | 0.0001 | 0.05 | 0.959 | Promotion Orientation and Emotion | | Female | 0.04 | 15.78 | < 0.001 |
| | Certainty | GDP | -0.04 | -4.21 | < 0.001 | | In/Outgroup: Connecting | GDP | -0.04 | -4.03 | < 0.001 | | Fashion and Trends | GDP | 0.01 | 1.40 | 0.174 |
| | | Rice % | -0.06 | -6.33 | < 0.001 | Self and Groups | | Rice % | -0.04 | -3.20 | 0.004 | | | Rice % | 0.004 | 0.47 | 0.645 |
| | | Female | 0.07 | 37.67 | < 0.001 | | | Female | -0.02 | -3.01 | 0.003 | | | Female | 0.26 | 137.71 | < 0.001 |
| | Possibility/ Openness | GDP | -0.03 | -3.22 | 0.004 | | In/Outgroup: Dividing | GDP | 0.01 | 1.08 | 0.292 | | Affect | GDP | -0.07 | -6.04 | < 0.001 |
| | | Rice % | -0.04 | -4.62 | < 0.001 | | | Rice % | 0.01 | 1.04 | 0.307 | | | Rice % | -0.02 | -1.53 | 0.139 |
| | | Female | 0.02 | 10.56 | < 0.001 | | | Female | 0.21 | 106.75 | < 0.001 | | | Female | 0.03 | 12.37 | < 0.001 |
| | Assent | GDP | -0.01 | -1.13 | 0.272 | | I | GDP | -0.08 | -6.65 | < 0.001 | | We | GDP | -0.03 | -3.54 | 0.002 |
| | | Rice % | -0.001 | -0.08 | 0.936 | | | Rice % | -0.01 | -0.40 | 0.694 | | | Rice % | -0.03 | -3.30 | 0.003 |
| | | Female | 0.18 | 90.88 | < 0.001 | | | Female | -0.05 | -17.78 | < 0.001 | | | | Province | | Prefecture |
| | Non-Fluencies | GDP | -0.01 | -1.22 | 0.233 | | Self | GDP | 0.005 | 0.56 | 0.581 | | Geogr. Units: Located Users: Distinct Terms: Total Terms: | | 29 249,361 4,955,629 1,002,453,505 | | 421 80,825 2,711,765 345,423,348 |
| | | Rice % | 0.06 | 5.20 | < 0.001 | | | Rice % | -0.005 | -0.49 | 0.626 | | | | | | |

Note: Analyses are hierarchical linear models nested in prefectures. Prefecture rice is the percentage of cultivated land devoted to rice paddies. We used the earliest rice statistics we could find for each province (2001 in most cases). Prefecture GDP data is from 2014.

## 4. Self and Groups

**Fashion and Trends**

Results for fashion and trends were mixed. Differences between provinces were not significant (Table 4). But differences between prefectures were significant. People in rice-farming prefectures used significantly more fashion/trend words after controlling for age ($\beta$ = 0.03, $P$ = 0.022, $r_{\text{pref}}$ = 0.02, Table S4B).

**Universalism versus Groups**

People in rice provinces used fewer universalistic ($\beta$ = -0.06, $P$ < 0.001, $r_{\text{prov}}$ = -0.76) and humanity words ($\beta$ = -0.09, $P$ < 0.001, $r_{\text{prov}}$ = -0.76). This fits with the idea that rice farming was built around close relationships rather than the loose ties of wheat farming[27,64]. Looking at the two in-group/out-group categories, people in rice provinces used fewer words that *connected* across in-groups and out-groups, such as *compatriot* (同胞, $\beta$ = -0.03, $P$ = 0.011, $r_{\text{prov}}$ = -0.53). But this trend did not extend to words that tend to imply divisions between in-groups and out-groups; people in rice-farming provinces were just as likely to use dividing words like *outsider* (外人, $\beta$ = 0.01, $P$ = 0.724, $r_{\text{prov}}$ = 0.25). In sum, people in rice-farming areas used fewer words about broad social ties and connecting across groups.

**Self, I, and We**

Against our predictions, people in rice provinces used less "we" ($\beta$ = -0.04, $P$ < 0.001, $r_{\text{pref}}$ = -0.01) and "I" ($\beta$ = -0.02, $P$ = 0.023, $r_{\text{pref}}$ = -0.01). People in rice prefectures used marginally fewer self words ($\beta$ = -0.01, $P$ = 0.156, $r_{\text{pref}}$ = -0.11). Rice-wheat differences in these categories were weaker between provinces than between prefectures (Table 3). These categories failed most tests of convergent validity with other markers of collectivism (Table S20).

The finding that people in rice-farming prefectures used both "I" and "we" significantly less is puzzling. One explanation could be "pronoun drop." Pronoun drop is when the speaker leaves out the pronoun and relies on the context or other cues to communicate the pronoun. For example, in Chinese, it is natural to say "didn't hear it" (没听见) and drop the "I." If rice areas are dropping pronouns more often than people in wheat areas, this could explain why rice areas are using less "I" and "we."

In line with this reasoning, one study compared cultures around the world and found that collectivistic cultures drop pronouns more[65]. The researchers argued that this is because naming the subject often emphasizes the individual actor, which makes more sense in a culture that emphasizes individual agency. In contrast, dropping the pronoun subtly emphasizes the situation. Emphasizing the context fits with the idea that people in collectivistic cultures see behavior more in terms of the situation[33,66].

In line with the conjecture that pronoun drop is more common in collectivistic cultures, people in rice areas used fewer of all types of pronouns ($\beta$ = -0.10, $P$ < 0.001, $r_{\text{prov}}$ = -0.60). If this trend within China replicates in future studies, it would present an interesting test of the theory that dropping is more common in collectivistic cultures[65]. Differences like this within China are valuable to theory building because these differences are among speakers of the same language family. Previous research that found that collectivistic cultures are more likely to drop pronouns were comparing entire languages (such as comparing Chinese and English), rather than differences within a single language[65].

**Rice-Wheat Border Analysis**

Comparing regions within China provides a cleaner comparison of cultural differences than comparing across countries. However, there are still differences between rice and wheat regions in China. For example, rice regions are at lower latitudes and tend to be hotter. Rice regions are also farther from herding cultures that have played an important role in Chinese history, such as the Mongolians.

One way to help rule out the influence of many potential confounds between rice and wheat regions is to analyze differences among only neighboring prefectures along the rice-wheat border. This analysis takes advantage of a convenient quirk of geography in China. While factors like temperature decrease bit by bit going north, the transition from rice farming to wheat farming is abrupt[67]. For example, Anhui province has nearby prefectures that farm 1% rice and 89% rice. Comparing prefectures along the rice-wheat border provides a more controlled comparison of places that differ strongly in rice and wheat but minimally in potential confounds like temperature.

We tested whether the rice-wheat differences for China as a whole replicated along the rice-wheat border provinces of Sichuan, Chongqing, Hubei, Anhui, and Jiangsu. Of the 11 word categories that differed significantly across China (Table 4), eight were significant along the rice-wheat border (Table 5). Two were marginally significant ("we" and in-group/out-group connecting, $P$s = 0.07). One was non-significant (universalism). In sum, analyses along the rice-wheat border mostly replicated the larger rice-wheat differences across China. Only one category failed to show a similar trend. This more controlled comparison suggests that rice-wheat differences are due to rice farming and not larger differences in temperature or latitude.

Table 5

*Differences in Word Use Are Similar Comparing Just Nearby Prefectures Along the Rice-Wheat Border*

| Word Category | Rice Side Words per 10,000 | Wheat Side Words per 10,000 | t | P | 95% CI | |
|---|---|---|---|---|---|---|
| Cognitive Processes | 861.12 | 871.21 | 2.91 | 0.004 | [3.29 | 16.89] |
| Causation | 87.50 | 89.19 | 2.84 | 0.005 | [0.52 | 2.86] |
| Certainty | 120.94 | 122.36 | 2.10 | 0.036 | [0.09 | 2.75] |
| Possibility/Openness | 132.45 | 134.58 | 2.61 | 0.009 | [0.53 | 3.72] |
| Assent | 460.57 | 446.12 | -5.58 | < 0.001 | [-19.53 | -9.38] |
| Non-Fluencies | 54.80 | 53.15 | -2.94 | 0.003 | [-2.76 | -0.55] |
| Universalism | 15.21 | 15.24 | 0.13 | 0.897 | [-0.38 | 0.44] |
| Humans | 96.90 | 101.32 | 5.34 | < 0.001 | [2.80 | 6.04] |
| In/Outgroup: Connecting | 19.45 | 19.89 | 1.78 | 0.074 | [-0.04 | 0.91] |
| Positivity/Optimism | 20.48 | 22.28 | 5.67 | < 0.001 | [1.18 | 2.42] |
| Achievement | 112.96 | 117.24 | 4.86 | < 0.001 | [2.56 | 6.01] |
| We | 23.36 | 23.85 | 1.77 | 0.077 | [-0.05 | 1.02] |

Note: This table tests along the rice-wheat border for the word categories that showed significant differences in the analysis over all of China (Table 4). The border runs through Sichuan, Chongqing, Hubei, Jiangsu, and Anhui. Prefectures in these provinces are defined as rice if they devote more than 50% of cultivated land to paddies. These nearby prefectures differ sharply in rice[17] but very little in temperature, latitude, distance from contact with herding cultures, and other potential confounds.

**Controlling for Pronoun Drop**

In response to an earlier draft, an anonymous reviewer suggested running analyses controlling for the percentage of pronoun drop. Controlling for pronoun drop could accomplish two things:

(1) It could separate social differences from differences in thought style. If pronoun drop is a marker of holistic thought, controlling for pronoun drop could allow us to test whether social differences such as self words and universalism words are separate from thought style differences.

(2) Pronoun drop could represent a larger pattern of dropping substantive words. It is possible that people who drop pronouns also tend to leave out key words and instead rely on the context to fill in the details[68]. For example, people could leave out the key substantive word by saying, "I wish you wouldn't be so…" or "Don't be like *that*." If pronoun drop is an indicator of leaving out key words, controlling for pronoun drop would give us one method to check whether rice-wheat differences are separate from patterns of omitting words.

To measure pronoun use, we calculated the percentages of words that were pronouns for each user. Results showed that rice-wheat differences remained significant after controlling for differences in pronoun use (Table 6). These results suggest that rice-wheat differences are independent of pronoun drop. Also, if pronoun drop correlates with a general pattern of dropping words (contextual communication), then this analysis suggests that the rice-wheat differences we found are independent from differences in contextual communication.

However, there was one exception. Rice-wheat differences in possibility/openness became non-significant after controlling for pronoun use ($P = 0.782$). People who used more pronouns tended to use more possibility/openness words. This presents an interesting puzzle for future research. We hazard a potential explanation. Possibility/openness often involves imagining new possibilities. Many possibility/openness words in the LIWC dictionary involve imagining different realities, such as, "imagine that…" Because these are new thoughts, they may be more abstract and less tied to a particular context. This explanation would fit with prior research that has described individualistic, Western cultures as "low context" communication cultures, in contrast to interdependent, "high context" communication cultures like China and Japan[68,69].

Overall, the results also suggest that rice-wheat differences are not an artifact of dropping key words. However, pronoun drop is only one measure of contextual communication. Future research could explore new methods of measuring contextual communication.

**Table 6**

*Differences Between Rice and Wheat Provinces Are Similar After Controlling for Individual Users' Frequency of Pronoun Drop*

| | Word Category | | β | t | P | | Word Category | | β | t | P | | Word Category | | β | t | P |
|---|---|---|---|---|---|---|---|---|---|---|---|---|---|---|---|---|---|
| | | Female | -0.07 | -48.10 | < 0.001 | | | Female | -0.09 | -42.65 | < 0.001 | | | Female | -0.07 | -33.81 | < 0.001 |
| | | % Pron. | 0.69 | 454.24 | < 0.001 | | | % Pron. | 0.06 | 29.41 | < 0.001 | | | % Pron. | 0.31 | 153.24 | < 0.001 |
| | Cognitive Processes | GDP | 0.04 | 1.58 | 0.127 | | Universalism | GDP | -0.002 | -0.07 | 0.945 | | Positivity/ Optimism | GDP | 0.01 | 0.32 | 0.750 |
| | | % Urban | -0.01 | -0.34 | 0.735 | | | % Urban | 0.004 | 0.11 | 0.917 | | | % Urban | -0.02 | -0.55 | 0.589 |
| | | Rice % | -0.02 | -2.78 | 0.011 | | | Rice % | -0.06 | -5.08 | < 0.001 | | | Rice % | -0.06 | -4.71 | < 0.001 |
| | | Female | -0.17 | -85.51 | < 0.001 | | | Female | -0.02 | -14.74 | < 0.001 | | | Female | -0.23 | -113.36 | < 0.001 |
| | | % Pron. | 0.24 | 118.91 | < 0.001 | | | % Pron. | 0.61 | 370.20 | < 0.001 | | | % Pron. | 0.06 | 30.18 | < 0.001 |
| | Causation | GDP | 0.03 | 0.73 | 0.473 | | Humans | GDP | 0.03 | 0.78 | 0.445 | | Achievement | GDP | 0.02 | 0.38 | 0.707 |
| | | % Urban | 0.002 | 0.04 | 0.967 | | | % Urban | -0.02 | -0.44 | 0.661 | | | % Urban | 0.05 | 0.62 | 0.544 |
| Cognition and Discourse | | Rice % | -0.06 | -5.02 | < 0.001 | | | Rice % | -0.04 | -3.65 | 0.001 | Promotion Orientation and Emotion | | Rice % | -0.08 | -5.39 | < 0.001 |
| | | Female | -0.02 | -13.31 | < 0.001 | | | Female | -0.02 | -5.65 | < 0.001 | | | Female | 0.06 | 24.46 | < 0.001 |
| | | % Pron. | 0.51 | 286.38 | < 0.001 | | | % Pron. | 0.05 | 12.70 | < 0.001 | | | % Pron. | -0.10 | -40.00 | < 0.001 |
| | Certainty | GDP | 0.005 | 0.24 | 0.815 | Self and Groups | In/Outgroup: Connecting | GDP | -0.03 | -0.73 | 0.476 | | Fashion and Trends | GDP | -0.01 | -0.26 | 0.796 |
| | | % Urban | 0.01 | 0.33 | 0.745 | | | % Urban | 0.01 | 0.34 | 0.741 | | | % Urban | 0.01 | 0.37 | 0.717 |
| | | Rice % | -0.02 | -3.03 | 0.006 | | | Rice % | 0.01 | 0.65 | 0.523 | | | Rice % | -0.01 | -0.69 | 0.497 |
| | | Female | -0.05 | -26.23 | < 0.001 | | | Female | -0.04 | -7.47 | < 0.001 | | | Female | 0.15 | 88.82 | < 0.001 |
| | | % Pron. | 0.55 | 311.01 | < 0.001 | | | % Pron. | 0.08 | 16.06 | < 0.001 | | | % Pron. | 0.51 | 296.80 | < 0.001 |
| | Possibility/ Openness | GDP | 0.03 | 1.44 | 0.162 | | In/Outgroup: Dividing | GDP | -0.07 | -1.45 | 0.160 | | Affect | GDP | 0.01 | 0.43 | 0.671 |
| | | % Urban | -0.01 | -0.34 | 0.734 | | | % Urban | 0.10 | 1.92 | 0.068 | | | % Urban | -0.04 | -1.32 | 0.200 |
| | | Rice % | 0.001 | 0.09 | 0.929 | | | Rice % | 0.02 | 1.16 | 0.260 | | | Rice % | 0.02 | 2.48 | 0.021 |
| | | Female | 0.14 | 71.44 | < 0.001 | | | Female | -0.07 | -26.75 | < 0.001 | | | Female | 0.14 | 70.65 | < 0.001 |
| | | % Pron. | 0.19 | 95.00 | < 0.001 | | | % Pron. | 0.11 | 41.11 | < 0.001 | Cog. & Dis.* | | % Pron. | 0.17 | 85.59 | < 0.001 |
| | Assent | GDP | 0.01 | 0.25 | 0.802 | | Self | GDP | -0.01 | -0.23 | 0.817 | | Non-Fluencies | GDP | -0.004 | -0.10 | 0.918 |
| | | % Urban | -0.02 | -0.35 | 0.729 | | | % Urban | 0.03 | 0.74 | 0.468 | | | % Urban | 0.004 | 0.09 | 0.931 |
| | | Rice % | 0.11 | 7.21 | < 0.001 | | | Rice % | 0.01 | 0.53 | 0.604 | | | Rice % | 0.08 | 5.98 | < 0.001 |

Note: This analysis controls for individual users' prevalence of pronoun use (percentage of words that are pronouns, such as *you* and *I*). This analysis does not include the "in-group/out-group connecting" category because it includes a pronoun and thus is highly correlated with pronoun use. Gender and pronoun use are characteristics of individual users. GDP, urbanization, and rice are province variables. Models are hierarchical linear models with users nested in provinces. *Cognition and discourse.

**Reverse Causality**

One challenge with trying to test whether rice farming influenced culture in China is reverse causality. If all of China can farm rice, and it is just people in certain areas who *choose* to farm rice, then perhaps some people were collectivistic to begin with, and they decided to farm rice. Perhaps people in certain areas had more social cohesion to begin with, and that led them to pick up rice farming.

One way to test this is to ask where it is physically possible to farm rice in China. The United Nations Food and Agriculture Organization estimates where it is possible to farm wetland rice using a range of environmental conditions, such as rainfall, temperature, soil, and terrain. What is important for our purposes is that these "rice suitability" values apply to plots of farmland regardless of whether people there are actually farming rice.

These suitability values can tell us two things. First, suitability can tell us whether all of China could grow rice if people wanted to. The answer here is clearly no. Large swaths of China—14 provinces in total—have suitability scores of zero. Instead, environmental suitability for rice strongly predicted where people actually farm rice in China, $\beta = 0.86$, $F = 1063.50$, $P < 0.001$. Thus, people farm rice in China mostly where it is ecologically possible; it is *not* the case that large parts of China *could* farm rice but just chose not to. This makes reverse causality less likely.

Second, we re-ran the main analyses removing actual rice farming and using environmental suitability instead (an instrumental variable analysis). By removing the variable that is potentially selected by humans (farming rice) and replacing it with a variable that is mostly out of traditional humans' hands (the climate), we can gain more insight into whether rice is shaping culture (causality) or whether certain types of people choose to farm rice (self-selection). The results replicated the main analysis both for prefecture rice suitability (Table 7) and province suitability (Table S7). These results suggest that reverse causality is not driving rice-wheat differences in China.

**Table 7**

*Prefecture Environmental Suitability for Rice Predicts Word Use on Weibo*

| | Word Category | | β | t | P | | Word Category | | β | t | P | | Word Category | | β | t | P |
|---|---|---|---|---|---|---|---|---|---|---|---|---|---|---|---|---|---|
| | | Female | 0.08 | 24.04 | < 0.001 | | | Female | -0.09 | -23.22 | < 0.001 | | | Female | -0.01 | -1.68 | 0.092 |
| | Cognitive Processes | GDP | -0.02 | -1.82 | 0.070 | | Universalism | GDP | 0.001 | 0.18 | 0.857 | | Positivity/ Optimism | GDP | -0.02 | -2.82 | 0.005 |
| | | Rice Suit. | -0.07 | -6.71 | < 0.001 | | | Rice Suit. | -0.01 | -1.16 | 0.248 | | | Rice Suit. | -0.06 | -6.04 | < 0.001 |
| | | Female | -0.13 | -37.56 | < 0.001 | | | Female | 0.12 | 33.84 | < 0.001 | | | Female | -0.23 | -67.30 | < 0.001 |
| | Causation | GDP | 0.02 | 2.49 | 0.014 | | Humans | GDP | -0.04 | -3.91 | < 0.001 | | Achievement | GDP | 0.02 | 3.10 | 0.002 |
| Cognition and Discourse | | Rice Suit. | -0.05 | -4.39 | < 0.001 | | | Rice Suit. | -0.05 | -4.19 | < 0.001 | Promotion Orientation and Emotion | | Rice Suit. | -0.04 | -4.15 | < 0.001 |
| | | Female | 0.10 | 27.56 | < 0.001 | Self and Groups | | Female | -0.03 | -3.57 | < 0.001 | | | Female | 0.02 | 3.91 | < 0.001 |
| | Certainty | GDP | -0.02 | -2.59 | 0.010 | | In/Outgroup: Connecting | GDP | -0.01 | -1.16 | 0.246 | | Fashion and Trends | GDP | 0.01 | 2.15 | 0.033 |
| | | Rice Suit. | -0.06 | -7.13 | < 0.001 | | | Rice Suit. | -0.004 | -0.41 | 0.682 | | | Rice Suit. | 0.01 | 1.36 | 0.178 |
| | | Female | 0.08 | 22.78 | < 0.001 | | | Female | -0.03 | -3.40 | < 0.001 | | | Female | 0.28 | 82.03 | < 0.001 |
| | Possibility/ Openness | GDP | -0.02 | -2.36 | 0.019 | | In/Outgroup: Dividing | GDP | -0.001 | -0.07 | 0.948 | | Affect | GDP | -0.05 | -6.40 | < 0.001 |
| | | Rice Suit. | -0.05 | -6.83 | < 0.001 | | | Rice Suit. | 0.01 | 0.85 | 0.398 | | | Rice Suit. | -0.06 | -5.90 | < 0.001 |
| | | Female | 0.21 | 59.73 | < 0.001 | | | Female | -0.05 | -10.54 | < 0.001 | C. & D.* | | Female | 0.20 | 57.44 | < 0.001 |
| | Assent | GDP | -0.02 | -2.22 | 0.027 | | Self | GDP | 0.01 | 1.70 | 0.091 | | Non-Fluencies | GDP | -0.02 | -2.34 | 0.020 |
| | | Rice Suit. | 0.02 | 1.93 | 0.056 | | | Rice Suit. | -0.01 | -1.17 | 0.244 | | | Rice Suit. | 0.02 | 1.79 | 0.075 |

Note: Analyses are hierarchical linear models nested in prefectures. Rice suitability is an instrumental variable that reduces the potential for reverse causality. Suitability is an index of environmental variables (such as rainfall) that determine where it is physically possible to grow paddy rice, regardless of whether people are farming rice there. We indexed herding using the percentage of traditionally herding ethnicities in each province, according to the 2000 Census. For example, Mongolian and Manchu ethnicities herded traditionally. The supplemental materials present the full list of herding groups. Provincial GDP per capita statistics are from 2014 in RMB. *Cognition and discourse.

**Are Differences Due to Dialect?**

Differences in dialects could confound tests of rice-wheat differences, especially because southern rice areas have more diverse dialects than the northern wheat areas[70]. Although written Chinese often allows words from different dialects to be written the same way, there are still words that are unique to different dialects. These dialectical words are unlikely to appear in the Chinese LIWC dictionary and thus may undercount words from non-Mandarin-speaking areas. We used a Chinese version of LIWC, which was previously validated[32].

**Cantonese.** We ran two analyses to pull apart rice-wheat differences from dialect differences. First, we ran analyses excluding Cantonese, which is arguably the most distinct and most culturally developed dialect. Cantonese is highly developed in part because of the historical importance of Hong Kong, which produced many popular movies and songs in the era when most media in Mainland China was limited to strictly socialist stories. Cantonese has the most developed system of non-Mandarin characters (such as 係), which do not appear in the LIWC dictionary. This could be an important confound in testing the rice theory because Cantonese areas are also some of the highest rice-producing areas in China.

To test whether rice-wheat differences were a confound of Cantonese, we re-ran the main analyses excluding the two Cantonese-speaking Mainland provinces (Guangdong and Guangxi). After excluding Cantonese-speaking areas, rice-wheat differences that were significant in the main analyses remained significant. Thus, rice-wheat differences exist independently from Cantonese-speaking regions.

**Mandarin Only.** Second, we ran a more conservative analysis limited to Mandarin-speaking provinces only. To categorize provinces, we used the *Language Atlas of China*[70], which denotes nine of the 29 provinces in the sample as speaking a non-Mandarin Chinese dialect. Although this eliminates many provinces, it still leaves enough provinces and prefectures to run analyses. It also leaves variation in rice. Mandarin areas still vary from 0% to 60% paddy rice.

Results from Mandarin-speaking areas show that rice-wheat differences remain significant (Table 8). Rice-wheat differences in cognitive words, achievement words, and all other categories that were robust in the main analyses remain robust in the Mandarin-only analyses.

**Table 8**

*Robustness to Dialect: Mandarin-Speaking Areas Only*

| | Word Category | | $\beta$ | $t$ | $P$ | | Word Category | | $\beta$ | $t$ | $P$ | | Word Category | | $\beta$ | $t$ | $P$ |
|---|---|---|---|---|---|---|---|---|---|---|---|---|---|---|---|---|---|
| Cognition and Discourse | Cognitive Processes | Female | 0.09 | 33.74 | < 0.001 | Self and Groups | Universalism | Female | -0.08 | -26.16 | < 0.001 | Promotion Orientation and Emotion | Positivity/ Optimism | Female | -0.004 | -1.42 | 0.156 |
| | | GDP | -0.04 | -3.64 | 0.003 | | | GDP | -0.02 | -2.14 | 0.056 | | | GDP | -0.05 | -3.67 | 0.003 |
| | | Rice % | -0.07 | -5.83 | < 0.001 | | | Rice % | -0.04 | -4.15 | 0.002 | | | Rice % | -0.08 | -5.77 | < 0.001 |
| | Causation | Female | -0.12 | -44.05 | < 0.001 | | Humans | Female | 0.12 | 42.75 | < 0.001 | | Achievement | Female | -0.22 | -83.99 | < 0.001 |
| | | GDP | -0.003 | -0.26 | 0.796 | | | GDP | -0.06 | -4.51 | < 0.001 | | | GDP | 0.004 | 0.26 | 0.802 |
| | | Rice %. | -0.07 | -5.63 | < 0.001 | | | Rice % | -0.08 | -5.80 | < 0.001 | | | Rice % | -0.07 | -5.11 | < 0.001 |
| | Certainty | Female | 0.10 | 35.76 | < 0.001 | | In/Outgroup: Connecting | Female | -0.01 | -2.17 | 0.030 | | Fashion and Trends | Female | 0.04 | 11.01 | < 0.001 |
| | | GDP | -0.04 | -5.81 | < 0.001 | | | GDP | -0.02 | -1.88 | 0.097 | | | GDP | 0.005 | 0.48 | 0.641 |
| | | Rice % | -0.05 | -6.78 | < 0.001 | | | Rice % | -0.01 | -0.50 | 0.633 | | | Rice % | -0.005 | -0.47 | 0.643 |
| | Possibility/ Openness | Female | 0.08 | 28.61 | < 0.001 | | In/Outgroup: Dividing | Female | -0.01 | -2.15 | 0.032 | | Affect | Female | 0.29 | 109.72 | < 0.001 |
| | | GDP | -0.03 | -2.90 | 0.013 | | | GDP | -0.001 | -0.04 | 0.967 | | | GDP | -0.07 | -4.54 | < 0.001 |
| | | Rice % | -0.04 | -4.08 | 0.001 | | | Rice % | -0.0004 | -0.02 | 0.983 | | | Rice % | -0.01 | -0.84 | 0.414 |
| | Assent | Female | 0.21 | 79.33 | < 0.001 | | Self | Female | -0.05 | -14.78 | < 0.001 | C. & D.* | Non-Fluencies | Female | 0.19 | 71.79 | < 0.001 |
| | | GDP | -0.01 | -0.40 | 0.699 | | | GDP | -0.01 | -0.77 | 0.459 | | | GDP | 0.001 | 0.09 | 0.932 |
| | | Rice % | 0.08 | 6.17 | < 0.001 | | | Rice % | -0.01 | -1.45 | 0.171 | | | Rice % | 0.06 | 4.81 | < 0.001 |

Note: Analyses are hierarchical linear models nested in provinces. Non-Mandarin provinces are classified as Guangxi, Guangdong, Hunan, Zhejiang, Jiangxi, Hainan, Anhui, Fujian, and Shanghai (based on the *Language Atlas of China*). Provincial GDP per capita statistics are from 2014 in RMB. *Cognition and discourse.

**Which Explains More: Rice, Development, or the Urban-Rural Divide?**

Next, we compared historical rice farming to two factors that researchers invoke far more often to explain cultural differences: modernization[9] and urbanization[15]. Modernization and urban-rural differences are easy to see in China. In contrast, historical patterns of farming are not obvious to people. Thus, it would be logical to predict that modernization and urbanization should more strongly shape differences in China than rice.

To compare rice, modernization, and urbanization, we analyzed provinces' rice, GDP per capita, and the percentage of urban residents. We ran hierarchical linear models with users nested in provinces for each LIWC word category. We put one predictor and one word category in each model, then took the $R$ squared. Then we averaged the $R$ squared across all word categories.

First, we did an analysis for the words we hypothesized would be related to rice. Then we compared the predictive power ($R$ squared) across *all* 79 LIWC 2015 categories, regardless of whether we predicted rice-wheat differences for that particular category. We did this because, we might bias the analysis in favor of rice if we limit it to only the variables we *predicted* would be related to rice. Using all categories avoids the problem of cherry-picking the categories we hypothesized were related to rice. In short, analyzing across all word categories is a more conservative test of the explanatory power of rice.

We anticipated that the analysis would explain a small portion of the variance for two reasons. (a) We analyzed across all variables, rather than variables theoretically linked to the predictors. (b) Because calculating group-level variance is technically complex, we calculated individual-level variance, which tends to be far smaller. Thus, we interpret these numbers in relative terms, rather than in absolute numbers.

Across all word categories, modernization explained 0.1% of the variation; urbanization explained 0.2% of the variation; and rice explained 0.3% of the variation (Figure 3). Although small, historical farming explained more variation in how people use words in China than urbanization and modernization.

**Figure 3**

*Rice Explains More Variance in Word Use Than GDP and Urbanization Among Hypothesized Linguistic Categories (Top) and All LIWC Word Categories (Bottom)*

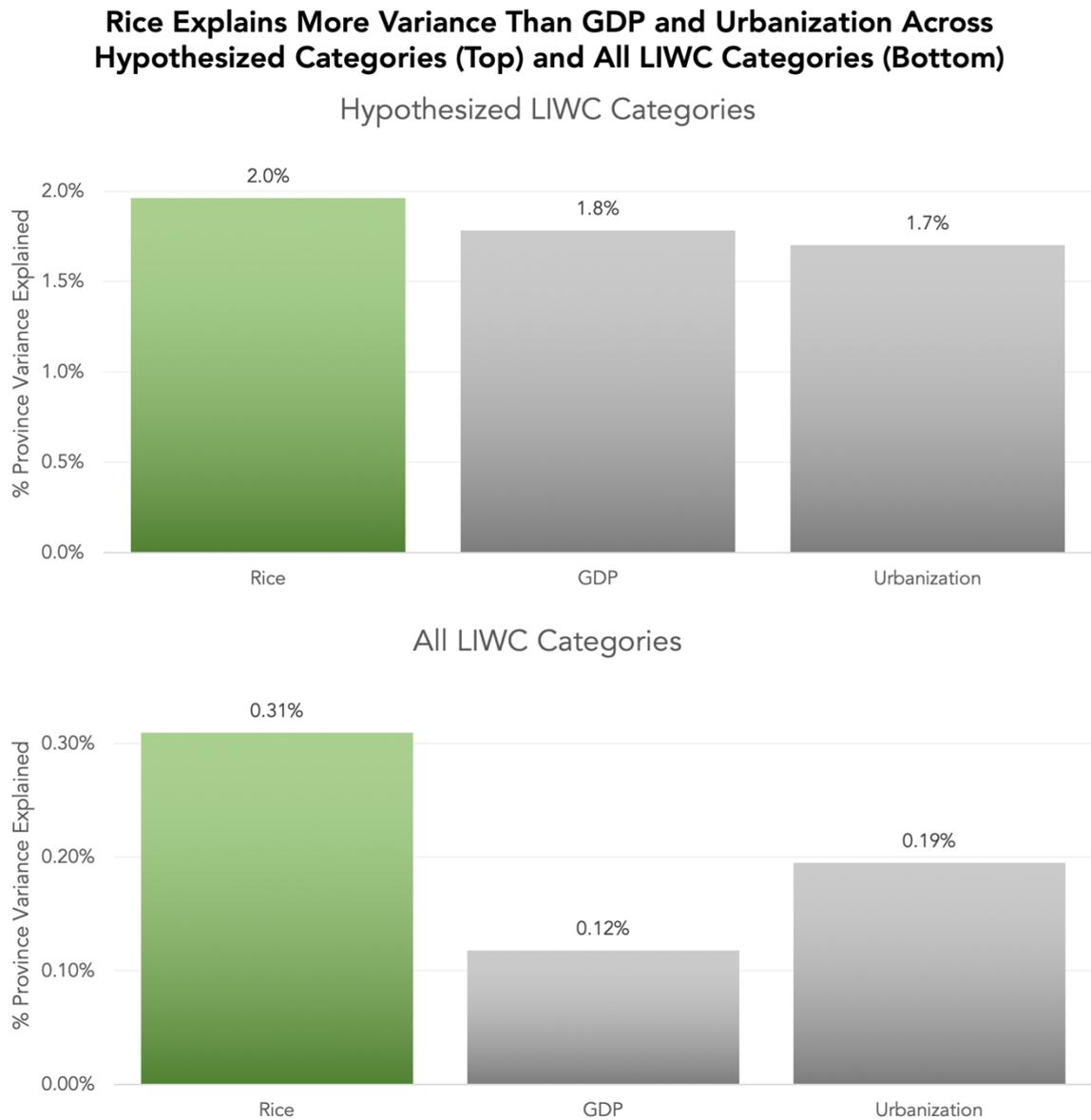

note: The top graph reports the average percentage of variance across provinces ($R$ squared) that each variable explains for our hypothesized LIWC word categories. However, testing just the word categories we hypothesized would be linked to rice could bias the results in favor of rice. To address this potential for bias, the bottom graph tests across all LIWC word categories, regardless of whether we hypothesized about them or not. Results from both analyses show that rice explains more regional variation in China than economic modernization (GDP per capita) and urbanization (measured by percent urban population).

**Are Rice-Wheat Differences Disappearing in Modern China?**

For much of its history, China has been an agricultural society. But in 1997, China dropped below 50% employment in agriculture for the first time in recorded history[71]. Now that most people in China are no longer farming, it raises the question of whether rice-wheat differences are slowly disappearing over time. If it is primarily *the act of farming itself* that causes rice culture, then differences should be disappearing in China's modern cities. But if rice culture is embedded more deeply in the culture—through socialization styles, institutions, schools, cultural products—then rice culture may still be just as alive as ever in cities like Shanghai.

We tested this idea using a simple method: we used China's city tier system to divide prefectures into China's modern mega cities (first-tier cities) to smaller cities and rural areas (second and third tier). For simplicity, we call this "urban" versus "rural," although any urban-rural division oversimplifies the continuum from farmland to mega city. We then compared the absolute size of rice-wheat differences in urban versus rural areas (using standardized regression coefficients). We compared this for the 11 word categories that showed consistent rice-wheat differences in the main analysis.

Out of 11 variables that showed consistent rice-wheat differences in the main analysis, eight were actually *larger* in urban areas than rural areas (Figure 4). Three were larger in rural areas. Two exceptions were thematically related: humans and universalism. This could suggest that living in large cities pushes people to consider others in broader terms, rather than in the narrower terms of specific relationships. This fits with the fact that people in big cities cross paths with hundreds or even thousands of people every day on sidewalks, subways, and shopping malls.

Yet these two exceptions aside, the bulk of the evidence suggested that rice-wheat differences are persisting, even in China's modern cities. If anything, the differences tended to be larger in China's most modern cities. This contradicts the idea that historical legacies are disappearing in the modern world. But it fits with the speculation that modernization in China is, in part, bringing about an embrace of traditional cultural patterns[72]. At the very least, the results strongly suggest that people do not need to have threshed rice to inherit rice culture.

**Figure 4**

*Most Rice-Wheat Differences Are Larger in Urban Areas (Blue) Than Rural Areas (Red)*

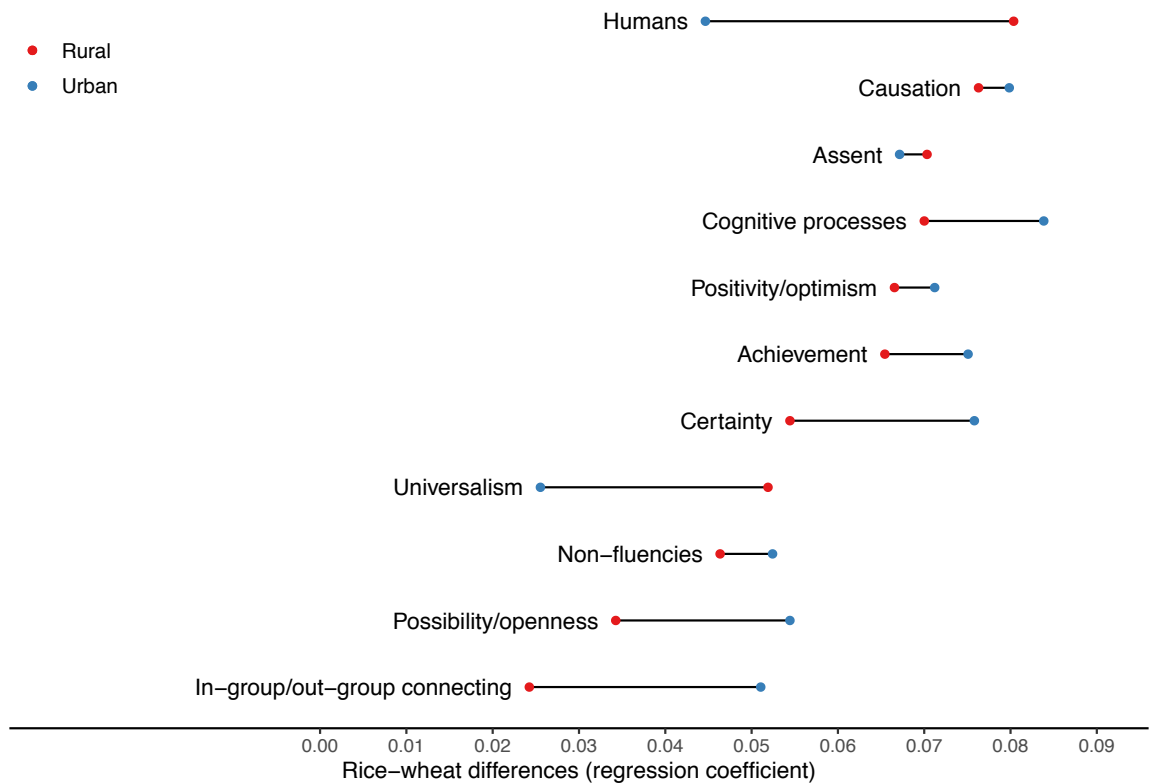

Note: The dots represent the effect size (regression coefficient) for rice-wheat differences between urban areas (blue) or between rural areas (red). Regression coefficients are absolute values (ignoring positive or negative). Differences were larger in urban areas for eight out of 11 variables. This contradicts the idea that rice-wheat differences are disappearing in modern environments.

**Rice Predicts Similar Word Patterns in Japan**

Beyond statistically controlling for potential confounding variables, another way to test the robustness of an effect is to test it again in a different country with a different language. Testing in Japan allows us to more strongly rule out potential confounds in China, such as the influence of herding populations in northern China. Japan also presents an interesting test of whether rice culture persists into the modern era, since Japan has been modernized for longer. On GDP per capita, Japan caught up to Western Europe in the 1970s, surpassed it in the 1980s, and has been wealthier for the last 40 years[67].

**Japan Twitter Data**

We analyzed 266 million terms from approximately 8.8 million tweets posted between 2014 and 2019, collected using the Twitter API with a bounding box around the country of Japan. We further geolocated them to the 47 Japanese prefectures by looking for matches between each prefecture name and the Twitter location field obtained from user profile information based on prior work[73].

We then tested whether the 10 LIWC word categories related to rice in China would also be related to rice farming in Japan. Since LIWC is not available in Japanese, we used methods from prior work to translate the word lists from Chinese and English into Japanese[73,74]. This method has been validated in prior studies on social media across cultures[73] and in spoken language in Japan[74].

**Rice in Japan**

We measured rice using data on the percentage of rice (稲) per planted area in Japan's 47 prefectures[75]. Japan's prefectures are the size of many US counties. The prefecture data goes back to 1975. It is reasonable to expect smaller differences in Japan than China. This is because prefecture rice varies in China from 0% to 95%. In Japan, rice varies from 18% to 90%[b].

Does 1975 data reflect historical farming patterns? Data from larger regions going back to 1950 shows less than 3% variation from 1950 to 2000. The 1975 data is also highly correlated with environmental suitability for rice $\beta = 0.78$, $P < 0.001$. In other words, this data seems to reflect deep historical patterns of rice farming that are largely determined by environmental conditions (Supplemental Section 16 provides more detail).

**Control Variables**

As in China, we controlled for economic development (GDP per capita), the percentage of urban residents, and education (Table S6).

**Results**

Rice predicted word use in Japan similar to China (Table 9). People in rice areas used language similarly to rice areas in China in the three major categories of cognition, promotion orientation, and self/groups. People in rice areas used fewer words related to cognitive processes, large human groups, and self words. As in China, the results in Japan were similar after controlling for regional differences in education (Table S6).

However, there were a few differences from the results in China. Rice areas in China used more assent words and non-fluencies, but rice areas in Japan did not show these patterns. Like China, rice-farming prefectures in Japan had fewer achievement words, although the difference was marginally significant ($\beta = -0.02$, $P = 0.055$, $r_{pref} = -0.15$). Finally, the difference between "I" and "we" was larger in Japan than in China. In Japan, prefectures with the most rice used significantly less "I" ($\beta = -0.02$, $P = 0.025$, $r_{pref} = -0.31$) but the same amount of "we" as prefectures with less rice ($\beta = -0.001$, $P = 0.950$, $r_{pref} = -$

---

[b] The Tokyo urban area has less rice. However, given the Tokyo area's high environmental suitability for rice, it likely farmed more rice historically. The outlying island of Okinawa also farms less rice.

0.11). As in China, rice prefectures used fewer pronouns in general ($\beta$ = -0.02, $P$ = 0.029, $r_{prov}$ = -0.28). This provides more evidence for the idea that collectivistic areas drop pronouns more frequently, even within the same language.

**Table 9**

*Prefectures' History of Rice Farming Predicts Word Use on Weibo in Japan*

| | Word Category | | $\beta$ | $t$ | $P$ | | Word Category | | $\beta$ | $t$ | $P$ |
|---|---|---|---|---|---|---|---|---|---|---|---|
| Cognition and Discourse | Cognitive Processes | GDP | -0.005 | -0.30 | 0.769 | Cognition and Discourse | Possibility/ Openness | GDP | 0.002 | 0.14 | 0.891 |
| | | % Urban | 0.03 | 3.82 | < 0.001 | | | % Urban | 0.03 | 3.83 | 0.001 |
| | | Rice % | -0.04 | -3.78 | < 0.001 | | | Rice % | -0.02 | -3.06 | 0.007 |
| | Causation | GDP | 0.04 | 2.42 | 0.027 | | Assent | GDP | -0.05 | -2.93 | 0.009 |
| | | % Urban | 0.01 | 0.82 | 0.418 | | | % Urban | -0.005 | -0.54 | 0.595 |
| | | Rice % | -0.02 | -2.40 | 0.022 | | | Rice % | -0.02 | -2.48 | 0.019 |
| | Certainty | GDP | -0.01 | -0.62 | 0.542 | | Non-Fluencies | GDP | -0.03 | -1.53 | 0.140 |
| | | % Urban | 0.01 | 1.60 | 0.118 | | | % Urban | 0.004 | 0.39 | 0.699 |
| | | Rice % | -0.03 | -2.93 | 0.006 | | | Rice % | -0.004 | -0.38 | 0.703 |
| Self and Groups | Humans | GDP | 0.003 | 0.19 | 0.850 | Promotion Orientation | Achievement | GDP | 0.03 | 1.40 | 0.178 |
| | | % Urban | 0.01 | 0.79 | 0.435 | | | % Urban | 0.02 | 2.38 | 0.023 |
| | | Rice % | -0.02 | -2.12 | 0.042 | | | Rice % | -0.02 | -1.99 | 0.055 |
| | I | GDP | -0.03 | -1.56 | 0.136 | | We | GDP | -0.01 | -0.43 | 0.669 |
| | | % Urban | 0.01 | 0.91 | 0.370 | | | % Urban | 0.01 | 1.03 | 0.307 |
| | | Rice % | -0.02 | -2.34 | 0.025 | | | Rice % | -0.001 | -0.06 | 0.950 |
| | Geogr. Units: 47 | | Located Users: 58,797 | | | Distinct Terms: 957,356 | | | Total Terms: 266,193,619 | | |

Note: Analyses are hierarchical linear models nested in prefectures. Prefecture rice is the percentage of rice planted area.

**Discussion**

Analysis of more than a billion terms on Weibo revealed different patterns of language use in historically rice-farming parts of China and Japan. These differences were independent of factors like age, gender, economic development, and urbanization. The large sample size gives high confidence that the differences are reliable. What's more, the rice-wheat differences replicated for provinces and prefectures.

The rice-wheat differences fell into three broad categories:

1. **Thought Style:** People in wheat-farming areas used more words related to thought and logic, whereas people in rice areas seemed to emphasize agreement (assent words) and express more caution toward knowledge (such as fewer certainty words and more non-fluencies).
2. **Achievement and Promotion:** People in rice-farming areas seemed to be more focused on prevention, whereas people in wheat-farming areas expressed more optimism and public striving.
3. **Broad versus Tight Relationships:** People in wheat areas used more words linked to humanity and universalism than people in rice areas, which fits with the idea that rice farming involved tight social ties with people close to them[17,64].

Rice-wheat differences remained when analyzing people from Mandarin-speaking areas and when excluding Cantonese-speaking areas. This matters because Cantonese is arguably the most different written form of Chinese. However, it would be illuminating if future studies can look for regional differences in *spoken* Chinese, where dialects have more room for expression.

**Limitations**

One major limitation of our data is that we do not know what people mean if we simply count the words they use. There is a telling illustration of this problem in a study of preachers' sermons. A team of researchers analyzed different categories of morality words that preachers used in more conservative and more liberal churches in the US[76]. Conservatives tend to emphasize the role of authority more than liberals, so the researchers expected that liberal churches would use fewer words about authority. Yet authority words were highly common sermons from Unitarians, one of the most liberal churches.

However, when the researchers used human judgment to understand *how* preachers were using these words, it was clear that Unitarians were encouraging their followers to reject authority. Ministers were exhorting their members to *question* authority. Unitarians cared deeply about authority, but they cared mostly about the harms of authority.

With our Weibo data, we can only say that people in rice-farming regions are using certain words more or less. We cannot know for sure whether they are endorsing these ideas or rejecting these ideas. This fits with a common disclaimer on Twitter: "Retweets do not equal endorsement."

Another limitation of Weibo data is censorship. Researchers have documented censorship of particular words on Weibo[77]. In response, Weibo users sometimes use similar-sounding words to evade censorship[77]. This could distort the words people use and create noise in the word categories. However, we did not analyze political topics, which should limit the effect of censorship. Also, the fact that many patterns replicated in Japan's Twitter data suggests that the relationships between rice and word use are reliable.

**Farming Legacies Alive in Modern China**

The fact that this study took place on Weibo is an important contextual detail. Weibo might be the *last* place to expect to find cultural differences rooted in farming legacies since Weibo is modern. Its users are younger and more educated than the broader population[78]. Yet even among this modern, smartphone generation, word use maps onto ancient patterns of rice farming.

It may be surprising that historical rice farming explained more variation in word use than urban-rural differences and economic development. When comparing word use across all LIWC categories—not just the categories theoretically linked to rice—rice explained more variation than urbanization and modernization. This is unexpected because economic development and the urban-rural divide are much more popular explanations of regional differences in China[11–13].

It is logical to think that rice farming could affect rice farmers themselves, but these differences should be fading over time, as China modernizes and fewer people farm. Yet on average, differences were stronger among the people most removed from farming—people in China's largest cities like Shanghai and Beijing. If verified in other samples, this finding raises the intriguing possibility that modernization is magnifying historical cultural legacies.

**Methods**

**Data**

Our materials are public messages posted on Weibo. Since Weibo does not provide streaming API tools to obtain random samples over time (as Twitter does), we used the Weibo API to query for all status updates from a given user, where users were crawled using a breadth-first search strategy beginning with random users. In total, we obtained about 29 million posts posted in 2014 and 2015 from 859,054 users. Based on prior work, we used the self-reported location from user's profile information [73]. Three prior studies have found that this method of locating participants is accurate and reliable on Twitter and Weibo[79–81].

We are limited in our analyses to 2014 and 2015 because Weibo makes it extremely difficult to collect data. Unlike Twitter, Weibo does not provide an API that makes access easy. One alternative is to scrape data, but Weibo requires registration to scrape, and registration requires a Chinese phone number. Thus, longitudinal data covering more years is not feasible.

**LIWC**

We started by testing the LIWC Chinese categories. We converted each post into a vector by counting tokens in user posts that match tokens in the LIWC dictionary. We then summed these token counts on the user level and normalize by the number of words posted by each user.

**Linguistic Features**

We automatically extracted the relative frequency of single words, LIWC, and theory-driven features across posts of all users. First, all posts of users were segmented into words using THULAC[82]. Then, we removed all words used by less than 1% of users to remove uncommonly used words. The features were extracted from all posts and aggregated to each user in the dataset to obtain language profiles for every user.

**Theory-Driven Features**

We also created new categories and sub-categories based on cultural psychological theories that were not represented in the LIWC categories: alone, fashion/trends, in-group/out-group, justice, and universalism words. We report all results in the supplemental materials and focus on categories that revealed meaningful regional differences in the main text.

**User Demographics**

Many Weibo users report their demographic information such as age, gender, and location. Of the entire set of users in our dataset, we obtained age for 13,776 users, gender for 558,264 users and location (city and province) for 564,139 users. Of these, we analyzed data from users with at least 1,000 words so that we have sufficient language from each user for analysis.

We analyzed these variables but acknowledge that this is a limited set of individual demographics. For example, we do not have demographics such as political beliefs, social status, and education. Although we do not have these variables at the individual level, we estimate regional differences in variables like education and wealth.

One anonymous reviewer asked about removing organization accounts from the analysis. This is done automatically in the analyses that include gender and age. That is because organizations like companies, news media, and non-profits usually do not put a gender and age on their account.

The assumption is that organizations are less representative of regional culture. However, this assumption may be flawed. Both individuals and organizations are a part of cultures. And as collective bodies, organizations may be even stronger reflections of local culture. Regardless, our main analyses include all accounts, and our analyses with gender and age are mostly individual accounts.

For location, we were able to map users in our dataset to 29 provinces and 421 prefectures (*shi* 市, often translated as "city" but more akin to prefectures or US counties). This retained 249,361 users with province-level information and 80,825 users with prefecture-level information.

**Regional Variables**

Table S1 reports the sources and measures for each regional variable. This table also describes the theoretical rationale for each variable.

**Rice.** We used the percentage of cultivated land devoted to paddy rice for the earliest year we could find. For provinces, this was from the 1996 *China Statistical Yearbook*. For most prefectures, the earliest yearbooks were from 2002. This measure excludes dryland rice, which does not use irrigation networks to provide standing water and thus requires less labor and coordination. How well does 1996 data represent historical rice farming? We compared this data to 1914 rice data available for 22 provinces[83]. The two were correlated highly $r(22) = 0.95$, $P < 0.001$, suggesting that the 1996 data reflects traditional regional differences in rice farming.

We use the term "wheat" to describe areas with little rice farming. This makes sense because provinces in China that don't farm rice tend to farm wheat $r(28) = 0.69$, $P < 0.001$. However, we use "wheat" as a catchall term that describes dryland crops in general. There is diversity in China's wheat region. Around the time of Confucius, millet was more common in northern China[17,84]. Later, corn arrived from the Americas; nowadays, a handful of provinces grow more corn than wheat. Future studies can dive more deeply into whether there are meaningful differences between different dryland crops like wheat, corn, and potatoes.

**GDP and modernization.** We used province and prefecture GDP data from 2014 as a measure of modernization[85]. Because there is evidence of a lag between modernization and cultural change[10], we also tested 1996 GDP per capita (Table S14).

We also tested several alternative measures of modernization because researchers have argued that GDP is not always the best measure of modernization[8]. Because local leaders in China receive promotions based in part on GDP statistics, some researchers have expressed suspicion about GDP numbers[86]. For that reason, we collected data on internet installation rates from the China Internet Network Information Center. This data is less politically sensitive and less likely to be faked. Supplemental Section 8 describes more alternative measures.

**Urbanization.** We used the percentage of urban residents per province as a measure of urbanization. We analyzed data from 2016 to represent modern urbanization and data from 2000 to represent time-lagged urbanization.

**Japan Data**

We measured economic development in Japan using prefecture GDP per capita from 2010. We used urbanization statistics on the ratio of urban population per prefecture from 2005 (the latest year from the statistical department report). We controlled for differences in regional education using the percentage of college graduates per prefecture in 2010.

Unfortunately, data on users' gender, age, and education were unavailable for Twitter users in Japan. However, the data from China suggests that this is not a major concern, because the China data showed highly similar results for rice controlling for user characteristics or not.

**Statistical Analysis**

We analyzed the data using hierarchical linear models with the LMER function in the lme4 package in the program R. We grouped observations at the province level for province-level rice analyses and at the prefecture level for prefecture-level rice analyses.

We isolated the patterns in users' language using dictionary-based open vocabulary features by building multi-level models predicting linguistic feature usage across individuals using controls from both individual users and users' locations. We used Benjamini-Hochberg $P$-correction and $P < 0.05$ to identify meaningful correlations. To test the hypothesis, we estimated a series of regression models which are variations of the following:

$$\text{LinguisticCategory}_i = \alpha_i + \beta_1 \text{Control}_{user} + \beta_2 \text{Control}_{region} + \Phi + \varepsilon$$

For each feature dimension in every linguistic category $i$, we built a regression model with user-level controls (age and gender) and regional control variables along with $\Phi$, which is the random effect (one of 421 counties or one of 29 provinces), and $\varepsilon$ is the error term.

All analyses control for gender. We then added in provincial or prefecture-level control variables, such as GDP. Because age is available for only a small sub-sample, we first analyzed whether age is a meaningful predictor for each word category. Then we analyze age controls in depth for the categories that age predicts over $r > 0.10$.

**Figure S1**

*Rice-Wheat Differences Are Largely Independent of Dialect*

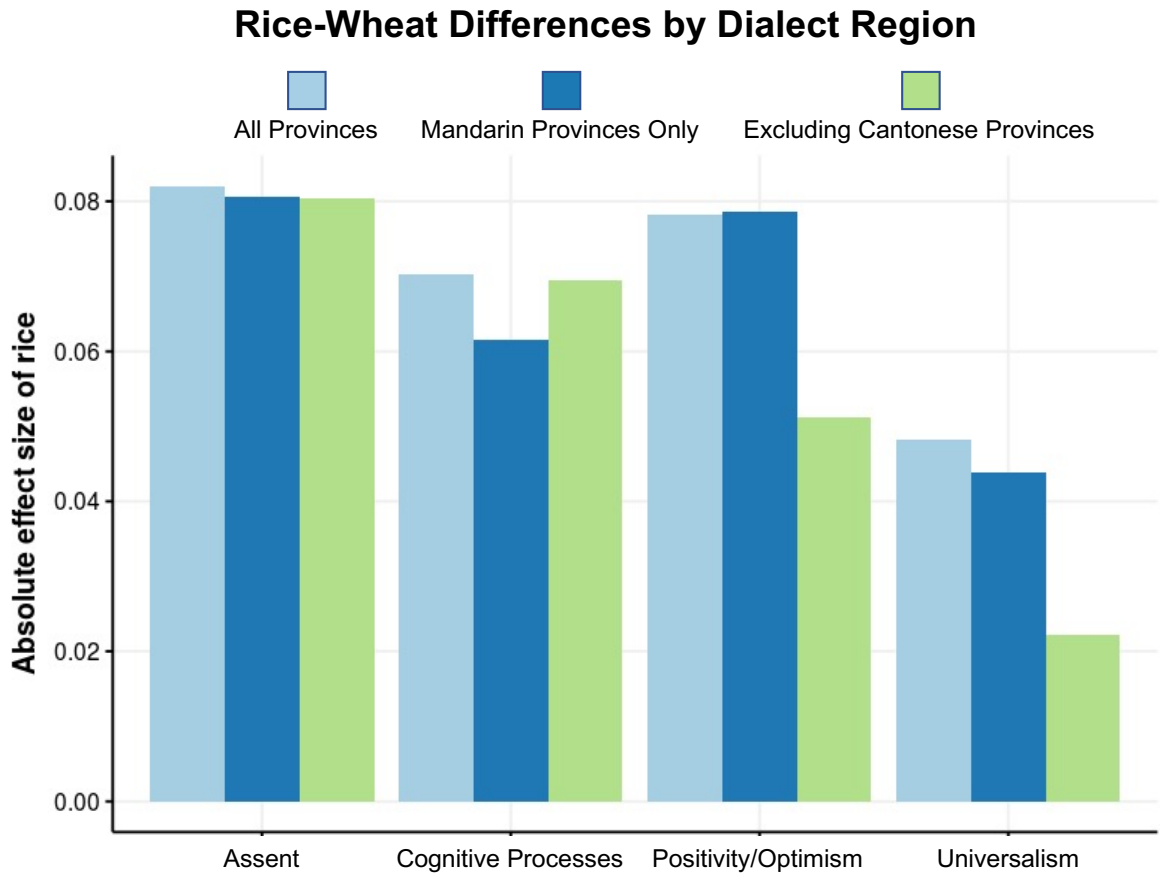

*Note:* To test whether differences in dialects are confounding the results, we ran analyses testing rice-wheat differences among all provinces (light blue) and then only for provinces in the Mandarin dialect group (dark blue). Green bars exclude Cantonese, which is the dialect outside of Mandarin with arguably the most developed writing system. Effect sizes are individual-level standardized regression coefficients. Effect sizes tend to be larger at the group level.

**Table S1A**
*Variables Tested, Sources, and Rationale in China*

| Variable | Measure | Source | Rationale |
|---|---|---|---|
| % Rice: Province & Prefecture | Paddy fields area/total cultivated area | *China Statistical Yearbook* 1996; Provincial Statistical Yearbooks (mostly 2002) | Paddy rice required more work and coordination to build and operate irrigation systems. |
| Modern GDP: Province & Prefecture | 2014 GDP per capita | *China Statistical Yearbook* 2015; *China City Statistical Yearbook* 2015 | Modernization theory argues that wealth and modernization cause cultures to become more individualistic. |
| Historical GDP | 1995 GDP per capita | *China Stat. Yearbook*, 1996 | Studies have found a lag between economic growth and cultural change. |
| % Urban | Urban residents/total population, 2016 | *China Pop. & Employment Statistical Yearbook*, 2017 | We use urbanization to test the difference between urban and rural areas. Urbanization is also an alternative indicator of modernization. |
| % Cultivated Land | Hectares of cultivated land/province land | *China Statistical Yearbook*, 1996 | This measures the density of farming in general, pulling apart general farming and rice farming in particular. |
| Environmental Rice Suitability: Province & Prefecture | Environmental suitability for high-input rainfed rice | UN Global Agro-Ecological Zones Database | Environmental suitability for rice (regardless of whether people actually farm rice there) helps address reverse causality—the possibility that areas that were collectivistic to begin with chose to farm rice. |
| % College Graduates | College graduates per school-age population, 1990, 2015 | *China Statistical Yearbook*, 1991, 2016 | Researchers have argued that education is an important vehicle of modernization. Education may also influence use of cognitive words. We test modern and historical statistics to account for lag in cultural change. |
| % Herding Cultures | Traditionally herding ethnicities/total pop. | *China Population Statistical Yearbook*, 2002 | Research has found that herding cultures tend to be more individualistic than nearby farming cultures. |
| Average Temperature | Average high, low in January and July | Zuzu Che Weather Records | Some researchers have argued that hotter areas are more collectivistic. Temperature is correlated with disease prevalence. |
| Climatic Demands | Sum dev. from 22C in avg. highs/lows July, Jan | Zuzu Che Weather Records | Climatic demands theory argues that cultures in harsher climates should be more collectivistic. |
| % Service Industry | People employed in service jobs/employed people, 2010, 1995 | *China Statistical Yearbook, 2011, 1996* | Some researchers have argued that service sector development is a better indicator of modernization than GDP. We test historical statistics because there is evidence for a lag between economic development and cultural change. |
| % Private Industry | People employed in priv. industry/employed people, 2010, 1995 | *China Statistical Yearbook, 2011, 1996* | In China, the shift from the state-controlled economy to the private sector is an indicator of economic modernization beyond GDP. We test historical statistics because there is evidence for a lag between development and cultural change. |
| Internet Penetration | Internet users/province population | China Internet Development Report, 2008 | Researchers have found evidence that GDP statistics in China are sometimes manipulated. Internet installation rates are less politically sensitive, presenting an alternative indicator of modernization. |
| Percent Han | People of Han ethnicity/province population | *China Population Statistical Yearbook, 2002* | The percent Han could be interpreted as a measure of ethnic homogeneity (lack of diversity) or as a proxy for Confucian heritage. |
| Pathogen Prevalence | Human-transmitted infectious diseases rates | Chen et al., 1990; Statistical Yearbooks | Pathogen prevalence theory argues that environments with higher rates of communicable disease tend to be more collectivistic. |

**Table S1B**

*Variables Tested, Sources, and Rationale in Japan*

| Variable | Measure | Source | Rationale |
| --- | --- | --- | --- |
| % Prefecture Rice | Rice planted area/total planted area, 1975 | Statistics and Information Department (2010). 7-14 Planted area and production of rice by prefecture (1985--2009). | Rice required more work and coordination to build and operate irrigation systems. This made rice areas more interdependent. |
| GDP: Prefecture | GDP per capita, 2005 | Cabinet Office. (2013). The 2013 annual report on prefectural-level accounts. Tokyo, Japan: Cabinet Office of Japan. | Modernization theory argues that wealth and modernization cause cultures to become more individualistic. |
| % Urban | Ratio of urban inhabitants, 2005 | Statistics Bureau, Ministry of Internal Affairs and Communications. (2016). Population, Population Density, Population of Densely Inhabited Districts and Area by Prefecture, All *Shi* 市 and All *Gun* 郡 (1898-2005). | We use urbanization to test the difference between urban and rural areas. Urbanization is also an alternative indicator of modernization. |
| Education | Percent of college graduates, 2010 | Statistics Bureau, Ministry of Internal Affairs and Communications. (2016). Social indicators by prefecture. | Education is an important vehicle of modernization. Education may also make people more likely to cognitive process words. |

**Table S2A**
*Rice-Wheat Differences Hold Controlling for Temperature (Province Level)*

| | Word Category | | β | t | P | | Word Category | | β | t | P | | Word Category | | β | t | P |
|---|---|---|---|---|---|---|---|---|---|---|---|---|---|---|---|---|---|
| Cognition and Discourse | Cognitive Processes | Female | 0.08 | 40.98 | < 0.001 | Self and Groups | Universalism | Female | -0.08 | -36.86 | < 0.001 | Promotion Orientation and Emotion | Positivity/ Optimism | Female | 0.001 | 0.67 | 0.504 |
| | | GDP | 0.01 | 0.28 | 0.780 | | | GDP | -0.01 | -0.22 | 0.828 | | | GDP | -0.02 | -0.37 | 0.712 |
| | | % Urban | -0.05 | -0.93 | 0.365 | | | % Urban | 0.01 | 0.22 | 0.827 | | | % Urban | -0.01 | -0.24 | 0.815 |
| | | % Herding | 0.01 | 0.96 | 0.345 | | | % Herding | 0.01 | 1.71 | 0.100 | | | % Herding | 0.01 | 1.17 | 0.252 |
| | | % Cult. Land | -0.01 | -0.57 | 0.574 | | | % Cult. Land | -0.01 | -1.00 | 0.330 | | | % Cult. Land | -0.01 | -0.44 | 0.666 |
| | | Temperature | 0.0004 | 0.03 | 0.978 | | | Temperature | -0.01 | -0.96 | 0.344 | | | Temperature | 0.01 | 0.67 | 0.509 |
| | | Rice % | -0.07 | -3.53 | 0.002 | | | Rice % | -0.04 | -2.63 | 0.016 | | | Rice % | -0.09 | -3.78 | 0.001 |
| | Causation | Female | -0.12 | -58.55 | < 0.001 | | Humans | Female | 0.11 | 55.51 | < 0.001 | | Achievement | Female | -0.21 | -109.01 | < 0.001 |
| | | GDP | 0.01 | 0.21 | 0.838 | | | GDP | -0.02 | -0.29 | 0.772 | | | GDP | 0.002 | 0.04 | 0.971 |
| | | % Urban | -0.01 | -0.11 | 0.916 | | | % Urban | -0.03 | -0.45 | 0.655 | | | % Urban | 0.01 | 0.19 | 0.849 |
| | | % Herding | 0.01 | 1.35 | 0.189 | | | % Herding | 0.02 | 1.91 | 0.069 | | | % Herding | 0.01 | 0.96 | 0.345 |
| | | % Cult. Land | 0.004 | 0.21 | 0.833 | | | % Cult. Land | -0.01 | -0.41 | 0.686 | | | % Cult. Land | 0.005 | 0.25 | 0.806 |
| | | Temperature | 0.005 | 0.31 | 0.759 | | | Temperature | 0.004 | 0.25 | 0.805 | | | Temperature | 0.001 | 0.06 | 0.953 |
| | | Rice % | -0.07 | -3.58 | 0.002 | | | Rice % | -0.08 | -3.28 | 0.003 | | | Rice % | -0.07 | -3.04 | 0.006 |
| | Certainty | Female | 0.09 | 45.00 | < 0.001 | | In/Outgroup: Connecting | Female | 0.0002 | 0.12 | 0.906 | | Fashion and Trends | Female | 0.04 | 15.56 | < 0.001 |
| | | GDP | -0.01 | -0.19 | 0.850 | | | GDP | 0.01 | 0.40 | 0.695 | | | GDP | -0.004 | -0.12 | 0.908 |
| | | % Urban | -0.02 | -0.60 | 0.556 | | | % Urban | -0.05 | -1.57 | 0.133 | | | % Urban | 0.01 | 0.37 | 0.712 |
| | | % Herding | 0.01 | 0.71 | 0.485 | | | % Herding | 0.02 | 2.39 | 0.025 | | | % Herding | 0.003 | 0.44 | 0.662 |
| | | % Cult. Land | -0.02 | -1.37 | 0.186 | | | % Cult. Land | -0.02 | -1.60 | 0.127 | | | % Cult. Land | 0.01 | 0.45 | 0.657 |
| | | Temperature | -0.01 | -0.60 | 0.551 | | | Temperature | 0.0004 | 0.04 | 0.966 | | | Temperature | -0.004 | -0.40 | 0.691 |
| | | Rice % | -0.05 | -3.38 | 0.003 | | | Rice % | -0.02 | -1.61 | 0.126 | | | Rice % | 0.01 | 0.74 | 0.466 |
| | Possibility/ Openness | Female | 0.08 | 37.82 | < 0.001 | | In/Outgroup: Dividing | Female | -0.01 | -2.91 | 0.004 | | Affect | Female | 0.26 | 137.08 | < 0.001 |
| | | GDP | 0.01 | 0.36 | 0.722 | | | GDP | -0.11 | -2.17 | 0.043 | | | GDP | -0.03 | -0.65 | 0.523 |
| | | % Urban | -0.04 | -0.87 | 0.396 | | | % Urban | 0.14 | 2.53 | 0.022 | | | % Urban | -0.03 | -0.70 | 0.490 |
| | | % Herding | 0.004 | 0.47 | 0.646 | | | % Herding | 0.01 | 0.47 | 0.641 | | | % Herding | 0.02 | 1.67 | 0.108 |
| | | % Cult. Land | -0.01 | -0.89 | 0.383 | | | % Cult. Land | 0.01 | 0.31 | 0.762 | | | % Cult. Land | -0.02 | -1.52 | 0.144 |
| | | Temperature | 0.004 | 0.31 | 0.761 | | | Temperature | 0.03 | 1.44 | 0.158 | | | Temperature | 0.01 | 0.40 | 0.693 |
| | | Rice % | -0.04 | -2.66 | 0.015 | | | Rice % | -0.02 | -0.73 | 0.477 | | | Rice % | -0.02 | -0.92 | 0.369 |
| | Assent | Female | 0.18 | 93.52 | < 0.001 | | Self | Female | -0.05 | -17.61 | < 0.001 | Cog. & Dis. | Non-Fluencies | Female | 0.18 | 90.23 | < 0.001 |
| | | GDP | -0.003 | -0.05 | 0.959 | | | GDP | -0.01 | -0.31 | 0.762 | | | GDP | -0.01 | -0.15 | 0.880 |
| | | % Urban | -0.02 | -0.30 | 0.770 | | | % Urban | 0.03 | 0.68 | 0.503 | | | % Urban | -0.01 | -0.13 | 0.897 |
| | | % Herding | -0.01 | -0.70 | 0.490 | | | % Herding | -0.01 | -0.64 | 0.528 | | | % Herding | -0.005 | -0.41 | 0.683 |
| | | % Cult. Land | -0.01 | -0.38 | 0.707 | | | % Cult. Land | -0.01 | -1.12 | 0.277 | | | % Cult. Land | -0.01 | -0.38 | 0.706 |
| | | Temperature | 0.005 | 0.27 | 0.786 | | | Temperature | 0.01 | 0.47 | 0.642 | | | Temperature | -0.003 | -0.16 | 0.876 |
| | | Rice % | 0.08 | 3.32 | 0.003 | | | Rice % | -0.02 | -1.14 | 0.269 | | | Rice % | 0.06 | 2.87 | 0.009 |

Note: This table tests yearly average temperature per province. Table S2B tests temperature at the prefecture level. Yearly average temperature (C) is the average highs and lows of January and July for the capital city of each province. GDP per capita statistics are from 2012 in RMB. The main text tables use 2014 GDP at the request of a reviewer but the two correlate *r* = 0.99.

**Table S2B**

*Differences Between Rice and Wheat Prefectures Are Independent from Temperature*

| | Word Category | | β | t | P | | Word Category | | β | t | P | | Word Category | | β | t | P |
|---|---|---|---|---|---|---|---|---|---|---|---|---|---|---|---|---|---|
| Cognition and Discourse | Cognitive Processes | Female | 0.08 | 24.19 | < 0.001 | Self and Groups | Universalism | Female | -0.09 | -23.28 | < 0.001 | Promotion Orientation and Emotion | Positivity/ Optimism | Female | -0.01 | -1.69 | 0.091 |
| | | GDP | -0.03 | -4.84 | < 0.001 | | | GDP | 0.002 | 0.29 | 0.772 | | | GDP | -0.04 | -5.51 | < 0.001 |
| | | Temp. | -0.001 | -0.10 | 0.920 | | | Temp. | -0.03 | -3.63 | < 0.001 | | | Temp. | 0.004 | 0.45 | 0.654 |
| | | Rice % | -0.08 | -7.48 | < 0.001 | | | Rice % | -0.02 | -3.01 | 0.003 | | | Rice % | -0.08 | -7.14 | < 0.001 |
| | Causation | Female | -0.13 | -37.55 | < 0.001 | | Humans | Female | 0.12 | 33.91 | < 0.001 | | Achievement | Female | -0.23 | -67.35 | < 0.001 |
| | | GDP | 0.01 | 1.63 | 0.105 | | | GDP | -0.04 | -6.27 | < 0.001 | | | GDP | 0.02 | 2.48 | 0.014 |
| | | Temp. | -0.02 | -2.59 | 0.010 | | | Temp. | -0.02 | -2.20 | 0.029 | | | Temp. | -0.02 | -2.21 | 0.028 |
| | | Rice % | -0.06 | -6.49 | < 0.001 | | | Rice % | -0.06 | -5.48 | < 0.001 | | | Rice % | -0.06 | -5.65 | < 0.001 |
| | Certainty | Female | 0.10 | 27.61 | < 0.001 | | In/Outgroup: Connecting | Female | -0.0005 | -0.13 | 0.900 | | Fashion and Trends | Female | 0.02 | 3.87 | < 0.001 |
| | | GDP | -0.04 | -5.72 | < 0.001 | | | GDP | -0.04 | -8.30 | < 0.001 | | | GDP | 0.02 | 2.85 | 0.005 |
| | | Temp. | -0.01 | -1.15 | 0.252 | | | Temp. | -0.01 | -0.85 | 0.397 | | | Temp. | -0.01 | -1.42 | 0.157 |
| | | Rice % | -0.06 | -6.03 | < 0.001 | | | Rice % | -0.03 | -3.63 | < 0.001 | | | Rice % | 0.02 | 2.11 | 0.036 |
| | Possibility/ Openness | Female | 0.08 | 22.96 | < 0.001 | | In/Outgroup: Dividing | Female | -0.03 | -3.40 | < 0.001 | | Affect | Female | 0.28 | 82.18 | < 0.001 |
| | | GDP | -0.03 | -4.76 | < 0.001 | | | GDP | 0.01 | 0.88 | 0.382 | | | GDP | -0.06 | -8.75 | < 0.001 |
| | | Temp. | 0.02 | 2.26 | 0.025 | | | Temp. | 0.02 | 1.48 | 0.141 | | | Temp. | 0.02 | 1.96 | 0.050 |
| | | Rice % | -0.06 | -6.24 | < 0.001 | | | Rice % | -0.02 | -1.35 | 0.181 | | | Rice % | -0.04 | -3.23 | 0.001 |
| | Assent | Female | 0.21 | 59.97 | < 0.001 | | Self | Female | -0.05 | -10.49 | < 0.001 | Cog. & Dis. | Non-Fluencies | Female | 0.20 | 57.61 | < 0.001 |
| | | GDP | -0.02 | -2.48 | 0.014 | | | GDP | 0.01 | 1.75 | 0.081 | | | GDP | -0.01 | -2.30 | 0.022 |
| | | Temp. | 0.04 | 4.79 | < 0.001 | | | Temp. | 0.02 | 1.75 | 0.082 | | | Temp. | 0.03 | 3.87 | < 0.001 |
| | | Rice % | 0.04 | 3.59 | < 0.001 | | | Rice % | -0.02 | -2.08 | 0.040 | | | Rice % | 0.02 | 2.48 | 0.014 |

Note: Analyses are hierarchical linear models nested in prefectures. Yearly average temperature (Celsius) is the average highs and lows of January and July for the capital city of each province. GDP statistics are from 2012 in RMB. The analyses in the main text use 2014 GDP at the request of a reviewer, but 2012 and 2014 prefecture GDP correlate $r = 0.98$, so the difference is negligible.

**Table S3A**

*Rice-Wheat Differences Are Robust to Differences in Education*

| | Word Category | | β | t | P | | Word Category | | β | t | P | | Word Category | | β | t | P |
|---|---|---|---|---|---|---|---|---|---|---|---|---|---|---|---|---|---|
| | | Female | 0.08 | 40.98 | < 0.001 | | | Female | -0.08 | -36.85 | < 0.001 | | | Female | 0.001 | 0.66 | 0.510 |
| | | % College | -0.06 | -1.72 | 0.101 | | | % College | 0.01 | 0.40 | 0.694 | | | % College | -0.07 | -1.71 | 0.101 |
| | Cognitive | % Urban | 0.01 | 0.26 | 0.798 | | | % Urban | -0.01 | -0.29 | 0.774 | | Positivity/ | % Urban | 0.01 | 0.37 | 0.714 |
| | Processes | % Herding | 0.01 | 1.09 | 0.286 | | Universalism | % Herding | 0.02 | 2.39 | 0.024 | | Optimism | % Herding | 0.01 | 0.83 | 0.412 |
| | | % Cult. Land | -0.02 | -1.04 | 0.312 | | | % Cult. Land | -0.01 | -0.77 | 0.452 | | | % Cult. Land | -0.02 | -1.14 | 0.265 |
| | | Rice % | -0.08 | -5.07 | < 0.001 | | | Rice % | -0.04 | -3.26 | 0.004 | | | Rice % | -0.09 | -4.81 | < 0.001 |
| | | Female | -0.12 | -58.55 | < 0.001 | | | Female | 0.11 | 55.50 | < 0.001 | | | Female | -0.21 | -109.01 | < 0.001 |
| | | % College | 0.004 | 0.11 | 0.913 | | | % College | -0.06 | -1.51 | 0.145 | | | % College | 0.001 | 0.04 | 0.971 |
| | Causation | % Urban | 0.001 | 0.05 | 0.962 | | Humans | % Urban | -0.001 | -0.04 | 0.970 | | Achievement | % Urban | 0.01 | 0.36 | 0.723 |
| | | % Herding | 0.01 | 1.54 | 0.138 | | | % Herding | 0.02 | 1.99 | 0.058 | | | % Herding | 0.01 | 1.43 | 0.165 |
| | | % Cult. Land | 0.004 | 0.24 | 0.809 | | | % Cult. Land | -0.02 | -0.96 | 0.347 | | | % Cult. Land | 0.005 | 0.27 | 0.792 |
| Cognition and Discourse | | Rice % | -0.07 | -3.83 | < 0.001 | Self and Groups | | Rice % | -0.09 | -4.40 | < 0.001 | Promotion Orientation and Emotion | | Rice % | -0.07 | -3.40 | 0.002 |
| | | Female | 0.09 | 45.01 | < 0.001 | | | Female | 0.0003 | 0.12 | 0.904 | | | Female | 0.04 | 15.56 | < 0.001 |
| | | % College | -0.03 | -1.20 | 0.247 | | | % College | 0.02 | 0.98 | 0.340 | | | % College | -0.02 | -1.08 | 0.295 |
| | Certainty | % Urban | -0.01 | -0.32 | 0.755 | | In/Outgroup: | % Urban | -0.06 | -2.92 | 0.009 | | Fashion and | % Urban | 0.03 | 1.41 | 0.172 |
| | | % Herding | 0.01 | 0.94 | 0.357 | | Connecting | % Herding | 0.02 | 3.03 | 0.005 | | Trends | % Herding | 0.003 | 0.57 | 0.575 |
| | | % Cult. Land | -0.02 | -1.70 | 0.104 | | | % Cult. Land | -0.01 | -1.29 | 0.213 | | | % Cult. Land | 0.002 | 0.21 | 0.835 |
| | | Rice % | -0.07 | -4.78 | < 0.001 | | | Rice % | -0.02 | -1.38 | 0.181 | | | Rice % | 0.001 | 0.10 | 0.922 |
| | | Female | 0.08 | 37.81 | < 0.001 | | | Female | -0.01 | -2.93 | 0.003 | | | Female | 0.26 | 137.09 | < 0.001 |
| | | % College | -0.06 | -2.50 | 0.021 | | | % College | 0.04 | 0.89 | 0.386 | | | % College | -0.08 | -2.78 | 0.012 |
| | Possibility/ | % Urban | 0.02 | 1.10 | 0.282 | | In/Outgroup: | % Urban | -0.004 | -0.12 | 0.908 | | Affect | % Urban | -0.01 | -0.26 | 0.801 |
| | Openness | % Herding | 0.002 | 0.29 | 0.775 | | Dividing | % Herding | -0.01 | -0.49 | 0.626 | | | % Herding | 0.01 | 1.52 | 0.139 |
| | | % Cult. Land | -0.02 | -1.70 | 0.105 | | | % Cult. Land | -0.004 | -0.21 | 0.837 | | | % Cult. Land | -0.04 | -2.82 | 0.010 |
| | | Rice % | -0.06 | -4.39 | < 0.001 | | | Rice % | 0.02 | 0.80 | 0.434 | | | Rice % | -0.03 | -2.12 | 0.045 |
| | | Female | 0.18 | 93.53 | < 0.001 | | | Female | -0.05 | -17.62 | < 0.001 | | | Female | 0.18 | 90.23 | < 0.001 |
| | | % College | 0.03 | 0.81 | 0.429 | | | % College | -0.02 | -0.62 | 0.540 | | | % College | -0.01 | -0.38 | 0.709 |
| | Assent | % Urban | -0.04 | -1.30 | 0.206 | | Self | % Urban | 0.03 | 1.20 | 0.242 | Cog. & Dis. | Non-Fluencies | % Urban | -0.005 | -0.15 | 0.884 |
| | | % Herding | -0.01 | -0.88 | 0.386 | | | % Herding | -0.01 | -1.13 | 0.268 | | | % Herding | -0.005 | -0.51 | 0.615 |
| | | % Cult. Land | -0.005 | -0.24 | 0.812 | | | % Cult. Land | -0.02 | -1.50 | 0.150 | | | % Cult. Land | -0.01 | -0.51 | 0.618 |
| | | Rice % | 0.09 | 4.39 | < 0.001 | | | Rice % | -0.02 | -1.25 | 0.226 | | | Rice % | 0.06 | 3.01 | 0.006 |

Note: Analyses are hierarchical linear models nested in provinces. Percent college graduates are college graduates per school-age population in the year 2015, derived from *China Statistical Yearbook* 2016. We use this as a proxy for differences in education levels for different provinces.

**Table S3B**

*Rice-Wheat Differences Are Robust to Differences in Time-Lagged Education*

| | Word Category | | β | t | P | | Word Category | | β | t | P | | Word Category | | β | t | P |
|---|---|---|---|---|---|---|---|---|---|---|---|---|---|---|---|---|---|
| | | Female | 0.08 | 40.99 | < 0.001 | | | Female | -0.08 | -36.86 | < 0.001 | | | Female | 0.001 | 0.66 | 0.509 |
| | | % Coll. 1990 | -0.08 | -2.89 | 0.009 | | | % Coll. 1990 | 0.01 | 0.26 | 0.796 | | | % Coll. 1990 | -0.09 | -2.70 | 0.013 |
| | Cognitive Processes | % Urban | 0.02 | 0.96 | 0.347 | | Universalism | % Urban | -0.003 | -0.16 | 0.872 | | Positivity/ Optimism | % Urban | 0.03 | 0.96 | 0.347 |
| | | % Herding | 0.01 | 1.33 | 0.193 | | | % Herding | 0.02 | 2.36 | 0.026 | | | % Herding | 0.01 | 1.04 | 0.307 |
| | | % Cult. Land | -0.02 | -1.14 | 0.268 | | | % Cult. Land | -0.01 | -0.85 | 0.404 | | | % Cult. Land | -0.02 | -1.21 | 0.240 |
| | | Rice % | -0.09 | -5.99 | < 0.001 | | | Rice % | -0.05 | -3.47 | 0.002 | | | Rice % | -0.10 | -5.55 | < 0.001 |
| | | Female | -0.12 | -58.56 | < 0.001 | | | Female | 0.11 | 55.50 | < 0.001 | | | Female | -0.21 | -109.01 | < 0.001 |
| | | % Coll. 1990 | -0.05 | -1.53 | 0.140 | | | % Coll. 1990 | -0.08 | -2.29 | 0.031 | | | % Coll. 1990 | -0.04 | -1.24 | 0.229 |
| | Causation | % Urban | 0.04 | 1.44 | 0.165 | | Humans | % Urban | 0.01 | 0.38 | 0.707 | | Achievement | % Urban | 0.04 | 1.46 | 0.157 |
| | | % Herding | 0.01 | 1.58 | 0.127 | | | % Herding | 0.02 | 2.24 | 0.034 | | | % Herding | 0.01 | 1.45 | 0.158 |
| | | % Cult. Land | -0.001 | -0.04 | 0.969 | | | % Cult. Land | -0.02 | -0.98 | 0.337 | | | % Cult. Land | 0.001 | 0.05 | 0.957 |
| Cognition and Discourse | | Rice % | -0.08 | -4.82 | < 0.001 | | | Rice % | -0.09 | -4.93 | < 0.001 | Promotion Orientation and Emotion | | Rice % | -0.08 | -4.15 | < 0.001 |
| | | Female | 0.09 | 45.01 | < 0.001 | | | Female | 0.0003 | 0.12 | 0.905 | | | Female | 0.04 | 15.57 | < 0.001 |
| | | % Coll. 1990 | -0.05 | -2.05 | 0.053 | | | % Coll. 1990 | 0.01 | 0.32 | 0.756 | | | % Coll. 1990 | -0.02 | -0.77 | 0.451 |
| | Certainty | % Urban | 0.004 | 0.19 | 0.854 | Self and Groups | In/Outgroup: Connecting | % Urban | -0.05 | -2.45 | 0.023 | | Fashion and Trends | % Urban | 0.02 | 1.16 | 0.259 |
| | | % Herding | 0.01 | 1.07 | 0.294 | | | % Herding | 0.02 | 2.91 | 0.007 | | | % Herding | 0.004 | 0.66 | 0.515 |
| | | % Cult. Land | -0.02 | -1.85 | 0.079 | | | % Cult. Land | -0.02 | -1.53 | 0.142 | | | % Cult. Land | 0.004 | 0.38 | 0.708 |
| | | Rice % | -0.07 | -5.44 | < 0.001 | | | Rice % | -0.02 | -1.76 | 0.093 | | | Rice % | 0.004 | 0.32 | 0.753 |
| | | Female | 0.08 | 37.82 | < 0.001 | | | Female | -0.01 | -2.93 | 0.003 | | | Female | 0.26 | 137.09 | < 0.001 |
| | | % Coll. 1990 | -0.08 | -3.40 | 0.003 | | | % Coll. 1990 | 0.02 | 0.58 | 0.568 | | | % Coll. 1990 | -0.05 | -1.75 | 0.095 |
| | Possibility/ Openness | % Urban | 0.03 | 1.61 | 0.122 | | In/Outgroup: Dividing | % Urban | 0.01 | 0.18 | 0.857 | | Affect | % Urban | -0.03 | -1.16 | 0.259 |
| | | % Herding | 0.003 | 0.50 | 0.620 | | | % Herding | -0.01 | -0.56 | 0.576 | | | % Herding | 0.01 | 1.66 | 0.109 |
| | | % Cult. Land | -0.02 | -1.74 | 0.097 | | | % Cult. Land | -0.01 | -0.36 | 0.726 | | | % Cult. Land | -0.03 | -2.24 | 0.036 |
| | | Rice % | -0.06 | -5.00 | < 0.001 | | | Rice % | 0.01 | 0.62 | 0.540 | | | Rice % | -0.02 | -1.47 | 0.156 |
| | | Female | 0.18 | 93.53 | < 0.001 | | | Female | -0.05 | -17.62 | < 0.001 | | | Female | 0.18 | 90.23 | < 0.001 |
| | | % Coll. 1990 | 0.05 | 1.49 | 0.152 | | | % Coll. 1990 | -0.03 | -1.10 | 0.286 | | | % Coll. 1990 | 0.01 | 0.39 | 0.702 |
| | Assent | % Urban | -0.06 | -1.92 | 0.068 | | Self | % Urban | 0.03 | 1.65 | 0.112 | Cog. & Dis. | Non-Fluencies | % Urban | -0.02 | -0.84 | 0.412 |
| | | % Herding | -0.01 | -0.98 | 0.335 | | | % Herding | -0.01 | -1.13 | 0.267 | | | % Herding | -0.004 | -0.46 | 0.649 |
| | | % Cult. Land | -0.004 | -0.24 | 0.812 | | | % Cult. Land | -0.02 | -1.60 | 0.125 | | | % Cult. Land | -0.01 | -0.34 | 0.736 |
| | | Rice % | 0.09 | 4.89 | < 0.001 | | | Rice % | -0.02 | -1.46 | 0.160 | | | Rice % | 0.06 | 3.47 | 0.002 |

Note: Analyses are hierarchical linear models nested in provinces. Percent college graduates are college graduates per school-age population in the year 1990, derived from *China Statistical Yearbook* 1991. We use this as a measure of regional differences in education in recent history.

**Table S4A**

*Differences Between Rice and Wheat Provinces Are Robust to Users' Age*

| | Word Category | | $\beta$ | $t$ | $P$ | | Word Category | | $\beta$ | $t$ | $P$ | | Word Category | | $\beta$ | $t$ | $P$ |
|---|---|---|---|---|---|---|---|---|---|---|---|---|---|---|---|---|---|
| Cognition and Discourse | Cognitive Processes | Female | 0.09 | 8.33 | < 0.001 | Self and Groups | Universalism | Female | -0.08 | -6.66 | < 0.001 | Promotion Orientation and Emotion | Positivity/ Optimism | Female | 0.02 | 1.61 | 0.108 |
| | | Age | 0.03 | 2.44 | 0.015 | | | Age | 0.03 | 2.89 | 0.004 | | | Age | 0.04 | 3.59 | < 0.001 |
| | | GDP | -0.02 | -0.39 | 0.702 | | | GDP | -0.11 | -1.61 | 0.128 | | | GDP | -0.05 | -0.69 | 0.500 |
| | | % Urban | -0.03 | -0.45 | 0.665 | | | % Urban | 0.11 | 1.52 | 0.155 | | | % Urban | 0.01 | 0.08 | 0.934 |
| | | Rice % | -0.08 | -4.63 | 0.002 | | | Rice % | -0.07 | -3.17 | 0.008 | | | Rice % | -0.11 | -4.70 | < 0.001 |
| | Causation | Female | -0.11 | -9.75 | < 0.001 | | Humans | Female | 0.12 | 10.50 | < 0.001 | | Achievement | Female | -0.20 | -18.41 | < 0.001 |
| | | Age | 0.09 | 7.91 | < 0.001 | | | Age | 0.04 | 3.25 | 0.001 | | | Age | 0.09 | 8.56 | < 0.001 |
| | | GDP | 0.01 | 0.15 | 0.880 | | | GDP | -0.04 | -0.46 | 0.652 | | | GDP | -0.05 | -0.63 | 0.537 |
| | | % Urban | -0.01 | -0.14 | 0.890 | | | % Urban | -0.02 | -0.18 | 0.862 | | | % Urban | 0.05 | 0.62 | 0.544 |
| | | Rice % | -0.09 | -3.79 | 0.002 | | | Rice % | -0.10 | -3.53 | 0.003 | | | Rice % | -0.10 | -4.42 | < 0.001 |
| | Certainty | Female | 0.11 | 10.08 | < 0.001 | | In/Outgroup: Connecting | Female | 0.01 | 0.99 | 0.321 | | Fashion and Trends | Female | 0.04 | 2.96 | 0.003 |
| | | Age | -0.0004 | -0.04 | 0.969 | | | Age | 0.01 | 0.58 | 0.563 | | | Age | 0.01 | 0.58 | 0.564 |
| | | GDP | -0.07 | -1.08 | 0.302 | | | GDP | -0.17 | -3.04 | 0.022 | | | GDP | -0.08 | -1.36 | 0.190 |
| | | % Urban | 0.02 | 0.34 | 0.739 | | | % Urban | 0.11 | 1.89 | 0.120 | | | % Urban | 0.08 | 1.35 | 0.198 |
| | | Rice % | -0.07 | -3.83 | 0.003 | | | Rice % | -0.06 | -3.35 | 0.023 | | | Rice % | 0.02 | 1.42 | 0.177 |
| | Possibility/ Openness | Female | 0.08 | 7.50 | < 0.001 | | In/Outgroup: Dividing | Female | 0.01 | 0.33 | 0.743 | | Affect | Female | 0.26 | 24.05 | < 0.001 |
| | | Age | 0.02 | 1.86 | 0.063 | | | Age | -0.02 | -0.67 | 0.504 | | | Age | -0.04 | -4.01 | < 0.001 |
| | | GDP | 0.003 | 0.05 | 0.962 | | | GDP | -0.19 | -1.56 | 0.120 | | | GDP | -0.05 | -0.68 | 0.511 |
| | | % Urban | -0.05 | -0.65 | 0.532 | | | % Urban | 0.21 | 1.76 | 0.079 | | | % Urban | -0.02 | -0.27 | 0.796 |
| | | Rice % | -0.04 | -2.02 | 0.075 | | | Rice % | -0.02 | -0.72 | 0.469 | | | Rice % | -0.02 | -1.17 | 0.269 |
| | Assent | Female | 0.16 | 14.61 | < 0.001 | | Self | Female | -0.03 | -2.18 | 0.029 | Cog. & Dis. | Non-Fluencies | Female | 0.17 | 15.59 | < 0.001 |
| | | Age | -0.09 | -7.82 | < 0.001 | | | Age | 0.04 | 2.97 | 0.003 | | | Age | -0.08 | -7.13 | < 0.001 |
| | | GDP | 0.01 | 0.07 | 0.948 | | | GDP | 0.05 | 0.59 | 0.562 | | | GDP | -0.05 | -0.67 | 0.515 |
| | | % Urban | -0.04 | -0.39 | 0.704 | | | % Urban | -0.06 | -0.65 | 0.521 | | | % Urban | 0.02 | 0.29 | 0.779 |
| | | Rice % | 0.10 | 3.41 | 0.007 | | | Rice % | 0.02 | 0.96 | 0.353 | | | Rice % | 0.07 | 3.15 | 0.012 |

Note: Analyses are hierarchical linear models nested in provinces. Provincial GDP per capita statistics are from 2012 in RMB. The analyses in the main text use 2014 GDP at the request of a reviewer, but 2012 and 2014 GDP correlate $r$ = 0.99, so the difference is negligible.

**Table S4B**

*Differences Between Rice and Wheat Prefectures Are Robust to Users' Age*

| | Word Category | | β | t | P | | Word Category | | β | t | P | | Word Category | | β | t | P |
|---|---|---|---|---|---|---|---|---|---|---|---|---|---|---|---|---|---|
| Cognition and Discourse | Cognitive Processes | Female | 0.09 | 7.40 | < 0.001 | Self and Groups | Universalism | Female | -0.08 | -6.21 | < 0.001 | Promotion Orientation and Emotion | Positivity/ Optimism | Female | 0.02 | 1.24 | 0.216 |
| | | Age | 0.02 | 2.04 | 0.041 | | | Age | 0.03 | 2.58 | 0.010 | | | Age | 0.04 | 3.27 | 0.001 |
| | | GDP | -0.04 | -2.85 | 0.005 | | | GDP | 0.02 | 1.50 | 0.140 | | | GDP | -0.04 | -2.67 | 0.009 |
| | | Rice % | -0.08 | -5.35 | < 0.001 | | | Rice % | -0.04 | -2.65 | 0.012 | | | Rice % | -0.08 | -4.89 | < 0.001 |
| | Causation | Female | -0.12 | -9.99 | < 0.001 | | Humans | Female | 0.11 | 9.73 | < 0.001 | | Achievement | Female | -0.20 | -17.59 | < 0.001 |
| | | Age | 0.09 | 7.66 | < 0.001 | | | Age | 0.03 | 2.39 | 0.017 | | | Age | 0.09 | 8.00 | < 0.001 |
| | | GDP | 0.02 | 1.37 | 0.173 | | | GDP | -0.04 | -3.18 | 0.002 | | | GDP | 0.01 | 1.09 | 0.278 |
| | | Rice % | -0.09 | -6.38 | < 0.001 | | | Rice % | -0.08 | -5.41 | < 0.001 | | | Rice % | -0.09 | -5.77 | < 0.001 |
| | Certainty | Female | 0.11 | 9.14 | < 0.001 | | In/Outgroup: Connecting | Female | 0.01 | 0.89 | 0.373 | | Fashion and Trends | Female | 0.03 | 2.30 | 0.021 |
| | | Age | -0.004 | -0.32 | 0.747 | | | Age | 0.01 | 0.64 | 0.524 | | | Age | 0.01 | 0.58 | 0.563 |
| | | GDP | -0.05 | -3.31 | 0.001 | | | GDP | -0.07 | -5.39 | < 0.001 | | | GDP | 0.003 | 0.23 | 0.818 |
| | | Rice % | -0.07 | -4.66 | < 0.001 | | | Rice % | -0.05 | -4.05 | < 0.001 | | | Rice % | 0.03 | 2.28 | 0.022 |
| | Possibility/ Openness | Female | 0.09 | 7.22 | < 0.001 | | In/Outgroup: Dividing | Female | 0.01 | 0.31 | 0.760 | | Affect | Female | 0.25 | 22.21 | < 0.001 |
| | | Age | 0.02 | 1.48 | 0.139 | | | Age | -0.01 | -0.42 | 0.675 | | | Age | -0.04 | -3.82 | < 0.001 |
| | | GDP | -0.04 | -2.41 | 0.018 | | | GDP | 0.04 | 1.16 | 0.248 | | | GDP | -0.05 | -3.76 | < 0.001 |
| | | Rice % | -0.04 | -2.66 | 0.009 | | | Rice % | -0.004 | -0.11 | 0.910 | | | Rice % | -0.03 | -2.01 | 0.047 |
| | Assent | Female | 0.16 | 13.79 | < 0.001 | | Self | Female | -0.04 | -2.64 | 0.008 | Cog. & Dis. | Non-fluencies | Female | 0.17 | 14.44 | < 0.001 |
| | | Age | -0.09 | -7.39 | < 0.001 | | | Age | 0.04 | 2.67 | 0.008 | | | Age | -0.08 | -6.60 | < 0.001 |
| | | GDP | -0.02 | -1.64 | 0.105 | | | GDP | 0.01 | 0.32 | 0.749 | | | GDP | -0.03 | -2.67 | 0.009 |
| | | Rice % | 0.06 | 4.60 | < 0.001 | | | Rice % | 0.003 | 0.15 | 0.882 | | | Rice % | 0.06 | 4.67 | < 0.001 |

Note: Analyses are hierarchical linear models nested in prefectures. Age data is available for a sub-sample of Weibo users. GDP per capita is 2012 prefecture data in RMB. The analyses in the main text use 2014 GDP at the request of a reviewer, but 2012 and 2014 prefecture GDP correlate $r = 0.98$, so the difference is negligible.

**Table S5**

*Robustness to Dialect: Excluding Cantonese-Speaking Areas*

| | Word Category | | $\beta$ | $t$ | $P$ | | Word Category | | $\beta$ | $t$ | $P$ | | Word Category | | $\beta$ | $t$ | $P$ |
|---|---|---|---|---|---|---|---|---|---|---|---|---|---|---|---|---|---|
| rowspan cognition | Cognitive Processes | Female | 0.08 | 35.89 | < 0.001 | rowspan Self and Groups | Universalism | Female | -0.08 | -25.99 | < 0.001 | rowspan Promotion Orientation and Emotion | Positivity/ Optimism | Female | 0.01 | 2.65 | 0.008 |
| | | GDP | -0.04 | -2.74 | 0.012 | | | GDP | 0.03 | 1.86 | 0.100 | | | GDP | -0.01 | -0.45 | 0.665 |
| | | Rice % | -0.08 | -6.16 | < 0.001 | | | Rice % | -0.04 | -4.85 | < 0.001 | | | Rice % | -0.06 | -5.07 | < 0.001 |
| | Causation | Female | -0.13 | -57.48 | < 0.001 | | Humans | Female | 0.10 | 35.52 | < 0.001 | | Achievement | Female | -0.22 | -103.89 | < 0.001 |
| | | GDP | 0.004 | 0.36 | 0.724 | | | GDP | -0.001 | -0.05 | 0.961 | | | GDP | 0.01 | 0.76 | 0.455 |
| | | Rice %. | -0.08 | -6.31 | < 0.001 | | | Rice % | -0.07 | -6.42 | < 0.001 | | | Rice % | -0.08 | -6.08 | < 0.001 |
| | Certainty | Female | 0.09 | 39.40 | < 0.001 | | In/Outgroup: Connecting | Female | 0.001 | 0.23 | 0.815 | | Fashion and Trends | Female | 0.04 | 11.02 | < 0.001 |
| | | GDP | -0.04 | -4.27 | < 0.001 | | | GDP | -0.03 | -1.58 | 0.152 | | | GDP | 0.02 | 1.43 | 0.182 |
| | | Rice % | -0.06 | -6.31 | < 0.001 | | | Rice % | -0.04 | -4.14 | 0.002 | | | Rice % | 0.004 | 0.66 | 0.520 |
| | Possibility/ Openness | Female | 0.07 | 31.93 | < 0.001 | | In/Outgroup: Dividing | Female | -0.01 | -2.00 | 0.046 | | Affect | Female | 0.27 | 128.92 | < 0.001 |
| | | GDP | -0.03 | -2.49 | 0.021 | | | GDP | 0.02 | 1.08 | 0.319 | | | GDP | -0.07 | -4.96 | < 0.001 |
| | | Rice % | -0.05 | -4.29 | < 0.001 | | | Rice % | 0.01 | 1.12 | 0.281 | | | Rice % | -0.03 | -2.22 | 0.037 |
| | Assent | Female | 0.20 | 92.78 | < 0.001 | | Self | Female | -0.04 | -10.07 | < 0.001 | Cog./Dis. | Non-Fluencies | Female | 0.19 | 86.48 | < 0.001 |
| | | GDP | -0.02 | -1.46 | 0.160 | | | GDP | 0.02 | 1.50 | 0.165 | | | GDP | -0.01 | -1.10 | 0.281 |
| | | Rice % | 0.09 | 7.22 | < 0.001 | | | Rice % | 0.005 | 0.59 | 0.562 | | | Rice % | 0.06 | 5.84 | < 0.001 |

Note: Analyses are hierarchical linear models nested in provinces. Cantonese provinces are classified as Guangxi and Guangdong (based on the *Language Atlas of China*). Provincial GDP per capita statistics are from 2012 in RMB. The analyses in the main text use 2014 GDP at the request of a reviewer, but 2012 and 2014 GDP correlate $r = 0.99$, so the difference is negligible.

**Table S6**

*Japanese Prefectures' History of Rice Farming Predicts Word Use on Weibo: Controlling for Education*

<table>
<tr><th colspan="2">Word Category</th><th>ß</th><th>t</th><th>P</th><th colspan="2">Word Category</th><th>ß</th><th>t</th><th>P</th></tr>
<tr><td rowspan="12">Cognition and Discourse</td><td rowspan="4">Cognitive Processes</td><td>GDP</td><td>-0.01</td><td>-0.41</td><td>0.685</td><td rowspan="12">Cognition and Discourse</td><td rowspan="4">Possibility/ Openness</td><td>GDP</td><td>-0.001</td><td>-0.12</td><td>0.909</td></tr>
<tr><td>% Urban</td><td>0.03</td><td>2.40</td><td>0.020</td><td>% Urban</td><td>0.01</td><td>1.30</td><td>0.202</td></tr>
<tr><td>Education</td><td>0.01</td><td>0.92</td><td>0.364</td><td>Education</td><td>0.02</td><td>2.52</td><td>0.019</td></tr>
<tr><td>Rice %</td><td>-0.03</td><td>-3.48</td><td>0.001</td><td>Rice %</td><td>-0.02</td><td>-2.61</td><td>0.016</td></tr>
<tr><td rowspan="4">Causation</td><td>GDP</td><td>0.03</td><td>2.53</td><td>0.026</td><td rowspan="4">Assent</td><td>GDP</td><td>-0.05</td><td>-2.89</td><td>0.010</td></tr>
<tr><td>% Urban</td><td>-0.01</td><td>-1.16</td><td>0.254</td><td>% Urban</td><td>-0.01</td><td>-0.58</td><td>0.566</td></tr>
<tr><td>Education</td><td>0.03</td><td>2.99</td><td>0.005</td><td>Education</td><td>0.004</td><td>0.28</td><td>0.782</td></tr>
<tr><td>Rice %</td><td>-0.02</td><td>-2.07</td><td>0.047</td><td>Rice %</td><td>-0.02</td><td>-2.34</td><td>0.026</td></tr>
<tr><td rowspan="4">Certainty</td><td>GDP</td><td>-0.01</td><td>-0.66</td><td>0.519</td><td rowspan="4">Non-Fluencies</td><td>GDP</td><td>-0.03</td><td>-1.51</td><td>0.144</td></tr>
<tr><td>% Urban</td><td>0.01</td><td>1.09</td><td>0.283</td><td>% Urban</td><td>0.003</td><td>0.23</td><td>0.819</td></tr>
<tr><td>Education</td><td>0.003</td><td>0.23</td><td>0.820</td><td>Education</td><td>0.002</td><td>0.14</td><td>0.889</td></tr>
<tr><td>Rice %</td><td>-0.02</td><td>-2.77</td><td>0.009</td><td>Rice %</td><td>-0.004</td><td>-0.34</td><td>0.732</td></tr>
<tr><td rowspan="8">Self and Groups</td><td rowspan="4">Humans</td><td>GDP</td><td>0.004</td><td>0.21</td><td>0.835</td><td rowspan="8">Promotion Orientation</td><td rowspan="4">Achievement</td><td>GDP</td><td>0.02</td><td>1.22</td><td>0.241</td></tr>
<tr><td>% Urban</td><td>0.01</td><td>0.86</td><td>0.394</td><td>% Urban</td><td>0.01</td><td>0.50</td><td>0.618</td></tr>
<tr><td>Education</td><td>-0.01</td><td>-0.39</td><td>0.700</td><td>Education</td><td>0.03</td><td>2.49</td><td>0.018</td></tr>
<tr><td>Rice %</td><td>-0.02</td><td>-2.13</td><td>0.040</td><td>Rice %</td><td>-0.02</td><td>-1.61</td><td>0.117</td></tr>
<tr><td rowspan="4">I</td><td>GDP</td><td>-0.03</td><td>-1.53</td><td>0.143</td><td rowspan="4">We</td><td>GDP</td><td>-0.004</td><td>-0.25</td><td>0.803</td></tr>
<tr><td>% Urban</td><td>0.01</td><td>0.76</td><td>0.450</td><td>% Urban</td><td>0.02</td><td>1.90</td><td>0.062</td></tr>
<tr><td>Education</td><td>-0.001</td><td>-0.07</td><td>0.943</td><td>Education</td><td>-0.02</td><td>-1.73</td><td>0.091</td></tr>
<tr><td>Rice %</td><td>-0.02</td><td>-2.27</td><td>0.030</td><td>Rice %</td><td>-0.004</td><td>-0.43</td><td>0.670</td></tr>
</table>

Note: Analyses are hierarchical linear models nested in prefectures. Prefecture rice is the percentage of rice planted area.

**Table S7**

*Provinces' Environmental Suitability for Rice Predicts Word Use on Weibo*

<table>
<tr><th colspan="2">Word Category</th><th>ß</th><th>t</th><th>P</th><th colspan="2">Word Category</th><th>ß</th><th>t</th><th>P</th><th colspan="2">Word Category</th><th>ß</th><th>t</th><th>P</th></tr>
<tr><td rowspan="4">Cogniti...</td><td rowspan="4">Cognitive Processes</td><td>Female</td><td>0.08</td><td>40.97</td><td>< 0.001</td><td rowspan="4">Self and...</td><td rowspan="4">Universalism</td><td>Female</td><td>-0.08</td><td>-36.86</td><td>< 0.001</td><td rowspan="4">Promotion...</td><td rowspan="4">Positivity/ Optimism</td><td>Female</td><td>0.001</td><td>0.66</td><td>0.510</td></tr>
<tr><td>GDP</td><td>0.03</td><td>0.55</td><td>0.585</td><td>GDP</td><td>-0.01</td><td>-0.39</td><td>0.699</td><td>GDP</td><td>0.01</td><td>0.15</td><td>0.884</td></tr>
<tr><td>% Urban</td><td>-0.08</td><td>-1.66</td><td>0.112</td><td>% Urban</td><td>-0.002</td><td>-0.06</td><td>0.950</td><td>% Urban</td><td>-0.07</td><td>-1.21</td><td>0.239</td></tr>
<tr><td>% Herding</td><td>0.01</td><td>1.25</td><td>0.222</td><td>% Herding</td><td>0.02</td><td>2.29</td><td>0.030</td><td>% Herding</td><td>0.01</td><td>1.17</td><td>0.252</td></tr>
</table>

| Category | Variable | Est. | t | p | | Category | Variable | Est. | t | p | | Category | Variable | Est. | t | p |
|---|---|---|---|---|---|---|---|---|---|---|---|---|---|---|---|---|
| | % Cult. Land | 0.01 | 0.50 | 0.622 | | | % Cult. Land | 0.002 | 0.16 | 0.871 | | | % Cult. Land | 0.01 | 0.44 | 0.666 |
| | Rice Suit. | -0.05 | -3.09 | 0.005 | | | Rice Suit. | -0.04 | -3.48 | 0.002 | | | Rice Suit. | -0.05 | -3.04 | 0.006 |
| Causation | Female | -0.12 | -58.56 | < 0.001 | Humans | Female | 0.11 | 55.50 | < 0.001 | Achievement | Female | -0.21 | -109.02 | < 0.001 |
| | GDP | 0.03 | 0.57 | 0.572 | | GDP | 0.004 | 0.07 | 0.945 | | GDP | 0.01 | 0.28 | 0.779 |
| | % Urban | -0.05 | -0.95 | 0.355 | | % Urban | -0.07 | -1.24 | 0.229 | | % Urban | -0.03 | -0.48 | 0.636 |
| | % Herding | 0.01 | 1.39 | 0.177 | | % Herding | 0.03 | 2.15 | 0.042 | | % Herding | 0.01 | 1.36 | 0.184 |
| | % Cult. Land | 0.02 | 1.16 | 0.260 | | % Cult. Land | 0.01 | 0.51 | 0.613 | | % Cult. Land | 0.02 | 1.18 | 0.249 |
| | Rice Suit. | -0.05 | -3.31 | 0.003 | | Rice Suit. | -0.05 | -2.70 | 0.013 | | Rice Suit. | -0.05 | -3.22 | 0.004 |
| Certainty | Female | 0.09 | 45.00 | < 0.001 | In/Outgroup: Connecting | Female | 0.0002 | 0.11 | 0.912 | Fashion and Trends | Female | 0.04 | 15.57 | < 0.001 |
| | GDP | -0.01 | -0.15 | 0.885 | | GDP | 0.02 | 0.55 | 0.591 | | GDP | -0.01 | -0.43 | 0.674 |
| | % Urban | -0.04 | -1.11 | 0.281 | | % Urban | -0.06 | -2.09 | 0.050 | | % Urban | 0.02 | 0.78 | 0.441 |
| | % Herding | 0.01 | 1.21 | 0.235 | | % Herding | 0.02 | 2.63 | 0.014 | | % Herding | 0.001 | 0.10 | 0.921 |
| | % Cult. Land | -0.001 | -0.08 | 0.936 | | % Cult. Land | -0.01 | -1.26 | 0.221 | | % Cult. Land | 0.003 | 0.28 | 0.779 |
| | Rice Suit. | -0.04 | -3.38 | 0.003 | | Rice Suit. | -0.02 | -1.74 | 0.096 | | Rice Suit. | -0.004 | -0.46 | 0.652 |
| Possibility/ Openness | Female | 0.08 | 37.81 | < 0.001 | In/Outgroup: Dividing | Female | -0.01 | -2.93 | 0.003 | Affect | Female | 0.26 | 137.08 | < 0.001 |
| | GDP | 0.03 | 0.69 | 0.500 | | GDP | -0.09 | -1.82 | 0.082 | | GDP | -0.02 | -0.47 | 0.641 |
| | % Urban | -0.06 | -1.58 | 0.128 | | % Urban | 0.12 | 2.25 | 0.036 | | % Urban | -0.05 | -1.13 | 0.269 |
| | % Herding | 0.005 | 0.55 | 0.590 | | % Herding | -0.004 | -0.39 | 0.700 | | % Herding | 0.02 | 1.81 | 0.082 |
| | % Cult. Land | -0.003 | -0.21 | 0.834 | | % Cult. Land | -0.003 | -0.18 | 0.863 | | % Cult. Land | -0.02 | -1.53 | 0.139 |
| | Rice Suit. | -0.03 | -2.37 | 0.026 | | Rice Suit. | -0.001 | -0.07 | 0.945 | | Rice Suit. | -0.01 | -0.53 | 0.601 |
| Assent | Female | 0.18 | 93.53 | < 0.001 | Self | Female | -0.05 | -17.62 | < 0.001 | Cog. & Dis Non-Fluencies | Female | 0.18 | 90.24 | < 0.001 |
| | GDP | -0.01 | -0.19 | 0.853 | | GDP | -0.003 | -0.11 | 0.917 | | GDP | -0.02 | -0.43 | 0.670 |
| | % Urban | 0.02 | 0.33 | 0.744 | | % Urban | 0.01 | 0.40 | 0.694 | | % Urban | 0.03 | 0.54 | 0.592 |
| | % Herding | -0.01 | -0.91 | 0.373 | | % Herding | -0.01 | -1.02 | 0.317 | | % Herding | -0.004 | -0.42 | 0.678 |
| | % Cult. Land | -0.03 | -1.55 | 0.136 | | % Cult. Land | -0.01 | -1.13 | 0.270 | | % Cult. Land | -0.02 | -1.25 | 0.226 |
| | Rice Suit. | 0.07 | 3.97 | < 0.001 | | Rice Suit. | -0.01 | -1.06 | 0.298 | | Rice Suit. | 0.05 | 3.09 | 0.005 |

Note: Analyses are hierarchical linear models nested in provinces. Rice suitability is an instrumental variable that reduces the potential for reverse causality. Suitability is an index of environmental variables (such as rainfall) that determine where it is physically possible to grow paddy rice, regardless of whether people are farming rice there. We indexed herding using the percentage of traditionally herding ethnicities in each province, according to the 2000 Census. For example, Mongolian and Manchu ethnicities herded traditionally. The supplemental materials present the full list of herding groups. Provincial GDP per capita statistics are from 2012 in RMB. The analyses in the main text use 2014 GDP at the request of a reviewer, but 2012 and 2014 prefecture GDP correlate $r = 0.98$, so the difference is negligible.

**Table S8A**

*Rice-Wheat Differences Robust to Provinces' Climatic Demands*

| | Word Category | β | t | P | | Word Category | β | t | P | | Word Category | β | t | P |
|---|---|---|---|---|---|---|---|---|---|---|---|---|---|---|
| Cognition and Discourse | Cognitive Processes | | | | Self and Groups | Universalism | | | | Promotion Orientation and Emotion | Positivity/Optimism | | | |
| | Female | 0.08 | 40.97 | < 0.001 | | Female | -0.08 | -36.85 | < 0.001 | | Female | 0.001 | 0.66 | 0.507 |
| | GDP | 0.02 | 0.58 | 0.571 | | GDP | -0.02 | -0.56 | 0.579 | | GDP | -0.004 | -0.08 | 0.939 |
| | % Urban | -0.06 | -1.38 | 0.184 | | % Urban | 0.02 | 0.55 | 0.589 | | % Urban | -0.03 | -0.58 | 0.567 |
| | % Herding | 0.001 | 0.12 | 0.905 | | % Herding | 0.02 | 1.90 | 0.069 | | % Herding | 0.01 | 0.66 | 0.516 |
| | % Cult. Land | -0.01 | -0.95 | 0.356 | | % Cult. Land | -0.01 | -0.82 | 0.424 | | % Cult. Land | -0.01 | -0.67 | 0.511 |
| | Clim. Dem. | 0.02 | 1.68 | 0.106 | | Clim. Dem. | 0.003 | 0.29 | 0.771 | | Clim. Dem. | 0.01 | 0.34 | 0.734 |
| | Rice % | -0.06 | -3.27 | 0.004 | | Rice % | -0.05 | -3.28 | 0.004 | | Rice % | -0.07 | -3.46 | 0.002 |
| | Causation | | | | | Humans | | | | | Achievement | | | |
| | Female | -0.12 | -58.56 | < 0.001 | | Female | 0.11 | 55.50 | < 0.001 | | Female | -0.12 | -58.56 | < 0.001 |
| | GDP | 0.02 | 0.48 | 0.635 | | GDP | -0.001 | -0.03 | 0.977 | | GDP | 0.02 | 0.48 | 0.635 |
| | % Urban | -0.02 | -0.40 | 0.690 | | % Urban | -0.05 | -0.85 | 0.402 | | % Urban | -0.02 | -0.40 | 0.690 |
| | % Herding | 0.01 | 0.72 | 0.476 | | % Herding | 0.01 | 1.12 | 0.273 | | % Herding | 0.01 | 0.72 | 0.476 |
| | % Cult. Land | -0.001 | -0.03 | 0.974 | | % Cult. Land | -0.01 | -0.74 | 0.465 | | % Cult. Land | -0.001 | -0.03 | 0.974 |
| | Clim. Dem. | 0.01 | 0.91 | 0.373 | | Clim. Dem. | 0.02 | 1.36 | 0.187 | | Clim. Dem. | 0.01 | 0.91 | 0.373 |
| | Rice % | -0.06 | -3.28 | 0.004 | | Rice % | -0.06 | -2.85 | 0.009 | | Rice % | -0.06 | -3.28 | 0.004 |
| | Certainty | | | | | In/Outgroup: Connecting | | | | | Fashion and Trends | | | |
| | Female | -0.12 | -58.56 | < 0.001 | | Female | 0.0003 | 0.12 | 0.904 | | Female | 0.04 | 15.57 | < 0.001 |
| | GDP | 0.02 | 0.48 | 0.635 | | GDP | 0.01 | 0.39 | 0.698 | | GDP | -0.01 | -0.26 | 0.800 |
| | % Urban | -0.02 | -0.40 | 0.690 | | % Urban | -0.05 | -1.63 | 0.121 | | % Urban | 0.02 | 0.52 | 0.606 |
| | % Herding | 0.01 | 0.72 | 0.476 | | % Herding | 0.02 | 2.48 | 0.020 | | % Herding | 0.004 | 0.53 | 0.599 |
| | % Cult. Land | -0.001 | -0.03 | 0.974 | | % Cult. Land | -0.02 | -1.56 | 0.137 | | % Cult. Land | 0.01 | 0.53 | 0.601 |
| | Clim. Dem. | 0.01 | 0.91 | 0.373 | | Clim. Dem. | -0.004 | -0.36 | 0.726 | | Clim. Dem. | 0.001 | 0.14 | 0.889 |
| | Rice % | -0.06 | -3.28 | 0.004 | | Rice % | -0.02 | -1.90 | 0.074 | | Rice % | 0.01 | 0.62 | 0.543 |
| | Possibility/Openness | | | | | In/Outgroup: Dividing | | | | | Affect | | | |
| | Female | 0.08 | 37.81 | < 0.001 | | Female | -0.01 | -2.91 | 0.004 | | Female | 0.26 | 137.08 | < 0.001 |
| | GDP | 0.02 | 0.68 | 0.501 | | GDP | -0.10 | -2.16 | 0.042 | | GDP | -0.02 | -0.42 | 0.681 |
| | % Urban | -0.05 | -1.28 | 0.215 | | % Urban | 0.13 | 2.65 | 0.016 | | % Urban | -0.05 | -1.09 | 0.290 |
| | % Herding | -0.002 | -0.26 | 0.794 | | % Herding | 0.01 | 1.11 | 0.273 | | % Herding | 0.01 | 1.00 | 0.327 |
| | % Cult. Land | -0.03 | -1.84 | 0.080 | | % Cult. Land | 0.01 | 0.56 | 0.579 | | % Cult. Land | -0.03 | -1.84 | 0.080 |
| | Clim. Dem. | 0.01 | 1.12 | 0.274 | | Clim. Dem. | -0.04 | -2.54 | 0.016 | | Clim. Dem. | 0.01 | 0.90 | 0.379 |
| | Rice % | -0.03 | -2.24 | 0.037 | | Rice % | -0.02 | -1.07 | 0.299 | | Rice % | -0.01 | -0.32 | 0.755 |
| | Assent | | | | | Self | | | | Cog. & Dis. | Non-Fluencies | | | |
| | Female | 0.18 | 93.52 | < 0.001 | | Female | -0.05 | -17.60 | < 0.001 | | Female | 0.18 | 90.23 | < 0.001 |
| | GDP | -0.001 | -0.03 | 0.977 | | GDP | -0.01 | -0.26 | 0.798 | | GDP | -0.01 | -0.28 | 0.784 |
| | % Urban | -0.02 | -0.32 | 0.752 | | % Urban | 0.03 | 0.69 | 0.499 | | % Urban | -0.001 | -0.01 | 0.989 |
| | % Herding | -0.01 | -0.50 | 0.618 | | % Herding | -0.003 | -0.39 | 0.698 | | % Herding | -0.002 | -0.16 | 0.878 |
| | % Cult. Land | -0.01 | -0.34 | 0.734 | | % Cult. Land | -0.01 | -1.08 | 0.292 | | % Cult. Land | -0.01 | -0.28 | 0.780 |
| | Clim. Dem. | -0.01 | -0.60 | 0.555 | | Clim. Dem. | -0.01 | -0.83 | 0.412 | | Clim. Dem. | -0.01 | -0.34 | 0.733 |
| | Rice % | 0.08 | 3.50 | 0.002 | | Rice % | -0.02 | -1.34 | 0.196 | | Rice % | 0.06 | 2.84 | 0.010 |

Note: Analyses are hierarchical linear models nested in provinces. Provincial GDP per capita statistics are from 2012 in RMB. The analyses in the main text use 2014 GDP at the request of a reviewer, but 2012 and 2014 GDP correlate *r* = 0.99, so the difference is negligible. Climatic demands are calculated as the sum deviation from 22C in the average highs and lows of July and January. Climatic demands theory argues that cultures in harsher climates should be more collectivistic.

**Table S8B**

*Rice-Wheat Differences Robust to Prefectures' Climatic Demands*

| | Word Category | | ß | t | P | | Word Category | | ß | t | P | | Word Category | | ß | t | P |
|---|---|---|---|---|---|---|---|---|---|---|---|---|---|---|---|---|---|
| Cognition and Discourse | Cognitive Processes | Female | 0.08 | 24.19 | < 0.001 | Self and Groups | Universalism | Female | -0.09 | -23.20 | < 0.001 | Promotion Orientation and Emotion | Positivity/ Optimism | Female | -0.01 | -1.68 | 0.092 |
| | | GDP | -0.03 | -4.86 | < 0.001 | | | GDP | 0.002 | 0.46 | 0.644 | | | GDP | -0.04 | -5.17 | < 0.001 |
| | | Clim. Dem. | 0.005 | 0.58 | 0.562 | | | Clim. Dem. | 0.01 | 1.40 | 0.165 | | | Clim. Dem. | -0.01 | -1.14 | 0.256 |
| | | Rice % | -0.08 | -8.18 | < 0.001 | | | Rice % | -0.04 | -5.31 | < 0.001 | | | Rice % | -0.08 | -8.39 | < 0.001 |
| | Causation | Female | -0.13 | -37.53 | < 0.001 | | Humans | Female | 0.12 | 33.91 | < 0.001 | | Achievement | Female | -0.23 | -67.34 | < 0.001 |
| | | GDP | 0.01 | 1.45 | 0.150 | | | GDP | -0.05 | -6.50 | < 0.001 | | | GDP | 0.01 | 2.23 | 0.027 |
| | | Clim. Dem. | 0.01 | 1.94 | 0.054 | | | Clim. Dem. | 0.02 | 2.82 | 0.005 | | | Clim. Dem. | 0.01 | 1.93 | 0.054 |
| | | Rice % | -0.07 | -8.10 | < 0.001 | | | Rice % | -0.07 | -6.26 | < 0.001 | | | Rice % | -0.06 | -6.90 | < 0.001 |
| | Certainty | Female | 0.10 | 27.62 | < 0.001 | | In/Outgroup: Connecting | Female | -0.0004 | -0.11 | 0.910 | | Fashion and Trends | Female | 0.02 | 3.91 | < 0.001 |
| | | GDP | -0.04 | -5.64 | < 0.001 | | | GDP | -0.04 | -8.11 | < 0.001 | | | GDP | 0.02 | 2.87 | 0.005 |
| | | Clim. Dem. | 0.01 | 0.94 | 0.350 | | | Clim. Dem. | 0.003 | 0.46 | 0.643 | | | Clim. Dem. | 0.003 | 0.47 | 0.636 |
| | | Rice % | -0.06 | -7.09 | < 0.001 | | | Rice % | -0.03 | -4.42 | < 0.001 | | | Rice % | 0.01 | 1.54 | 0.125 |
| | Possibility/ Openness | Female | 0.08 | 22.94 | < 0.001 | | In/Outgroup: Dividing | Female | -0.03 | -3.36 | < 0.001 | | Affect | Female | 0.28 | 82.17 | < 0.001 |
| | | GDP | -0.03 | -4.60 | < 0.001 | | | GDP | 0.02 | 1.35 | 0.180 | | | GDP | -0.06 | -8.43 | < 0.001 |
| | | Clim. Dem. | -0.01 | -1.59 | 0.114 | | | Clim. Dem. | -0.04 | -2.77 | 0.006 | | | Clim. Dem. | -0.01 | -1.42 | 0.155 |
| | | Rice % | -0.05 | -6.19 | < 0.001 | | | Rice % | -0.03 | -2.00 | 0.048 | | | Rice % | -0.03 | -2.93 | 0.004 |
| | Assent | Female | 0.21 | 59.94 | < 0.001 | | Self | Female | -0.05 | -10.48 | < 0.001 | C. & D. | Non-Fluencies | Female | 0.20 | 57.56 | < 0.001 |
| | | GDP | -0.01 | -2.11 | 0.036 | | | GDP | 0.01 | 2.09 | 0.038 | | | GDP | -0.01 | -2.25 | 0.026 |
| | | Clim. Dem. | -0.03 | -3.62 | < 0.001 | | | Clim. Dem. | -0.02 | -2.34 | 0.020 | | | Clim. Dem. | -0.01 | -2.07 | 0.039 |
| | | Rice % | 0.05 | 5.55 | < 0.001 | | | Rice % | -0.02 | -2.37 | 0.019 | | | Rice % | 0.04 | 4.47 | < 0.001 |

Note: Analyses are hierarchical linear models nested in prefectures. Climatic demands are calculated as the sum deviation from 22C in the average highs and lows of July and January. Climatic demands theory argues that cultures in harsher climates should be more collectivistic. The analyses in the main text use 2014 GDP at the request of a reviewer, but 2012 and 2014 prefecture GDP correlate $r = 0.98$, so the difference is negligible.

**Table S9**

*Provincial Rice-Wheat Differences Are Robust to Climatic Demands Combined with GDP*

| | Word Category | | β | t | P | | Word Category | | β | t | P | | Word Category | | β | t | P |
|---|---|---|---|---|---|---|---|---|---|---|---|---|---|---|---|---|---|
| **Cognition and Discourse** | Cognitive Processes | Female | 0.08 | 40.97 | < 0.001 | **Self and Groups** | Universalism | Female | -0.08 | -36.84 | < 0.001 | **Promotion Orientation and Emotion** | Positivity/ Optimism | Female | 0.001 | 0.67 | 0.505 |
| | | GDP | -0.01 | -0.12 | 0.906 | | | GDP | 0.08 | 1.13 | 0.272 | | | GDP | 0.06 | 0.61 | 0.547 |
| | | % Urban | -0.06 | -1.21 | 0.242 | | | % Urban | 0.007 | 0.20 | 0.841 | | | % Urban | -0.04 | -0.73 | 0.473 |
| | | % Herding | 0.001 | 0.14 | 0.890 | | | % Herding | 0.02 | 1.92 | 0.066 | | | % Herding | 0.01 | 0.62 | 0.543 |
| | | % Cult. Land | -0.01 | -0.79 | 0.437 | | | % Cult. Land | -0.02 | -1.19 | 0.248 | | | % Cult. Land | -0.02 | -0.82 | 0.422 |
| | | Clim. Dem. | 0.01 | 0.36 | 0.721 | | | Clim. Dem. | 0.04 | 1.55 | 0.134 | | | Clim. Dem. | 0.03 | 0.81 | 0.426 |
| | | Clim. x GDP | 0.03 | 0.44 | 0.666 | | | Clim. x GDP | -0.10 | -1.59 | 0.125 | | | Clim. x GDP | -0.07 | -0.74 | 0.469 |
| | | Rice % | -0.05 | -2.84 | 0.010 | | | Rice % | -0.05 | -3.74 | 0.001 | | | Rice % | -0.08 | -3.49 | 0.002 |
| | Causation | Female | -0.12 | -58.55 | < 0.001 | | Humans | Female | 0.11 | 55.50 | < 0.001 | | Achievement | Female | -0.21 | -109.01 | < 0.001 |
| | | GDP | 0.08 | 0.96 | 0.348 | | | GDP | 0.07 | 0.66 | 0.516 | | | GDP | 0.14 | 1.42 | 0.171 |
| | | % Urban | -0.03 | -0.58 | 0.566 | | | % Urban | -0.06 | -1.00 | 0.327 | | | % Urban | -0.01 | -0.26 | 0.795 |
| | | % Herding | 0.01 | 0.68 | 0.501 | | | % Herding | 0.01 | 1.07 | 0.294 | | | % Herding | 0.01 | 0.86 | 0.398 |
| | | % Cult. Land | -0.004 | -0.23 | 0.821 | | | % Cult. Land | -0.02 | -0.90 | 0.378 | | | % Cult. Land | -0.004 | -0.21 | 0.833 |
| | | Clim. Dem. | 0.04 | 1.16 | 0.260 | | | Clim. Dem. | 0.05 | 1.30 | 0.207 | | | Clim. Dem. | 0.05 | 1.60 | 0.124 |
| | | Clim. x GDP | -0.07 | -0.83 | 0.414 | | | Clim. x GDP | -0.07 | -0.77 | 0.452 | | | Clim. x GDP | -0.14 | -1.53 | 0.139 |
| | | Rice % | -0.07 | -3.36 | 0.003 | | | Rice % | -0.07 | -2.93 | 0.008 | | | Rice % | -0.07 | -3.46 | 0.002 |
| | Certainty | Female | 0.09 | 44.99 | < 0.001 | | In/Outgroup: Connecting | Female | 0.0003 | 0.13 | 0.895 | | Fashion and Trends | Female | 0.04 | 15.57 | < 0.001 |
| | | GDP | -0.01 | -0.22 | 0.832 | | | GDP | 0.09 | 1.59 | 0.127 | | | GDP | 0.03 | 0.53 | 0.602 |
| | | % Urban | -0.03 | -0.72 | 0.478 | | | % Urban | -0.06 | -2.01 | 0.060 | | | % Urban | 0.01 | 0.35 | 0.732 |
| | | % Herding | 0.0001 | 0.01 | 0.994 | | | % Herding | 0.02 | 2.53 | 0.019 | | | % Herding | 0.004 | 0.52 | 0.605 |
| | | % Cult. Land | -0.02 | -1.61 | 0.124 | | | % Cult. Land | -0.02 | -1.94 | 0.069 | | | % Cult. Land | 0.004 | 0.33 | 0.743 |
| | | Clim. Dem. | 0.02 | 0.76 | 0.458 | | | Clim. Dem. | 0.03 | 1.23 | 0.232 | | | Clim. Dem. | 0.02 | 0.72 | 0.478 |
| | | Clim. x GDP | 0.01 | 0.14 | 0.890 | | | Clim. x GDP | -0.09 | -1.58 | 0.129 | | | Clim. x GDP | -0.04 | -0.74 | 0.468 |
| | | Rice % | -0.05 | -3.11 | 0.006 | | | Rice % | -0.03 | -2.43 | 0.026 | | | Rice % | 0.004 | 0.29 | 0.777 |
| | Possibility/ Openness | Female | 0.08 | 37.81 | < 0.001 | | In/Outgroup: Dividing | Female | -0.01 | -2.89 | 0.004 | | Affect | Female | 0.26 | 137.07 | < 0.001 |
| | | GDP | -0.0004 | -0.01 | 0.996 | | | GDP | -0.003 | -0.03 | 0.973 | | | GDP | -0.08 | -0.91 | 0.371 |
| | | % Urban | -0.05 | -1.13 | 0.272 | | | % Urban | 0.12 | 2.39 | 0.028 | | | % Urban | -0.04 | -0.86 | 0.402 |
| | | % Herding | -0.002 | -0.24 | 0.813 | | | % Herding | 0.01 | 1.12 | 0.270 | | | % Herding | 0.01 | 1.02 | 0.316 |
| | | % Cult. Land | -0.02 | -1.08 | 0.292 | | | % Cult. Land | 0.005 | 0.28 | 0.780 | | | % Cult. Land | -0.03 | -1.57 | 0.131 |
| | | Clim. Dem. | 0.005 | 0.18 | 0.862 | | | Clim. Dem. | -0.01 | -0.20 | 0.841 | | | Clim. Dem. | -0.01 | -0.32 | 0.751 |
| | | Clim. x GDP | 0.03 | 0.36 | 0.719 | | | Clim. x GDP | -0.10 | -1.12 | 0.272 | | | Clim. x GDP | 0.06 | 0.82 | 0.423 |
| | | Rice % | -0.03 | -1.91 | 0.071 | | | Rice % | -0.03 | -1.46 | 0.161 | | | Rice % | -0.0001 | -0.01 | 0.994 |
| | Assent | Female | 0.18 | 93.52 | < 0.001 | | Self | Female | -0.05 | -17.59 | < 0.001 | **Cog. & Discourse** | Non-Fluencies | Female | 0.18 | 90.22 | < 0.001 |
| | | GDP | -0.15 | -1.55 | 0.137 | | | GDP | 0.11 | 1.65 | 0.114 | | | GDP | -0.16 | -1.76 | 0.094 |
| | | % Urban | 0.004 | 0.08 | 0.935 | | | % Urban | 0.01 | 0.28 | 0.779 | | | % Urban | 0.02 | 0.40 | 0.695 |
| | | % Herding | -0.01 | -0.44 | 0.666 | | | % Herding | -0.004 | -0.46 | 0.649 | | | % Herding | -0.001 | -0.08 | 0.936 |
| | | % Cult. Land | 0.001 | 0.05 | 0.960 | | | % Cult. Land | -0.02 | -1.62 | 0.123 | | | % Cult. Land | 0.003 | 0.15 | 0.886 |
| | | Clim. Dem. | -0.06 | -1.83 | 0.081 | | | Clim. Dem. | 0.03 | 1.38 | 0.181 | | | Clim. Dem. | -0.06 | -1.80 | 0.085 |
| | | Clim. x GDP | 0.16 | 1.74 | 0.096 | | | Clim. x GDP | -0.12 | -2.01 | 0.056 | | | Clim. x GDP | 0.16 | 1.84 | 0.080 |
| | | Rice % | 0.09 | 4.03 | < 0.001 | | | Rice % | -0.03 | -2.10 | 0.049 | | | Rice % | 0.07 | 3.47 | 0.002 |

**Table S10**

*Differences Between Rice and Wheat Prefectures Are Robust to Climatic Demands Combined with GDP*

| Group | Word Category | | ß | t | P | Group | Word Category | | ß | t | P | Group | Word Category | | ß | t | P |
|---|---|---|---|---|---|---|---|---|---|---|---|---|---|---|---|---|---|
| Cognition and Discourse | Cognitive Processes | Female | 0.08 | 24.19 | < 0.001 | Self and Groups | Universalism | Female | -0.09 | -23.19 | < 0.001 | Promotion Orientation and Emotion | Positivity/ Optimism | Female | -0.01 | -1.68 | 0.092 |
| | | GDP | -0.01 | -0.63 | 0.530 | | | GDP | 0.04 | 2.75 | 0.007 | | | GDP | -0.01 | -0.58 | 0.565 |
| | | Clim. Dem. | 0.01 | 1.07 | 0.287 | | | Clim. Dem. | 0.03 | 3.01 | 0.003 | | | Clim. Dem. | 0.001 | 0.10 | 0.924 |
| | | Clim. x GDP | -0.02 | -0.90 | 0.367 | | | Clim. x GDP | -0.05 | -2.70 | 0.008 | | | Clim. x GDP | -0.02 | -1.06 | 0.288 |
| | | Rice % | -0.08 | -8.20 | < 0.001 | | | Rice % | -0.04 | -5.83 | < 0.001 | | | Rice % | -0.08 | -8.46 | < 0.001 |
| | Causation | Female | -0.13 | -37.53 | < 0.001 | | Humans | Female | 0.12 | 33.91 | < 0.001 | | Achievement | Female | -0.23 | -67.34 | < 0.001 |
| | | GDP | 0.03 | 1.49 | 0.139 | | | GDP | -0.003 | -0.11 | 0.912 | | | GDP | 0.05 | 2.53 | 0.012 |
| | | Clim. Dem. | 0.02 | 2.06 | 0.039 | | | Clim. Dem. | 0.04 | 3.34 | < 0.001 | | | Clim. Dem. | 0.03 | 2.71 | 0.007 |
| | | Clim. x GDP | -0.02 | -1.09 | 0.278 | | | Clim. x GDP | -0.05 | -1.98 | 0.048 | | | Clim. x GDP | -0.04 | -1.92 | 0.056 |
| | | Rice % | -0.07 | -8.18 | < 0.001 | | | Rice % | -0.07 | -6.62 | < 0.001 | | | Rice % | -0.06 | -7.23 | < 0.001 |
| | Certainty | Female | 0.10 | 27.62 | < 0.001 | | In/Outgroup: Connecting | Female | -0.0004 | -0.11 | 0.909 | | Fashion and Trends | Female | 0.02 | 3.90 | < 0.001 |
| | | GDP | -0.01 | -0.37 | 0.710 | | | GDP | -0.05 | -2.76 | 0.007 | | | GDP | 0.01 | 0.35 | 0.724 |
| | | Clim. Dem. | 0.02 | 1.72 | 0.087 | | | Clim. Dem. | 0.001 | 0.11 | 0.915 | | | Clim. Dem. | -0.003 | -0.22 | 0.826 |
| | | Clim. x GDP | -0.03 | -1.46 | 0.147 | | | Clim. x GDP | 0.004 | 0.22 | 0.828 | | | Clim. x GDP | 0.01 | 0.60 | 0.551 |
| | | Rice % | -0.07 | -7.32 | < 0.001 | | | Rice % | -0.03 | -4.27 | < 0.001 | | | Rice % | 0.01 | 1.62 | 0.108 |
| | Possibility/ Openness | Female | 0.08 | 22.95 | < 0.001 | | In/Outgroup: Dividing | Female | -0.03 | -3.36 | < 0.001 | | Affect | Female | 0.28 | 82.17 | < 0.001 |
| | | GDP | -0.01 | -0.29 | 0.774 | | | GDP | 0.01 | 0.39 | 0.694 | | | GDP | -0.10 | -4.11 | < 0.001 |
| | | Clim. Dem. | -0.0003 | -0.03 | 0.978 | | | Clim. Dem. | -0.04 | -1.60 | 0.111 | | | Clim. Dem. | -0.03 | -2.17 | 0.031 |
| | | Clim. x GDP | -0.02 | -1.22 | 0.225 | | | Clim. x GDP | 0.002 | 0.04 | 0.970 | | | Clim. x GDP | 0.04 | 1.65 | 0.100 |
| | | Rice % | -0.05 | -6.37 | < 0.001 | | | Rice % | -0.03 | -1.96 | 0.054 | | | Rice % | -0.03 | -2.46 | 0.015 |
| | Assent | Female | 0.21 | 59.94 | < 0.001 | | Self | Female | -0.05 | -10.46 | < 0.001 | Cog. & Dis. | Non-Fluencies | Female | 0.20 | 57.56 | < 0.001 |
| | | GDP | -0.05 | -2.36 | 0.019 | | | GDP | 0.07 | 3.29 | 0.001 | | | GDP | -0.02 | -1.28 | 0.203 |
| | | Clim. Dem. | -0.04 | -3.69 | < 0.001 | | | Clim. Dem. | 0.01 | 0.85 | 0.396 | | | Clim. Dem. | -0.02 | -1.76 | 0.078 |
| | | Clim. x GDP | 0.04 | 1.80 | 0.073 | | | Clim. x GDP | -0.06 | -2.76 | 0.006 | | | Clim. x GDP | 0.01 | 0.61 | 0.542 |
| | | Rice % | 0.05 | 5.82 | < 0.001 | | | Rice % | -0.03 | -2.87 | 0.005 | | | Rice % | 0.04 | 4.51 | < 0.001 |

Note: Analyses are hierarchical linear models nested in prefectures. Climatic demands are calculated as the sum deviation from 22C in the average highs and lows of July and January. Climatic demands theory argues that cultures in harsher climates should be more collectivistic. A further version of this theory is that climatic demands have a weaker influence on culture in more developed areas because people use resources to buffer against the climate. The analyses in the main text use 2014 GDP at the request of a reviewer, but 2012 and 2014 prefecture GDP correlate $r = 0.98$, so the difference is negligible.

**Table S11**

*Alternative Measures of Modernization: Service Sector*

| | Word Category | | ß | t | P | | Word Category | ß | t | P | | Word Category | | ß | t | P |
|---|---|---|---|---|---|---|---|---|---|---|---|---|---|---|---|---|
| **Cognition and Discourse** | Cognitive Processes | Female | 0.08 | 40.98 | < 0.001 | Self and Groups | Universalism | | | | Promotion Orientation and Emotion | Positivity/ Optimism | Female | 0.001 | 0.66 | 0.509 |
| | | Service Ind. | -0.06 | -2.19 | 0.040 | | Female | -0.08 | -36.86 | < 0.001 | | | Service Ind. | -0.05 | -1.47 | 0.156 |
| | | % Urban | 0.01 | 0.31 | 0.756 | | Service Ind. | 0.01 | 0.42 | 0.678 | | | % Urban | -0.002 | -0.07 | 0.948 |
| | | % Herding | 0.01 | 1.30 | 0.205 | | % Urban | -0.01 | -0.28 | 0.780 | | | % Herding | 0.01 | 1.00 | 0.327 |
| | | % Cult. Land | -0.02 | -1.19 | 0.247 | | % Herding | 0.02 | 2.37 | 0.025 | | | % Cult. Land | -0.02 | -1.07 | 0.295 |
| | | Rice % | -0.08 | -5.35 | < 0.001 | | % Cult. Land | -0.01 | -0.76 | 0.456 | | | Rice % | -0.08 | -4.68 | < 0.001 |
| | | | | | | | Rice % | -0.05 | -3.71 | 0.001 | | | | | | |
| | Causation | Female | -0.12 | -58.56 | < 0.001 | | Humans | | | | | Achievement | Female | -0.21 | -109.01 | < 0.001 |
| | | Service Ind. | -0.02 | -0.77 | 0.448 | | Female | 0.11 | 55.50 | < 0.001 | | | Service Ind. | -0.01 | -0.39 | 0.700 |
| | | % Urban | 0.02 | 0.79 | 0.438 | | Service Ind. | -0.05 | -1.66 | 0.112 | | | % Urban | 0.02 | 0.77 | 0.452 |
| | | % Herding | 0.01 | 1.54 | 0.137 | | % Urban | -0.01 | -0.22 | 0.824 | | | % Herding | 0.01 | 1.43 | 0.164 |
| | | % Cult. Land | -0.00004 | -0.002 | 0.998 | | % Herding | 0.02 | 2.17 | 0.040 | | | % Cult. Land | 0.003 | 0.15 | 0.884 |
| | | Rice % | -0.07 | -4.49 | < 0.001 | | % Cult. Land | -0.02 | -1.01 | 0.324 | | | Rice % | -0.07 | -3.87 | < 0.001 |
| | | | | | | | Rice % | -0.08 | -4.46 | < 0.001 | | | | | | |
| | Certainty | Female | 0.09 | 45.01 | < 0.001 | | In/Outgroup: Connecting | | | | | Fashion and Trends | Female | 0.04 | 15.57 | < 0.001 |
| | | Service Ind. | -0.03 | -1.23 | 0.234 | | Female | 0.0003 | 0.12 | 0.905 | | | Service Ind. | 0.01 | 0.59 | 0.559 |
| | | % Urban | -0.01 | -0.55 | 0.589 | | Service Ind. | 0.003 | 0.18 | 0.863 | | | % Urban | 0.001 | 0.04 | 0.969 |
| | | % Herding | 0.01 | 1.06 | 0.298 | | % Urban | -0.04 | -2.46 | 0.022 | | | % Herding | 0.004 | 0.68 | 0.501 |
| | | % Cult. Land | -0.02 | -1.72 | 0.101 | | % Herding | 0.02 | 2.90 | 0.007 | | | % Cult. Land | 0.01 | 0.69 | 0.497 |
| | | Rice % | -0.06 | -4.95 | < 0.001 | | % Cult. Land | -0.02 | -1.50 | 0.150 | | | Rice % | 0.01 | 0.81 | 0.428 |
| | | | | | | | Rice % | -0.02 | -1.97 | 0.063 | | | | | | |
| | Possibility/ Openness | Female | 0.08 | 37.81 | < 0.001 | | In/Outgroup: Dividing | | | | | Affect | Female | 0.26 | 137.08 | < 0.001 |
| | | Service Ind. | -0.05 | -2.34 | 0.030 | | Female | -0.01 | -2.93 | 0.003 | | | Service Ind. | -0.03 | -1.26 | 0.223 |
| | | % Urban | 0.01 | 0.71 | 0.483 | | Service Ind. | 0.07 | 2.16 | 0.043 | | | % Urban | -0.04 | -1.70 | 0.103 |
| | | % Herding | 0.004 | 0.53 | 0.601 | | % Urban | -0.03 | -1.01 | 0.322 | | | % Herding | 0.01 | 1.65 | 0.112 |
| | | % Cult. Land | -0.02 | -1.65 | 0.115 | | % Herding | -0.01 | -0.66 | 0.512 | | | % Cult. Land | -0.03 | -2.18 | 0.040 |
| | | Rice % | -0.05 | -4.03 | < 0.001 | | % Cult. Land | 0.003 | 0.17 | 0.870 | | | Rice % | -0.02 | -1.08 | 0.294 |
| | | | | | | | Rice % | 0.02 | 0.96 | 0.347 | | | | | | |
| | Assent | Female | 0.18 | 93.53 | < 0.001 | | Self | | | | Cog. & Dis. | Non-Fluencies | Female | 0.18 | 90.23 | < 0.001 |
| | | Service Ind. | 0.03 | 0.81 | 0.426 | | Female | -0.05 | -17.61 | < 0.001 | | | Service Ind. | -0.01 | -0.43 | 0.668 |
| | | % Urban | -0.04 | -1.36 | 0.187 | | Service Ind. | 0.02 | 0.65 | 0.521 | | | % Urban | -0.01 | -0.20 | 0.845 |
| | | % Herding | -0.01 | -0.97 | 0.342 | | % Urban | 0.004 | 0.21 | 0.836 | | | % Herding | -0.004 | -0.48 | 0.638 |
| | | % Cult. Land | -0.004 | -0.24 | 0.816 | | % Herding | -0.01 | -1.06 | 0.300 | | | % Cult. Land | -0.01 | -0.52 | 0.606 |
| | | Rice % | 0.08 | 4.60 | < 0.001 | | % Cult. Land | -0.01 | -1.11 | 0.279 | | | Rice % | 0.06 | 3.43 | 0.002 |
| | | | | | | | Rice % | -0.01 | -0.89 | 0.384 | | | | | | |

Note: Service sector employment is the percentage of employed people that are employed in the service sector per province. Modernization theorist Inglehart has argued that the shift to the service sector economy is a better indicator of modernization than GDP. Analyses are hierarchical linear models nested in provinces.

**Table S12**

*Alternative Measures of Modernization: Private Industry*

| | Word Category | | β | t | P | | Word Category | | β | t | P | | Word Category | | β | t | P |
|---|---|---|---|---|---|---|---|---|---|---|---|---|---|---|---|---|---|
| | | Female | 0.08 | 40.98 | < 0.001 | | | Female | -0.08 | -36.86 | < 0.001 | | | Female | 0.001 | 0.66 | 0.507 |
| | | Private Ind. | -0.02 | -0.69 | 0.496 | | | Private Ind. | 0.07 | 3.16 | 0.005 | | | Private Ind. | -0.001 | -0.03 | 0.976 |
| | Cognitive | % Urban | -0.02 | -0.97 | 0.344 | | | % Urban | -0.03 | -2.30 | 0.031 | | Positivity/ | % Urban | -0.03 | -1.31 | 0.204 |
| | Processes | % Herding | 0.01 | 1.40 | 0.174 | | Universalism | % Herding | 0.01 | 1.36 | 0.183 | | Optimism | % Herding | 0.01 | 0.92 | 0.366 |
| | | % Cult. Land | -0.003 | -0.20 | 0.847 | | | % Cult. Land | -0.03 | -2.39 | 0.025 | | | % Cult. Land | -0.01 | -0.58 | 0.565 |
| | | Rice % | -0.06 | -3.59 | 0.002 | | | Rice % | -0.07 | -5.54 | < 0.001 | | | Rice % | -0.08 | -3.67 | 0.001 |
| | | Female | -0.12 | -58.55 | < 0.001 | | | Female | 0.11 | 55.51 | < 0.001 | | | Female | -0.21 | -109.01 | < 0.001 |
| | | Private Ind. | 0.01 | 0.34 | 0.734 | | | Private Ind. | 0.02 | 0.62 | 0.539 | | | Private Ind. | 0.01 | 0.27 | 0.792 |
| | Causation | % Urban | -0.002 | -0.07 | 0.944 | | Humans | % Urban | -0.06 | -2.12 | 0.045 | | Achievement | % Urban | 0.01 | 0.31 | 0.763 |
| | | % Herding | 0.01 | 1.29 | 0.208 | | | % Herding | 0.02 | 1.69 | 0.103 | | | % Herding | 0.01 | 1.23 | 0.231 |
| Cognition and Discourse | | % Cult. Land | 0.001 | 0.05 | 0.960 | Self and Groups | | % Cult. Land | -0.02 | -0.76 | 0.454 | Promotion Orientation and Emotion | | % Cult. Land | 0.003 | 0.13 | 0.900 |
| | | Rice % | -0.07 | -3.95 | < 0.001 | | | Rice % | -0.08 | -3.78 | < 0.001 | | | Rice % | -0.07 | -3.47 | 0.002 |
| | | Female | 0.09 | 45.01 | < 0.001 | | | Female | 0.0003 | 0.12 | 0.904 | | | Female | 0.04 | 15.56 | < 0.001 |
| | | Private Ind. | -0.02 | -0.60 | 0.554 | | | Private Ind. | 0.03 | 1.21 | 0.240 | | | Private Ind. | -0.01 | -0.59 | 0.564 |
| | Certainty | % Urban | -0.02 | -1.25 | 0.226 | | In/Outgroup: | % Urban | -0.05 | -3.63 | 0.002 | | Fashion and | % Urban | 0.02 | 1.06 | 0.302 |
| | | % Herding | 0.01 | 1.20 | 0.240 | | Connecting | % Herding | 0.02 | 2.29 | 0.030 | | Trends | % Herding | 0.01 | 0.85 | 0.401 |
| | | % Cult. Land | -0.01 | -0.97 | 0.344 | | | % Cult. Land | -0.02 | -2.04 | 0.054 | | | % Cult. Land | 0.01 | 0.76 | 0.454 |
| | | Rice % | -0.05 | -3.67 | 0.001 | | | Rice % | -0.03 | -2.45 | 0.023 | | | Rice % | 0.01 | 0.90 | 0.377 |
| | | Female | 0.08 | 37.82 | < 0.001 | | | Female | -0.01 | -2.93 | 0.003 | | | Female | 0.26 | 137.08 | < 0.001 |
| | | Private Ind. | -0.04 | -1.44 | 0.166 | | | Private Ind. | 0.07 | 1.87 | 0.076 | | | Private Ind. | -0.03 | -1.13 | 0.272 |
| | Possibility/ | % Urban | -0.003 | -0.17 | 0.869 | | In/Outgroup: | % Urban | -0.01 | -0.51 | 0.616 | | Affect | % Urban | -0.04 | -2.16 | 0.042 |
| | Openness | % Herding | 0.01 | 1.04 | 0.309 | | Dividing | % Herding | -0.02 | -1.32 | 0.197 | | | % Herding | 0.02 | 1.95 | 0.063 |
| | | % Cult. Land | -0.003 | -0.21 | 0.836 | | | % Cult. Land | -0.03 | -1.43 | 0.167 | | | % Cult. Land | -0.02 | -1.19 | 0.246 |
| | | Rice % | -0.03 | -2.14 | 0.043 | | | Rice % | -0.02 | -0.73 | 0.470 | | | Rice % | -0.002 | -0.12 | 0.908 |
| | | Female | 0.18 | 93.52 | < 0.001 | | | Female | -0.05 | -17.61 | < 0.001 | | | Female | 0.18 | 90.23 | < 0.001 |
| | | Private Ind. | -0.03 | -0.81 | 0.426 | | | Private Ind. | 0.02 | 0.74 | 0.466 | | | Private Ind. | -0.04 | -1.03 | 0.314 |
| | Assent | % Urban | -0.005 | -0.18 | 0.857 | | Self | % Urban | 0.01 | 0.29 | 0.772 | Cog. & Dis. | Non-Fluencies | % Urban | 0.003 | 0.14 | 0.886 |
| | | % Herding | -0.01 | -0.60 | 0.556 | | | % Herding | -0.01 | -1.27 | 0.215 | | | % Herding | -0.001 | -0.05 | 0.959 |
| | | % Cult. Land | -0.002 | -0.09 | 0.932 | | | % Cult. Land | -0.02 | -1.56 | 0.132 | | | % Cult. Land | 0.001 | 0.06 | 0.950 |
| | | Rice % | 0.09 | 4.30 | < 0.001 | | | Rice % | -0.02 | -1.31 | 0.204 | | | Rice % | 0.07 | 3.69 | 0.001 |

Note: Private industry employment is the percentage of employed people that are employed in private industry per province. This represents modernization in China's shift from the state-run economy to private enterprise. Analyses are hierarchical linear models nested in provinces.

**Table S13**

*Alternative Measures of Modernization: Internet Penetration*

| Group | Word Category | Variable | ß | t | P | Group | Word Category | Variable | ß | t | P | Group | Word Category | Variable | ß | t | P |
|---|---|---|---|---|---|---|---|---|---|---|---|---|---|---|---|---|---|
| Cognition and Discourse | Cognitive Processes | Female | 0.08 | 40.98 | < 0.001 | Self and Groups | Universalism | Female | -0.08 | -36.84 | < 0.001 | Promotion Orientation and Emotion | Positivity/ Optimism | Female | 0.001 | 0.67 | 0.506 |
| | | Internet Pen. | -0.04 | -1.21 | 0.240 | | | Internet Pen. | 0.04 | 1.52 | 0.144 | | | Internet Pen. | 0.01 | 0.23 | 0.823 |
| | | % Urban | 0.0003 | 0.01 | 0.992 | | | % Urban | -0.03 | -1.30 | 0.205 | | | % Urban | -0.04 | -1.14 | 0.267 |
| | | % Herding | 0.01 | 1.35 | 0.188 | | | % Herding | 0.02 | 2.38 | 0.025 | | | % Herding | 0.01 | 0.96 | 0.346 |
| | | % Cult. Land | -0.01 | -0.76 | 0.455 | | | % Cult. Land | -0.01 | -0.63 | 0.534 | | | % Cult. Land | -0.01 | -0.62 | 0.544 |
| | | Rice % | -0.07 | -4.22 | < 0.001 | | | Rice % | -0.05 | -4.35 | < 0.001 | | | Rice % | -0.08 | -4.20 | < 0.001 |
| | Causation | Female | -0.12 | -58.55 | < 0.001 | | Humans | Female | 0.11 | 55.51 | < 0.001 | | Achievement | Female | -0.21 | -109.00 | < 0.001 |
| | | Internet Pen. | 0.02 | 0.61 | 0.550 | | | Internet Pen. | -0.004 | -0.09 | 0.933 | | | Internet Pen. | 0.05 | 1.28 | 0.214 |
| | | % Urban | -0.01 | -0.41 | 0.688 | | | % Urban | -0.04 | -1.08 | 0.293 | | | % Urban | -0.03 | -0.76 | 0.452 |
| | | % Herding | 0.01 | 1.49 | 0.149 | | | % Herding | 0.02 | 2.07 | 0.049 | | | % Herding | 0.01 | 1.36 | 0.184 |
| | | % Cult. Land | 0.01 | 0.33 | 0.746 | | | % Cult. Land | -0.01 | -0.55 | 0.587 | | | % Cult. Land | 0.01 | 0.49 | 0.629 |
| | | Rice % | -0.07 | -4.45 | < 0.001 | | | Rice % | -0.07 | -3.83 | < 0.001 | | | Rice % | -0.07 | -4.21 | < 0.001 |
| | Certainty | Female | 0.09 | 45.00 | < 0.001 | | In/Outgroup: Connecting | Female | 0.0003 | 0.13 | 0.895 | | Fashion and Trends | Female | 0.04 | 15.58 | < 0.001 |
| | | Internet Pen. | -0.03 | -0.96 | 0.349 | | | Internet Pen. | 0.03 | 1.43 | 0.168 | | | Internet Pen. | 0.03 | 1.08 | 0.291 |
| | | % Urban | -0.01 | -0.33 | 0.741 | | | % Urban | -0.07 | -3.18 | 0.004 | | | % Urban | -0.01 | -0.55 | 0.587 |
| | | % Herding | 0.01 | 1.14 | 0.267 | | | % Herding | 0.02 | 2.95 | 0.007 | | | % Herding | 0.004 | 0.63 | 0.531 |
| | | % Cult. Land | -0.02 | -1.56 | 0.134 | | | % Cult. Land | -0.01 | -1.37 | 0.187 | | | % Cult. Land | 0.01 | 0.78 | 0.445 |
| | | Rice % | -0.05 | -4.34 | < 0.001 | | | Rice % | -0.03 | -2.48 | 0.023 | | | Rice % | 0.004 | 0.42 | 0.680 |
| | Possibility/ Openness | Female | 0.08 | 37.81 | < 0.001 | | In/Outgroup: Dividing | Female | -0.01 | -2.88 | 0.004 | | Affect | Female | 0.26 | 137.07 | < 0.001 |
| | | Internet Pen. | -0.03 | -1.12 | 0.276 | | | Internet Pen. | 0.11 | 3.10 | 0.005 | | | Internet Pen. | -0.06 | -1.84 | 0.079 |
| | | % Urban | 0.004 | 0.16 | 0.877 | | | % Urban | -0.07 | -2.10 | 0.046 | | | % Urban | -0.02 | -0.56 | 0.580 |
| | | % Herding | 0.004 | 0.61 | 0.547 | | | % Herding | -0.01 | -0.81 | 0.422 | | | % Herding | 0.01 | 1.87 | 0.074 |
| | | % Cult. Land | -0.02 | -1.13 | 0.271 | | | % Cult. Land | 0.004 | 0.29 | 0.775 | | | % Cult. Land | -0.03 | -2.26 | 0.034 |
| | | Rice % | -0.04 | -2.92 | 0.008 | | | Rice % | -0.01 | -0.39 | 0.703 | | | Rice % | -0.01 | -0.36 | 0.719 |
| | Assent | Female | 0.18 | 93.52 | < 0.001 | | Self | Female | -0.05 | -17.59 | < 0.001 | Cog. & Dis. | Non-Fluencies | Female | 0.18 | 90.22 | < 0.001 |
| | | Internet Pen. | -0.04 | -0.93 | 0.365 | | | Internet Pen. | 0.07 | 2.61 | 0.015 | | | Internet Pen. | -0.07 | -1.91 | 0.069 |
| | | % Urban | 0.01 | 0.26 | 0.800 | | | % Urban | -0.04 | -1.68 | 0.104 | | | % Urban | 0.04 | 1.23 | 0.232 |
| | | % Herding | -0.01 | -0.90 | 0.377 | | | % Herding | -0.01 | -1.35 | 0.186 | | | % Herding | -0.003 | -0.33 | 0.747 |
| | | % Cult. Land | -0.01 | -0.64 | 0.531 | | | % Cult. Land | -0.01 | -0.94 | 0.360 | | | % Cult. Land | -0.01 | -0.76 | 0.454 |
| | | Rice % | 0.09 | 4.67 | < 0.001 | | | Rice % | -0.02 | -1.83 | 0.082 | | | Rice % | 0.07 | 4.23 | < 0.001 |

Note: Internet penetration statistics are the percentages of internet users to total province population from the *China Internet Development Report 2008*. Internet penetration is an alternative indicator of modernization that is less sensitive to manipulation. Analyses are hierarchical linear models nested in provinces.

**Table S14**

*Alternative Measures of Modernization: Historical GDP per Capita*

| | Word Category | | ß | t | P | | Word Category | | ß | t | P | | Word Category | | ß | t | P |
|---|---|---|---|---|---|---|---|---|---|---|---|---|---|---|---|---|---|
| | | Female | 0.08 | 40.98 | < 0.001 | | | Female | -0.08 | -36.85 | < 0.001 | | | Female | 0.001 | 0.66 | 0.507 |
| | | GDP 1995 | -0.05 | -1.30 | 0.206 | | | GDP 1995 | 0.07 | 2.28 | 0.033 | | | GDP 1995 | -0.02 | -0.33 | 0.742 |
| | Cognitive | % Urban | 0.003 | 0.10 | 0.922 | | | % Urban | -0.05 | -1.99 | 0.060 | | Positivity/ | % Urban | -0.02 | -0.63 | 0.533 |
| | Processes | % Herding | 0.02 | 1.70 | 0.101 | | Universalism | % Herding | 0.01 | 1.36 | 0.187 | | Optimism | % Herding | 0.01 | 1.03 | 0.313 |
| | | % Cult. Land | 0.002 | 0.10 | 0.919 | | | % Cult. Land | -0.02 | -1.94 | 0.066 | | | % Cult. Land | -0.01 | -0.43 | 0.669 |
| | | Rice % | -0.06 | -3.35 | 0.003 | | | Rice % | -0.06 | -4.87 | < 0.001 | | | Rice % | -0.07 | -3.50 | 0.002 |
| | | Female | -0.12 | -58.55 | < 0.001 | | | Female | 0.11 | 55.51 | < 0.001 | | | Female | -0.21 | -109.01 | < 0.001 |
| | | GDP 1995 | -0.001 | -0.02 | 0.983 | | | GDP 1995 | -0.01 | -0.12 | 0.904 | | | GDP 1995 | 0.02 | 0.33 | 0.741 |
| | | % Urban | 0.005 | 0.15 | 0.884 | | | % Urban | -0.04 | -1.03 | 0.312 | | | % Urban | 0.002 | 0.06 | 0.956 |
| | Causation | % Herding | 0.01 | 1.38 | 0.180 | | Humans | % Herding | 0.02 | 1.90 | 0.068 | | Achievement | % Herding | 0.01 | 1.14 | 0.267 |
| | | % Cult. Land | 0.004 | 0.21 | 0.838 | | | % Cult. Land | -0.01 | -0.43 | 0.674 | | | % Cult. Land | 0.002 | 0.08 | 0.934 |
| Cognition and Discourse | | Rice % | -0.07 | -3.75 | 0.001 | Self and Groups | | Rice % | -0.07 | -3.36 | 0.003 | Promotion Orientation and Emotion | | Rice % | -0.07 | -3.50 | 0.002 |
| | | Female | 0.09 | 45.00 | < 0.001 | | | Female | 0.0003 | 0.12 | 0.903 | | | Female | 0.04 | 15.57 | < 0.001 |
| | | GDP 1995 | -0.05 | -1.43 | 0.166 | | | GDP 1995 | 0.02 | 0.81 | 0.428 | | | GDP 1995 | 0.02 | 0.64 | 0.526 |
| | | % Urban | 0.002 | 0.06 | 0.951 | | In/Outgroup: | % Urban | -0.06 | -2.56 | 0.018 | | Fashion and | % Urban | -0.004 | -0.16 | 0.873 |
| | Certainty | % Herding | 0.01 | 1.61 | 0.120 | | Connecting | % Herding | 0.02 | 2.29 | 0.030 | | Trends | % Herding | 0.002 | 0.33 | 0.741 |
| | | % Cult. Land | -0.01 | -0.60 | 0.557 | | | % Cult. Land | -0.02 | -1.83 | 0.082 | | | % Cult. Land | 0.002 | 0.17 | 0.869 |
| | | Rice % | -0.05 | -3.37 | 0.003 | | | Rice % | -0.03 | -2.21 | 0.038 | | | Rice % | 0.003 | 0.23 | 0.816 |
| | | Female | 0.08 | 37.81 | < 0.001 | | | Female | -0.01 | -2.93 | 0.003 | | | Female | 0.26 | 137.08 | < 0.001 |
| | | GDP 1995 | -0.05 | -1.34 | 0.194 | | | GDP 1995 | 0.14 | 2.98 | 0.007 | | | GDP 1995 | -0.03 | -0.85 | 0.406 |
| | Possibility/ | % Urban | 0.01 | 0.35 | 0.731 | | In/Outgroup: | % Urban | -0.07 | -1.99 | 0.059 | | Affect | % Urban | -0.04 | -1.31 | 0.203 |
| | Openness | % Herding | 0.01 | 1.07 | 0.297 | | Dividing | % Herding | -0.02 | -1.96 | 0.059 | | | % Herding | 0.02 | 1.83 | 0.079 |
| | | % Cult. Land | -0.003 | -0.22 | 0.830 | | | % Cult. Land | -0.04 | -2.02 | 0.057 | | | % Cult. Land | -0.02 | -1.25 | 0.223 |
| | | Rice % | -0.03 | -2.15 | 0.042 | | | Rice % | -0.03 | -1.39 | 0.176 | | | Rice % | -0.004 | -0.25 | 0.805 |
| | | Female | 0.18 | 93.52 | < 0.001 | | | Female | -0.05 | -17.61 | < 0.001 | | | Female | 0.18 | 90.23 | < 0.001 |
| | | GDP 1995 | -0.01 | -0.14 | 0.892 | | | GDP 1995 | 0.04 | 1.23 | 0.233 | | | GDP 1995 | -0.03 | -0.71 | 0.487 |
| | | % Urban | -0.02 | -0.42 | 0.676 | | | % Urban | -0.01 | -0.50 | 0.624 | Cog. & Dis. | Non- | % Urban | 0.01 | 0.20 | 0.845 |
| | Assent | % Herding | -0.01 | -0.81 | 0.428 | | Self | % Herding | -0.01 | -1.52 | 0.139 | | Fluencies | % Herding | -0.001 | -0.11 | 0.916 |
| | | % Cult. Land | -0.01 | -0.36 | 0.724 | | | % Cult. Land | -0.02 | -1.82 | 0.083 | | | % Cult. Land | -0.001 | -0.05 | 0.964 |
| | | Rice % | 0.08 | 3.91 | < 0.001 | | | Rice % | -0.02 | -1.59 | 0.126 | | | Rice % | 0.07 | 3.48 | 0.002 |

Note: Provincial GDP per capita statistics are year 1995 in RMB. Studies have found a lag between economic growth and cultural change, so we tested this historical indicator of economic development. Analyses are hierarchical linear models nested in provinces.

**Table S15**

*Alternative Measures of Modernization: Historical Service Industry*

| | Word Category | β | t | P | | Word Category | β | t | P | | Word Category | β | t | P |
|---|---|---|---|---|---|---|---|---|---|---|---|---|---|---|
| **Cognition and Discourse** | Cognitive Processes | | | | | Universalism | | | | | Positivity/ Optimism | | | |
| | Female | 0.08 | 40.98 | < 0.001 | | Female | -0.08 | -36.86 | < 0.001 | | Female | 0.001 | 0.66 | 0.508 |
| | Serv. Ind. 1995 | -0.03 | -1.62 | 0.118 | | Serv. Ind. 1995 | 0.01 | 0.57 | 0.574 | | Serv. Ind. 1995 | -0.03 | -1.32 | 0.197 |
| | % Urban | -0.01 | -0.55 | 0.584 | | % Urban | -0.01 | -0.33 | 0.741 | | % Urban | -0.01 | -0.60 | 0.557 |
| | % Herding | 0.002 | 0.15 | 0.880 | | % Herding | 0.02 | 2.28 | 0.031 | | % Herding | 0.001 | 0.10 | 0.920 |
| | % Cult. Land | -0.01 | -0.85 | 0.405 | | % Cult. Land | -0.01 | -0.80 | 0.433 | | % Cult. Land | -0.02 | -0.90 | 0.379 |
| | Rice % | -0.08 | -5.07 | < 0.001 | | Rice % | -0.05 | -3.62 | 0.001 | | Rice % | -0.08 | -4.65 | < 0.001 |
| | Causation | | | | | Humans | | | | | Achievement | | | |
| | Female | -0.12 | -58.56 | < 0.001 | | Female | 0.11 | 55.50 | < 0.001 | | Female | -0.21 | -109.01 | < 0.001 |
| | Serv. Ind. 1995 | -0.03 | -1.69 | 0.104 | | Serv. Ind. 1995 | -0.04 | -1.59 | 0.125 | | Serv. Ind. 1995 | -0.02 | -1.05 | 0.303 |
| | % Urban | 0.03 | 1.39 | 0.178 | | % Urban | -0.02 | -0.79 | 0.439 | | % Urban | 0.03 | 1.29 | 0.210 |
| | % Herding | 0.004 | 0.40 | 0.696 | | % Herding | 0.01 | 0.92 | 0.367 | | % Herding | 0.01 | 0.63 | 0.537 |
| | % Cult. Land | -0.001 | -0.05 | 0.958 | | % Cult. Land | -0.02 | -0.82 | 0.419 | | % Cult. Land | 0.002 | 0.09 | 0.926 |
| | Rice % | -0.08 | -4.90 | < 0.001 | | Rice % | -0.08 | -4.45 | < 0.001 | | Rice % | -0.07 | -4.09 | < 0.001 |
| | Certainty | | | | **Self and Groups** | In/Outgroup: Connecting | | | | **Promotion Orientation and Emotion** | Fashion and Trends | | | |
| | Female | 0.09 | 45.01 | < 0.001 | | Female | 0.0002 | 0.12 | 0.907 | | Female | 0.04 | 15.57 | < 0.001 |
| | Serv. Ind. 1995 | -0.01 | -0.93 | 0.362 | | Serv. Ind. 1995 | -0.02 | -1.27 | 0.217 | | Serv. Ind. 1995 | 0.01 | 0.60 | 0.556 |
| | % Urban | -0.02 | -1.18 | 0.250 | | % Urban | -0.03 | -1.95 | 0.063 | | % Urban | 0.003 | 0.19 | 0.854 |
| | % Herding | 0.003 | 0.36 | 0.720 | | % Herding | 0.01 | 1.79 | 0.087 | | % Herding | 0.01 | 0.89 | 0.378 |
| | % Cult. Land | -0.02 | -1.56 | 0.135 | | % Cult. Land | -0.02 | -1.90 | 0.073 | | % Cult. Land | 0.01 | 0.65 | 0.521 |
| | Rice % | -0.06 | -4.82 | < 0.001 | | Rice % | -0.03 | -2.43 | 0.025 | | Rice % | 0.01 | 0.83 | 0.417 |
| | Possibility/ Openness | | | | | In/Outgroup: Dividing | | | | | Affect | | | |
| | Female | 0.08 | 37.82 | < 0.001 | | Female | -0.01 | -2.94 | 0.003 | | Female | 0.26 | 137.08 | < 0.001 |
| | Serv. Ind. 1995 | -0.02 | -1.29 | 0.209 | | Serv. Ind. 1995 | 0.03 | 1.30 | 0.204 | | Serv. Ind. 1995 | -0.02 | -1.05 | 0.306 |
| | % Urban | -0.01 | -0.41 | 0.689 | | % Urban | -0.004 | -0.13 | 0.896 | | % Urban | -0.05 | -2.51 | 0.019 |
| | % Herding | -0.002 | -0.29 | 0.778 | | % Herding | 0.003 | 0.20 | 0.844 | | % Herding | 0.01 | 0.77 | 0.447 |
| | % Cult. Land | -0.02 | -1.18 | 0.253 | | % Cult. Land | -0.003 | -0.18 | 0.861 | | % Cult. Land | -0.03 | -2.03 | 0.054 |
| | Rice % | -0.05 | -3.59 | 0.002 | | Rice % | 0.02 | 0.78 | 0.442 | | Rice % | -0.02 | -1.06 | 0.302 |
| | Assent | | | | | Self | | | | **Cog. & Dis.** | Non-Fluencies | | | |
| | Female | 0.18 | 93.52 | < 0.001 | | Female | -0.05 | -17.61 | < 0.001 | | Female | 0.18 | 90.23 | < 0.001 |
| | Serv. Ind. 1995 | 0.01 | 0.49 | 0.627 | | Serv. Ind. 1995 | -0.0001 | -0.01 | 0.994 | | Serv. Ind. 1995 | 0.02 | 0.99 | 0.332 |
| | % Urban | -0.03 | -1.20 | 0.243 | | % Urban | 0.02 | 0.89 | 0.379 | | % Urban | -0.03 | -1.36 | 0.185 |
| | % Herding | -0.01 | -0.53 | 0.598 | | % Herding | -0.01 | -0.88 | 0.384 | | % Herding | 0.002 | 0.16 | 0.875 |
| | % Cult. Land | -0.01 | -0.39 | 0.700 | | % Cult. Land | -0.02 | -1.34 | 0.195 | | % Cult. Land | -0.004 | -0.25 | 0.805 |
| | Rice % | 0.08 | 4.48 | < 0.001 | | Rice % | -0.01 | -1.03 | 0.314 | | Rice % | 0.07 | 3.81 | < 0.001 |

Note: Studies have found a lag between economic growth and cultural change, so we tested this historical indicator of economic modernization. Analyses are hierarchical linear models nested in provinces. These analyses use the percentage of workers employed in the service industry as an indicator of modernization. As economies modernize, they tend to shift toward the service sector.

**Table S16**

*Alternative Measures of Modernization: Historical Private Industry*

| | Word Category | | ß | t | P | | Word Category | | ß | t | P | | Word Category | | ß | t | P |
|---|---|---|---|---|---|---|---|---|---|---|---|---|---|---|---|---|---|
| | | Female | 0.08 | 40.98 | < 0.001 | | | Female | -0.08 | -36.86 | < 0.001 | | | Female | 0.001 | 0.66 | 0.506 |
| | | Priv. Ind. 1995 | 0.02 | 0.69 | 0.495 | | | Priv. Ind. 1995 | 0.02 | 1.29 | 0.210 | | | Priv. Ind. 1995 | 0.02 | 0.93 | 0.360 |
| | Cognitive | % Urban | -0.04 | -2.22 | 0.038 | | Universalism | % Urban | -0.01 | -0.69 | 0.500 | | Positivity/ | % Urban | -0.05 | -2.20 | 0.039 |
| | Processes | % Herding | 0.01 | 1.25 | 0.222 | | | % Herding | 0.02 | 2.45 | 0.021 | | Optimism | % Herding | 0.01 | 1.03 | 0.312 |
| | | % Cult. Land | -0.01 | -0.69 | 0.498 | | | % Cult. Land | -0.01 | -1.21 | 0.242 | | | % Cult. Land | -0.02 | -0.85 | 0.405 |
| | | Rice % | -0.08 | -4.55 | < 0.001 | | | Rice % | -0.05 | -4.22 | 0.001 | | | Rice % | -0.08 | -4.40 | < 0.001 |
| | | Female | -0.12 | -58.55 | < 0.001 | | | Female | 0.11 | 55.51 | < 0.001 | | | Female | -0.21 | -109.01 | < 0.001 |
| | | Priv. Ind. 1995 | 0.02 | 0.77 | 0.451 | | | Priv. Ind. 1995 | 0.03 | 1.24 | 0.226 | | | Priv. Ind. 1995 | 0.01 | 0.45 | 0.659 |
| | Causation | % Urban | -0.004 | -0.23 | 0.820 | | Humans | % Urban | -0.06 | -2.79 | 0.011 | | Achievement | % Urban | 0.01 | 0.35 | 0.732 |
| | | % Herding | 0.01 | 1.58 | 0.127 | | | % Herding | 0.02 | 2.19 | 0.038 | | | % Herding | 0.01 | 1.46 | 0.157 |
| | | % Cult. Land | 0.001 | 0.06 | 0.950 | | | % Cult. Land | -0.01 | -0.80 | 0.433 | | | % Cult. Land | 0.003 | 0.17 | 0.865 |
| Cognition and Discourse | | Rice % | -0.07 | -4.43 | < 0.001 | | | Rice % | -0.08 | -4.29 | < 0.001 | Promotion Orientation and Emotion | | Rice % | -0.07 | -3.79 | < 0.001 |
| | | Female | 0.09 | 45.01 | < 0.001 | | | Female | 0.0002 | 0.12 | 0.906 | | | Female | 0.04 | 15.57 | < 0.001 |
| | | Priv. Ind. 1995 | 0.01 | 0.60 | 0.553 | Self and Groups | | Priv. Ind. 1995 | -0.004 | -0.25 | 0.802 | | | Priv. Ind. 1995 | 0.01 | 0.85 | 0.403 |
| | Certainty | % Urban | -0.04 | -2.43 | 0.025 | | In/Outgroup: | % Urban | -0.04 | -3.07 | 0.006 | | Fashion and | % Urban | 0.003 | 0.24 | 0.813 |
| | | % Herding | 0.01 | 1.07 | 0.296 | | Connecting | % Herding | 0.02 | 2.91 | 0.007 | | Trends | % Herding | 0.004 | 0.67 | 0.507 |
| | | % Cult. Land | -0.02 | -1.49 | 0.151 | | | % Cult. Land | -0.02 | -1.52 | 0.143 | | | % Cult. Land | 0.003 | 0.32 | 0.749 |
| | | Rice % | -0.06 | -4.57 | < 0.001 | | | Rice % | -0.02 | -1.81 | 0.084 | | | Rice % | 0.003 | 0.29 | 0.776 |
| | | Female | 0.08 | 37.82 | < 0.001 | | | Female | -0.01 | -2.93 | 0.003 | | | Female | 0.26 | 137.08 | < 0.001 |
| | | Priv. Ind. 1995 | 0.01 | 0.76 | 0.458 | | | Priv. Ind. 1995 | 0.03 | 1.09 | 0.290 | | | Priv. Ind. 1995 | 0.003 | 0.17 | 0.868 |
| | Possibility/ | % Urban | -0.03 | -1.91 | 0.071 | | In/Outgroup: | % Urban | 0.01 | 0.38 | 0.707 | | Affect | % Urban | -0.06 | -3.65 | 0.001 |
| | Openness | % Herding | 0.004 | 0.54 | 0.597 | | Dividing | % Herding | -0.01 | -0.63 | 0.535 | | | % Herding | 0.01 | 1.61 | 0.120 |
| | | % Cult. Land | -0.01 | -1.08 | 0.293 | | | % Cult. Land | -0.01 | -0.76 | 0.458 | | | % Cult. Land | -0.03 | -1.83 | 0.081 |
| | | Rice % | -0.05 | -3.32 | 0.003 | | | Rice % | -0.002 | -0.10 | 0.923 | | | Rice % | -0.01 | -0.80 | 0.434 |
| | | Female | 0.18 | 93.52 | < 0.001 | | | Female | -0.05 | -17.62 | < 0.001 | | | Female | 0.18 | 90.23 | < 0.001 |
| | | Priv. Ind. 1995 | -0.03 | -1.07 | 0.297 | | | Priv. Ind. 1995 | 0.02 | 1.44 | 0.163 | | | Priv. Ind. 1995 | -0.02 | -0.72 | 0.480 |
| | Assent | % Urban | -0.01 | -0.33 | 0.741 | | Self | % Urban | 0.003 | 0.22 | 0.829 | Cog. & Dis. | Non- | % Urban | -0.01 | -0.32 | 0.751 |
| | | % Herding | -0.01 | -1.03 | 0.312 | | | % Herding | -0.01 | -1.12 | 0.272 | | Fluencies | % Herding | -0.005 | -0.50 | 0.622 |
| | | % Cult. Land | -0.005 | -0.27 | 0.793 | | | % Cult. Land | -0.02 | -1.72 | 0.100 | | | % Cult. Land | -0.005 | -0.26 | 0.794 |
| | | Rice % | 0.09 | 4.66 | < 0.001 | | | Rice % | -0.02 | -1.60 | 0.123 | | | Rice % | 0.07 | 3.66 | 0.001 |

Note: Studies have found a lag between economic growth and cultural change, so we tested this historical indicator of economic modernization. Analyses are hierarchical linear models nested in provinces. This analysis uses the percentage of workers employed in private industry as a marker of modernization. This marker may be particularly important in representing China's shift from state-owned enterprises to private markets.

**Table S17**

*Rice-Wheat Differences Robust to Differences in Ethnic Homogeneity*

| | Word Category | | ß | t | P | | Word Category | | ß | t | P | | Word Category | | ß | t | P |
|---|---|---|---|---|---|---|---|---|---|---|---|---|---|---|---|---|---|
| | | Female | 0.08 | 40.98 | < 0.001 | | | Female | -0.08 | -36.85 | < 0.001 | | | Female | 0.001 | 0.67 | 0.506 |
| | | GDP | -0.005 | -0.11 | 0.917 | | | GDP | -0.02 | -0.60 | 0.553 | | | GDP | -0.02 | -0.44 | 0.663 |
| | Cognitive | % Urban | -0.02 | -0.48 | 0.639 | | | % Urban | 0.03 | 0.76 | 0.456 | | Positivity/ | % Urban | -0.01 | -0.10 | 0.918 |
| | Processes | % Han | -0.01 | -1.25 | 0.223 | | Universalism | % Han | -0.01 | -1.17 | 0.256 | | Optimism | % Han | -0.01 | -1.07 | 0.294 |
| | | % Cult. Land | -0.01 | -0.29 | 0.775 | | | % Cult. Land | -0.02 | -1.04 | 0.309 | | | % Cult. Land | -0.01 | -0.36 | 0.719 |
| | | Rice % | -0.08 | -5.69 | < 0.001 | | | Rice % | -0.06 | -5.53 | < 0.001 | | | Rice % | -0.08 | -5.41 | < 0.001 |
| | | Female | -0.12 | -58.55 | < 0.001 | | | Female | 0.11 | 55.51 | < 0.001 | | | Female | -0.21 | -109.01 | < 0.001 |
| | | GDP | 0.002 | 0.03 | 0.975 | | | GDP | -0.04 | -0.74 | 0.466 | | | GDP | 0.01 | 0.13 | 0.896 |
| | | % Urban | 0.01 | 0.14 | 0.887 | | | % Urban | 0.01 | 0.12 | 0.903 | Promotion Orientation and Emotion | | % Urban | 0.01 | 0.18 | 0.855 |
| | Causation | % Han | -0.01 | -1.14 | 0.266 | | Humans | % Han | -0.02 | -1.97 | 0.061 | | Achievement | % Han | -0.004 | -0.44 | 0.664 |
| | | % Cult. Land | 0.004 | 0.19 | 0.848 | | | % Cult. Land | -0.004 | -0.18 | 0.859 | | | % Cult. Land | -0.003 | -0.15 | 0.882 |
| | | Rice % | -0.08 | -5.56 | < 0.001 | | | Rice % | -0.09 | -5.53 | < 0.001 | | | Rice % | -0.08 | -5.07 | < 0.001 |
| Cognition and Discourse | | Female | 0.09 | 45.01 | < 0.001 | | | Female | 0.0003 | 0.13 | 0.899 | | | Female | 0.04 | 15.57 | < 0.001 |
| | | GDP | -0.02 | -0.48 | 0.636 | Self and Groups | | GDP | 0.01 | 0.31 | 0.759 | | | GDP | -0.005 | -0.16 | 0.877 |
| | | % Urban | -0.01 | -0.19 | 0.849 | | In/Outgroup: | % Urban | -0.05 | -1.15 | 0.264 | | Fashion and | % Urban | 0.02 | 0.46 | 0.647 |
| | Certainty | % Han | -0.005 | -0.65 | 0.524 | | Connecting | % Han | -0.01 | -1.09 | 0.286 | | Trends | % Han | -0.001 | -0.10 | 0.923 |
| | | % Cult. Land | -0.02 | -1.26 | 0.221 | | | % Cult. Land | -0.03 | -1.89 | 0.076 | | | % Cult. Land | 0.003 | 0.26 | 0.799 |
| | | Rice % | -0.06 | -5.93 | < 0.001 | | | Rice % | -0.04 | -3.42 | 0.003 | | | Rice % | 0.003 | 0.30 | 0.766 |
| | | Female | 0.08 | 37.82 | < 0.001 | | | Female | -0.01 | -2.93 | 0.003 | | | Female | 0.26 | 137.09 | < 0.001 |
| | | GDP | 0.01 | 0.15 | 0.883 | | | GDP | -0.10 | -1.77 | 0.091 | | | GDP | -0.06 | -1.36 | 0.189 |
| | Possibility/ | % Urban | -0.03 | -0.64 | 0.529 | | In/Outgroup: | % Urban | 0.12 | 2.06 | 0.054 | | Affect | % Urban | 0.001 | 0.03 | 0.977 |
| | Openness | % Han | -0.01 | -0.79 | 0.435 | | Dividing | % Han | -0.002 | -0.24 | 0.812 | | | % Han | -0.02 | -2.42 | 0.023 |
| | | % Cult. Land | -0.01 | -0.61 | 0.548 | | | % Cult. Land | 0.003 | 0.18 | 0.863 | | | % Cult. Land | -0.02 | -0.97 | 0.345 |
| | | Rice % | -0.04 | -3.78 | 0.001 | | | Rice % | 0.01 | 0.40 | 0.692 | | | Rice % | -0.02 | -1.74 | 0.096 |
| | | Female | 0.18 | 93.52 | < 0.001 | | | Female | -0.05 | -17.61 | < 0.001 | | | Female | 0.18 | 90.23 | < 0.001 |
| | | GDP | -0.001 | -0.02 | 0.986 | | | GDP | 0.01 | 0.23 | 0.818 | Cog. & Dis. | | GDP | -0.02 | -0.41 | 0.686 |
| | | % Urban | -0.02 | -0.39 | 0.701 | | | % Urban | 0.003 | 0.08 | 0.937 | | Non- | % Urban | 0.01 | 0.09 | 0.930 |
| | Assent | % Han | 0.003 | 0.29 | 0.777 | | Self | % Han | 0.01 | 1.17 | 0.253 | | Fluencies | % Han | -0.002 | -0.22 | 0.828 |
| | | % Cult. Land | -0.003 | -0.14 | 0.886 | | | % Cult. Land | -0.02 | -1.45 | 0.163 | | | % Cult. Land | 0.0001 | 0.004 | 0.996 |
| | | Rice % | 0.09 | 5.64 | < 0.001 | | | Rice % | -0.01 | -0.86 | 0.397 | | | Rice % | 0.07 | 4.41 | < 0.001 |

Note: We indexed ethnic homogeneity using the percentage of ethnic Han per province, according to the 2000 Census. Ethnic Han are relevant because (1) different ethnicities may have substantively different cultures, (2) some ethnic groups speak in dialects or even language families that are different from the majority Han, and (3) this percentage reflects ethnic homogeneity. However, percent Han did not meaningfully predict regional differences in language use. Provincial GDP per capita statistics are from 2012 in RMB. The analyses in the main text use 2014 GDP at the request of a reviewer, but 2012 and 2014 GDP correlate $r = 0.99$, so the difference is negligible. Analyses are hierarchical linear models nested in provinces.

**Table S18**
*Rice-Wheat Differences Robust to Pathogen Prevalence*

| | Word Category | | β | t | P | | Word Category | | β | t | P | | Word Category | | β | t | P |
|---|---|---|---|---|---|---|---|---|---|---|---|---|---|---|---|---|---|
| Cognition and Discourse | Cognitive Processes | Female | 0.08 | 40.99 | < 0.001 | Self and Groups | Universalism | Female | -0.08 | -36.85 | < 0.001 | Promotion Orientation and Emotion | Positivity/ Optimism | Female | 0.001 | 0.66 | 0.507 |
| | | Path. Prev. | -0.01 | -0.87 | 0.393 | | | Path. Prev. | -0.01 | -0.79 | 0.437 | | | Path. Prev. | -0.002 | -0.15 | 0.879 |
| | | % Urban | -0.04 | -2.38 | 0.027 | | | % Urban | -0.005 | -0.34 | 0.740 | | | % Urban | -0.04 | -1.78 | 0.089 |
| | | % Herding | 0.01 | 1.10 | 0.280 | | | % Herding | 0.02 | 2.24 | 0.033 | | | % Herding | 0.01 | 0.95 | 0.349 |
| | | % Cult. Land | -0.01 | -0.65 | 0.523 | | | % Cult. Land | -0.01 | -1.01 | 0.325 | | | % Cult. Land | -0.01 | -0.67 | 0.507 |
| | | Rice % | -0.07 | -4.50 | < 0.001 | | | Rice % | -0.05 | -3.84 | < 0.001 | | | Rice % | -0.08 | -4.23 | < 0.001 |
| | Causation | Female | -0.12 | -58.55 | < 0.001 | | Humans | Female | 0.11 | 55.51 | < 0.001 | | Achievement | Female | -0.21 | -109.01 | < 0.001 |
| | | Path. Prev. | -0.01 | -1.40 | 0.175 | | | Path. Prev. | -0.004 | -0.27 | 0.787 | | | Path. Prev. | -0.01 | -0.98 | 0.337 |
| | | % Urban | -0.01 | -0.51 | 0.618 | | | % Urban | -0.05 | -2.23 | 0.036 | | | % Urban | 0.003 | 0.13 | 0.898 |
| | | % Herding | 0.01 | 1.39 | 0.176 | | | % Herding | 0.02 | 2.02 | 0.054 | | | % Herding | 0.01 | 1.31 | 0.201 |
| | | % Cult. Land | 0.001 | 0.09 | 0.930 | | | % Cult. Land | -0.01 | -0.57 | 0.574 | | | % Cult. Land | 0.003 | 0.17 | 0.867 |
| | | Rice % | -0.07 | -4.38 | < 0.001 | | | Rice % | -0.07 | -3.93 | < 0.001 | | | Rice % | -0.07 | -3.79 | < 0.001 |
| | Certainty | Female | 0.09 | 45.01 | < 0.001 | | In/Outgroup: Connecting | Female | 0.0002 | 0.12 | 0.907 | | Fashion and Trends | Female | 0.04 | 15.57 | < 0.001 |
| | | Path. Prev. | -0.005 | -0.53 | 0.602 | | | Path. Prev. | 0.005 | 0.63 | 0.534 | | | Path. Prev. | 0.004 | 0.53 | 0.602 |
| | | % Urban | -0.03 | -2.46 | 0.023 | | | % Urban | -0.04 | -3.10 | 0.006 | | | % Urban | 0.01 | 1.07 | 0.295 |
| | | % Herding | 0.01 | 0.97 | 0.342 | | | % Herding | 0.02 | 3.00 | 0.006 | | | % Herding | 0.005 | 0.75 | 0.458 |
| | | % Cult. Land | -0.02 | -1.44 | 0.164 | | | % Cult. Land | -0.02 | -1.56 | 0.136 | | | % Cult. Land | 0.01 | 0.60 | 0.554 |
| | | Rice % | -0.06 | -4.60 | < 0.001 | | | Rice % | -0.02 | -2.15 | 0.045 | | | Rice % | 0.01 | 0.64 | 0.526 |
| | Possibility/ Openness | Female | 0.08 | 37.82 | < 0.001 | | In/Outgroup: Dividing | Female | -0.01 | -2.93 | 0.003 | | Affect | Female | 0.26 | 137.08 | < 0.001 |
| | | Path. Prev. | -0.01 | -0.61 | 0.545 | | | Path. Prev. | 0.004 | 0.25 | 0.803 | | | Path. Prev. | 0.02 | 1.68 | 0.105 |
| | | % Urban | -0.03 | -1.85 | 0.079 | | | % Urban | 0.03 | 1.23 | 0.235 | | | % Urban | -0.05 | -3.06 | 0.006 |
| | | % Herding | 0.003 | 0.43 | 0.674 | | | % Herding | -0.01 | -0.53 | 0.598 | | | % Herding | 0.02 | 1.91 | 0.067 |
| | | % Cult. Land | -0.01 | -1.00 | 0.331 | | | % Cult. Land | -0.01 | -0.46 | 0.654 | | | % Cult. Land | -0.02 | -1.74 | 0.095 |
| | | Rice % | -0.04 | -3.18 | 0.004 | | | Rice % | 0.01 | 0.40 | 0.691 | | | Rice % | -0.01 | -1.02 | 0.320 |
| | Assent | Female | 0.18 | 93.52 | < 0.001 | | Self | Female | -0.05 | -17.61 | < 0.001 | Cog. & Dis. | Non-Fluencies | Female | 0.18 | 90.23 | < 0.001 |
| | | Path. Prev. | 0.01 | 0.98 | 0.338 | | | Path. Prev. | -0.002 | -0.17 | 0.866 | | | Path. Prev. | -0.003 | -0.29 | 0.774 |
| | | % Urban | -0.01 | -0.49 | 0.629 | | | % Urban | 0.01 | 1.00 | 0.327 | | | % Urban | -0.02 | -0.93 | 0.361 |
| | | % Herding | -0.01 | -0.85 | 0.406 | | | % Herding | -0.01 | -1.07 | 0.291 | | | % Herding | -0.005 | -0.50 | 0.618 |
| | | % Cult. Land | -0.01 | -0.38 | 0.705 | | | % Cult. Land | -0.02 | -1.38 | 0.184 | | | % Cult. Land | -0.01 | -0.45 | 0.660 |
| | | Rice % | 0.08 | 4.41 | < 0.001 | | | Rice % | -0.01 | -1.05 | 0.304 | | | Rice % | 0.06 | 3.61 | 0.002 |

Note: Pathogen prevalence statistics measure rates of human-transmitted infectious diseases based on a study in 1976 (Chen, Campbell, Li, & Peto, 1990) that we supplemented with data from more recent Statistical Yearbooks to increase the sample of provinces. Analyses are hierarchical linear models nested in provinces.

**Table S19**

*Rice-Wheat Border Differences for All Word Categories in Table 3*

| Word Category | Rice Side Words per 10,000 | Wheat Side Words per 10,000 | t | P | 95% CI | |
|---|---|---|---|---|---|---|
| Cognitive Processes | 861.12 | 871.21 | 2.91 | 0.004 | [3.29 | 16.89] |
| Causation | 87.50 | 89.19 | 2.84 | 0.005 | [0.52 | 2.86] |
| Certainty | 120.94 | 122.36 | 2.10 | 0.036 | [0.09 | 2.75] |
| Possibility/Openness | 132.45 | 134.58 | 2.61 | 0.009 | [0.53 | 3.72] |
| Assent | 460.57 | 446.12 | -5.58 | < 0.001 | [-19.53 | -9.38] |
| Non-Fluencies | 54.80 | 53.15 | -2.94 | 0.003 | [-2.76 | -0.55] |
| Universalism | 15.21 | 15.24 | 0.13 | 0.897 | [-0.38 | 0.44] |
| Humans | 96.90 | 101.32 | 5.34 | < 0.001 | [2.80 | 6.04] |
| In/Outgroup: Connecting | 19.45 | 19.89 | 1.78 | 0.074 | [-0.04 | 0.91] |
| In/Outgroup: Dividing | 3.47 | 3.43 | -0.34 | 0.735 | [-0.30 | 0.21] |
| I | 127.90 | 130.77 | 2.51 | 0.012 | [0.63 | 5.12] |
| Self | 6.66 | 6.40 | -2.32 | 0.020 | [-0.47 | -0.04] |
| Positivity/Optimism | 20.48 | 22.28 | 5.67 | < 0.001 | [1.18 | 2.42] |
| Achievement | 112.96 | 117.24 | 4.86 | < 0.001 | [2.56 | 6.01] |
| Fashion and Trends | 10.79 | 10.42 | -1.95 | 0.051 | [-0.73 | 0.00] |
| Affect | 710.86 | 713.00 | 0.74 | 0.462 | [-3.55 | 7.83] |
| We | 23.36 | 23.85 | 1.77 | 0.077 | [-0.05 | 1.02] |

Note: This table presents rice-wheat border tests for all word categories in Table 4. The rice-wheat border analysis in the main text focuses on the word categories that were significant for China as a whole. The border runs through Sichuan, Chongqing, Hubei, Jiangsu, and Anhui. Prefectures in these provinces are defined as rice if they devote more than 50% of cultivated land to paddies. These nearby prefectures differ sharply in rice[17] but very little in temperature, latitude, distance from contact with herding cultures, and other potential confounds.

**Table S20**

*Convergent Validity Tests with Provincial/Prefecture Collectivism Indexes, Norm Tightness, Holistic Thought, and Self-Inflation*

| | Word Category | Markers That Are Higher in Collectivistic Cultures | | | | | | | | Lower Self-Inflation | |
|---|---|---|---|---|---|---|---|---|---|---|---|
| | | Province Collectivism | | Pref. Collectivism Index | | Norm Tightness | | Holistic Thought | | | |
| | | *r* | *p* | *r* | *p* | *r* | *p* | *r* | *p* | *r* | *p* |
| Individualistic | Self | -0.04 | 0.836 | -0.18 | 0.201 | 0.10 | 0.592 | -0.24 | 0.200 | 0.33 | 0.083 |
| | I | 0.12 | 0.512 | -0.09 | 0.557 | -0.67 | < 0.001 | -0.21 | 0.257 | 0.04 | 0.825 |
| | I/(I + We) | 0.28 | .126 | 0.11 | 0.437 | -0.11 | 0.550 | 0.21 | 0.271 | -0.16 | 0.432 |
| | I (Controlling for General Pronoun Use) | 0.38 | 0.039 | 0.11 | 0.456 | -0.35 | 0.061 | 0.14 | 0.461 | -0.10 | 0.636 |
| Collectivistic | We | -0.28 | 0.122 | -0.20 | 0.174 | -0.60 | < 0.001 | -0.45 | 0.013 | 0.24 | 0.228 |
| | Fashion/Trends | -0.23 | 0.204 | 0.14 | 0.321 | 0.31 | 0.088 | -0.21 | 0.266 | 0.14 | 0.476 |

Note: Green shaded rows correlate in the theoretically consistent direction. Red shaded rows correlate in the inconsistent direction. The province and prefectural collectivism indexes are Z scores of (% 3-generation households - % living alone - % nuclear families – divorce to marriage ratio) based on prior indexes in the US and China[60,61]. The tightness of social norms comes from a survey of 11,662 people from 31 provinces. Holistic thought comes from tests using the triad categorization task with 1,019 students from 30 provinces. Self-inflation data comes from 515 college students from 28 provinces who completed the sociogram task. In the sociogram task, participants draw circles to represent the self and their friends. People in individualistic cultures draw the self much larger than friends on average, whereas people in collectivistic cultures show less self-inflation.

**Table S21A**

*Discriminant Validity: Newly Created Word Categories (Top) Correlations with LIWC (Left)*

| LIWC Word Categories | Newly Created Word Categories | | | | | |
| --- | --- | --- | --- | --- | --- | --- |
| | Universalism | In/Outgroup: Connecting | In/Outgroup: Dividing | Self | Positivity/ Optimism | Fashion/ Trends |
| Achieve | 0.16 | 0.04 | 0.07 | 0.21 | 0.52 | 0.05 |
| Adverb | 0.02 | -0.02 | 0.01 | 0.08 | 0.28 | -0.13 |
| Affect | 0.00 | -0.05 | -0.06 | 0.02 | 0.19 | -0.07 |
| Affiliation | 0.09 | 0.01 | -0.02 | 0.01 | 0.13 | 0.08 |
| Anger | 0.08 | 0.05 | 0.03 | 0.07 | 0.08 | -0.11 |
| Anxiety | 0.04 | -0.01 | 0.00 | 0.03 | 0.13 | -0.07 |
| Assent | -0.09 | -0.05 | -0.03 | -0.06 | -0.08 | -0.08 |
| Auxiliary Verbs | 0.09 | -0.02 | 0.01 | 0.13 | 0.41 | -0.10 |
| Biological Processes | -0.06 | -0.06 | -0.05 | -0.04 | 0.09 | -0.01 |
| Body | -0.09 | -0.05 | -0.05 | -0.07 | -0.01 | 0.01 |
| Cause | 0.13 | 0.03 | 0.08 | 0.16 | 0.27 | -0.07 |
| Certainty | 0.10 | 0.01 | 0.02 | 0.14 | 0.35 | -0.05 |
| Cognitive Processes | 0.10 | -0.01 | 0.04 | 0.17 | 0.42 | -0.13 |
| Comparisons | 0.09 | 0.00 | 0.03 | 0.12 | 0.28 | -0.04 |
| Conjunctions | 0.11 | 0.02 | 0.04 | 0.17 | 0.37 | -0.11 |
| Death | 0.03 | 0.07 | 0.01 | -0.01 | -0.07 | -0.10 |
| Differences | 0.08 | 0.01 | 0.03 | 0.15 | 0.36 | -0.14 |
| Discrepancy | 0.08 | -0.02 | 0.00 | 0.13 | 0.42 | -0.10 |
| Drives | 0.16 | 0.06 | 0.06 | 0.14 | 0.35 | 0.02 |
| Family | -0.06 | 0.00 | 0.00 | -0.01 | -0.01 | -0.04 |
| Feel | -0.06 | -0.04 | -0.02 | -0.02 | 0.04 | 0.00 |
| Female | -0.03 | -0.01 | 0.01 | -0.02 | 0.00 | 0.02 |
| Focus Future | 0.02 | 0.01 | 0.02 | 0.02 | 0.07 | -0.05 |
| Focus Past | -0.01 | 0.03 | 0.02 | 0.02 | 0.05 | -0.11 |
| Focus Present | 0.00 | 0.06 | 0.02 | 0.03 | 0.08 | -0.05 |
| Friends | -0.02 | 0.10 | 0.01 | -0.01 | 0.00 | 0.01 |
| Function Words | 0.03 | -0.03 | 0.00 | 0.08 | 0.29 | -0.16 |
| General Particles[1] | 0.09 | 0.00 | 0.02 | 0.11 | 0.25 | -0.11 |
| Health | 0.04 | 0.00 | -0.01 | 0.07 | 0.25 | -0.06 |
| Hear | -0.03 | -0.01 | -0.02 | -0.03 | -0.04 | -0.03 |
| Home | -0.06 | 0.00 | 0.01 | -0.04 | -0.12 | 0.02 |
| Humans | 0.13 | 0.02 | 0.02 | 0.18 | 0.38 | -0.07 |
| I | -0.09 | -0.05 | -0.04 | -0.01 | 0.09 | -0.11 |
| Informal | -0.16 | -0.09 | -0.05 | -0.12 | -0.18 | -0.10 |
| Ingestion | -0.09 | -0.04 | -0.03 | -0.06 | -0.06 | -0.02 |
| Insight | 0.01 | -0.02 | 0.03 | 0.07 | 0.19 | -0.13 |
| Interrogatives | 0.04 | 0.01 | 0.03 | 0.04 | 0.11 | -0.06 |
| Impersonal Pronouns | 0.14 | 0.01 | 0.02 | 0.14 | 0.35 | -0.13 |
| Leisure | 0.02 | -0.01 | -0.03 | 0.00 | 0.01 | 0.13 |
| Male | -0.01 | -0.01 | 0.01 | 0.00 | -0.01 | -0.05 |
| Modal Particles | -0.17 | -0.07 | -0.06 | -0.15 | -0.26 | -0.07 |

Note: Highlighting: $r \geq 0.20$, $r \geq 0.30$, $r \geq 0.40$, $r \geq 0.50$. Researchers often use correlations above 0.90 as clear signs of redundancy, 0.80 as a warning sign, and below 0.80 as acceptable[59]. The in-group/out-group connecting category does not include "we." [1]General particles are grammatical function words, such as the possessive marker 的, which functions like the 's in "John's."

**Table S21B**

*Discriminant Validity: Newly Created Word Categories (Top) Correlations with LIWC (Left)*

| LIWC Word Categories | Newly Created Word Categories | | | | | |
|---|---|---|---|---|---|---|
| | Universalism | In/Outgroup: Connecting | In/Outgroup: Dividing | Self | Positivity/ Optimism | Fashion/ Trends |
| Money | 0.02 | 0.03 | 0.09 | 0.07 | 0.03 | 0.07 |
| Motion | 0.02 | 0.01 | 0.02 | 0.05 | 0.14 | -0.04 |
| Negations | 0.08 | 0.00 | 0.02 | 0.13 | 0.36 | -0.14 |
| Netspeak | -0.11 | -0.04 | -0.04 | -0.10 | -0.21 | 0.08 |
| Non-Fluencies | -0.11 | -0.05 | -0.04 | -0.07 | -0.10 | -0.09 |
| Numbers | -0.05 | -0.02 | -0.01 | -0.03 | -0.04 | -0.08 |
| Particles | -0.06 | -0.06 | -0.03 | -0.04 | -0.03 | -0.11 |
| Perceptual Processes | -0.07 | -0.05 | -0.05 | -0.06 | -0.05 | 0.04 |
| Positive Emotion | -0.06 | -0.08 | -0.08 | -0.06 | 0.03 | 0.02 |
| Power | 0.11 | 0.09 | 0.09 | 0.14 | 0.12 | -0.03 |
| Personal Pronouns | -0.01 | -0.04 | -0.04 | 0.01 | 0.23 | -0.09 |
| Prepositions | 0.14 | 0.07 | 0.08 | 0.18 | 0.27 | -0.10 |
| Postposition[1] | 0.01 | 0.02 | 0.03 | 0.02 | 0.01 | -0.04 |
| Progressive Markers[2] | -0.10 | -0.03 | 0.00 | -0.07 | -0.13 | -0.12 |
| Pronouns | 0.05 | -0.03 | -0.02 | 0.06 | 0.30 | -0.11 |
| Quantifiers | 0.05 | -0.02 | 0.01 | 0.07 | 0.18 | -0.03 |
| Measure Words[3] | -0.04 | 0.02 | 0.02 | 0.00 | -0.06 | -0.02 |
| Relativity | 0.08 | 0.06 | 0.06 | 0.08 | 0.18 | -0.08 |
| Religion | 0.12 | 0.03 | -0.01 | 0.05 | 0.09 | -0.05 |
| Reward | -0.03 | -0.04 | -0.04 | 0.00 | 0.17 | 0.01 |
| Risk | 0.12 | 0.06 | 0.06 | 0.16 | 0.31 | -0.11 |
| Sad | 0.07 | 0.01 | -0.02 | 0.07 | 0.29 | -0.10 |
| See | 0.01 | -0.01 | -0.02 | -0.03 | -0.07 | 0.10 |
| Sexual | -0.02 | -0.05 | -0.05 | -0.04 | 0.05 | 0.06 |
| She/He | 0.02 | 0.00 | 0.01 | 0.03 | 0.05 | -0.05 |
| Social | 0.08 | 0.01 | 0.02 | 0.10 | 0.29 | -0.06 |
| Space | 0.18 | 0.06 | 0.09 | 0.08 | 0.10 | -0.05 |
| Specific Article[4] | 0.05 | 0.02 | 0.03 | 0.02 | 0.02 | -0.04 |
| Swear Words | -0.05 | -0.01 | 0.01 | -0.04 | -0.10 | -0.07 |
| Tense Marker[5] | -0.06 | 0.00 | 0.02 | -0.03 | -0.01 | -0.14 |
| Possibility/Openness (Tentat)[6] | 0.05 | -0.01 | 0.02 | 0.10 | 0.26 | -0.06 |
| They | 0.07 | 0.06 | 0.03 | 0.07 | 0.06 | -0.04 |
| Time | -0.02 | 0.04 | 0.02 | 0.03 | 0.16 | -0.07 |
| Work | 0.12 | 0.10 | 0.12 | 0.15 | 0.18 | 0.00 |
| We | 0.07 | 0.04 | -0.01 | 0.05 | 0.16 | -0.01 |
| You | 0.05 | -0.04 | -0.04 | 0.00 | 0.27 | -0.03 |
| You Plural[7] | -0.06 | 0.00 | 0.00 | -0.06 | -0.08 | -0.02 |

Note: Highlighting: $r \geq 0.20$, $r \geq 0.30$, $r \geq 0.40$, $r \geq 0.50$, $r \leq -0.20$. The in-group/out-group connecting category does not include "we." [1]Postposition: Grammatical word that comes after the word it modifies, such as 的 to mark possession. [2]Progressive markers communicate progressive tense, such as 了. [3]Measure words count objects, such as "bowl" in "one bowl of soup" (一碗汤). [4]Specific articles denote a specific object/actor, such as 各 (each) and 某 ([a] certain [person/thing]). [5]Tense markers mark time, such as "recently" (近日). [5]We title the LIWC category of "tentative" as "possibility/openness." [7]You plural: Second person plural, such as 你们 (similar to "y'all" in English).

# Supplemental Materials

## 1. Classifying Dialects

To categorize provinces into dialect groups, we relied on the *Language Atlas of China* ([map](); 中国语言地图集). For the Mandarin-only analysis, we classified provinces as Mandarin or non-Mandarin. For the "excluding Cantonese" analysis, we classified provinces as Cantonese or non-Cantonese.

As is probably the case for most large countries around the world, there are debates about what should count as distinct groups. Three provinces stick out as particularly open for debate:

- Shanxi is sometimes categorized as having its own Jin dialect and sometimes as Mandarin, as in the *Language Atlas of China*. We followed the classification of Shanxi as Mandarin.
- A geographically small part of southern Jiangsu is categorized as Wu dialect. However, because most of the province is Mandarin, we classified Jiangsu as Mandarin.
- The province of Anhui is more split than Jiangsu, with areas of Mandarin, Wu, Hui, and Gan. We classified Anhui as non-Mandarin to reflect this diversity.

## 2. Pathogen Prevalence

We indexed pathogen prevalence based on statistics from the earliest data we could find. This study surveyed disease in 49 counties in 1976[87]. That study did not have data from a handful of provinces. To compensate for this limited data, we added statistics from the earliest statistical yearbooks we could find (from 2001 to 2008 depending on the province).

A total of 10 provinces had data from both sources. This allowed us to test how well the two sources agreed, especially considering that the statistics are separated by about 30 years. The two sources were meaningfully correlated, although it was not significant in such a small sample $r(8) = 0.43$, $P = 0.25$. With these two data sources, we had pathogen data for 15 provinces in our sample. More details on these measures are in the supplemental materials of the earlier rice-wheat study in China[24].

## 3. Herding Cultures

Although we compare rice and wheat farming, China is also home to cultures that traditionally herded, such as Mongolians, Manchu, and Tibetans. It is important to test the effect of herding because there is evidence that herding cultures are more independent and analytic-thinking than nearby farming communities[30,88]. In addition, herding cultures are more often located near China's wheat-farming areas, which means they are more likely to have been influenced by individualistic herding cultures. We tested this possibility by indexing the percentage of the provincial population from traditional herding cultures in the 2000 Census.

China's biggest historic herding groups were Tibetans, Uyghurs, Mongolians, Manchus, and Turkic groups like the Kyrgyz. We used the following list: Uyghur, Mongolians, Manchu, Tibetan, Kyrgyz, Salar, Daur, Xibo, Tajik, Uzbek, Ewenke, Yugur, Tatar, Elunchun, Hezhe, Menba, Luoba, and Kazakh. For many of these smaller cultural groups, it is difficult to accurately estimate the percentage that herding made up of their traditional subsistence. However, decisions to include or exclude many of the small groups makes little difference to the analyses, because the groups with sparse historical records have such small populations.

### 4. City Tier System

Although it is common in China to refer to prefectures as first tier, second tier, and third tier, there are different lists of exactly which prefectures fall into which categories. To avoid the bias inherent in choosing which cities fall into which categories, we relied on a publicly available list from Baidu Baike (similar to Wikipedia). The full list is available in the supplemental materials.

### 5. Rice Suitability

We measured rice suitability using the United Nations Food and Agriculture Organization's Global Agro-Ecological Zones database. We used values for high-input rainfed current cultivated wetland rice. These scores estimate the suitability for plots of land regardless of whether people are farming rice there or not. The models use climate data from 1961 to 1990.

It is worth keeping in mind that these models are not designed exactly for our purpose—estimating historical rice suitability. Were we to design these models, we would estimate historical climate data farther back in history. We would also calculate suitability using historical breeds of rice and pre-modern tools. Thus, we believe the United Nations' estimates are useful but likely overestimate the suitability of rice historically.

It's also important to remember that suitability changes over time—not just as the climate changes, but as new rice strains develop, and as humans create new methods of farming. Suitability is not completely fixed. However, we believe these estimates are useful in providing broad outlines of historical suitability.

The database provides summary scores for each province, so we used these values directly in the province analysis. However, the database does not aggregate the values for prefectures. Thus, we used two methods to estimate suitability at the prefecture level. First, for any province that the United Nations database aggregated suitability to a score of zero, we simply assigned zero to all prefectures in the province.

Second, for provinces with aggregate scores above zero, we estimated suitability using prefecture-level temperature and rain. Temperature and rainfall both strongly predict actual rice farming, but when put in the same model, rainfall ($\beta = 0.74$, $P < 0.001$) is a much stronger predictor than temperature ($\beta = 0.11$, $P = 0.030$). We compared simple models to models that used squared predictors, gated values, and interactions between temperature and rainfall. However, these models produced highly similar results, so we chose the basic models for the sake of simplicity and to avoid over-fitting.

### 6. Newly Created Theory-Driven Categories

The LIWC dictionary was not built with the goal of measuring cultural differences in mind. Thus, there were categories that we were interested in but not reflected in LIWC. We generated four new categories based on prior theorizing. In this section, we lay out the theoretical basis for each category.

**In-group/out-group.** Some cultural psychologists have argued that collectivistic cultures do not so much prioritize the needs of groups, but rather draw a distinction between people in familiar groups (in-group) versus unfamiliar groups (out-groups)[89]. In China, Shanghai and other southern rice-farming regions have a reputation as being *paiwai* (排外), excluding to outsiders. This fits with the finding that people from rice-growing provinces in China drew a larger distinction between friends and strangers than people from wheat-growing provinces[24,55].

We created two sub-categories of in-group/out-group words to take into account that some words carry a positive connotation, and some carry a negative connotation. For example, people usually use the word *comrade/brethren* (同胞) to emphasize connection and unity among people of different groups. In contrast, people usually use the term *outsider/out-of-towner* (外地人) to criticize or look down on people outside of the local group.

**Universalism.** Similar to in-group/out-group words, we created a category of universalism words. These words represent a focus on humans without regard to their group membership. This reflects the opposite of the in-group/out-group distinction that researchers have theorized is a part of collectivistic cultures[89].

**Fashion/trends.** We speculate that fashion and trends may be more important in collectivistic cultures. This would make sense with the idea that people in collectivistic cultures have more socially shared standards, whereas people in individualistic cultures are more likely to have individually defined standards[54]. If standards for beauty are more socially consensual in collectivistic cultures, it would be more important to stay on top of social trends. Based on this reasoning, we created a fashion/trends category including words like *hot* (to describe ideas and trends, 热门), *out-of-style* (过时), and *celebrities* (名流).

**Alone.** Previous studies have found that people in individualistic regions are more likely to spend time alone than people in collectivistic regions. For example, Americans in individualistic states are more likely to live alone or drive alone[61]. In China, people were more likely to be sitting alone in cafes in more individualistic wheat-farming areas than in collectivistic rice-farming areas[25]. These words are a limited set of words that have a connotation of spending time alone, such as *individually* (单独), *be alone* (独处). It does not include words about *feeling* alone, like "lonely."

**Justice.** We created a word category related to justice, such as *fair* (公平), *equal* (平等), *public welfare* (公益). The theory behind this category is the research on cultural differences in moral visions. Research using moral dilemmas across cultures has found that Americans are more likely to focus on abstract justice, whereas participants in India were more likely to focus on relational duties[90].

### 7. Wenxin Dictionary

We also ran analyses with the Wenxin dictionary[91], which is an expanded form of the LIWC Chinese dictionary. Because the results were largely similar to the results for LIWC, we report results from LIWC, which is more broadly used in research.

### 8. Alternative Measures of Modernization

Some researchers have argued that GDP is not the best measure of modernization[92]. For example, some countries have become wealthy through mining or oil but have not really modernized their economies. Modernization theorist Ronald Inglehart has advocated for using the percentage of people employed in the service sector as a measure of modernization and the turn away from farming and heavy industry[93]. In China, the move from the government-led economy to the private sector may also reflect modernization. Thus, we tested recent (year 2010) and historical (1996) percentage of employed people working in the service sector and the percentage of people employed in private industry.

### 9. Urbanization

We used the percentage of urban residents (城镇人口比例) in each province as an index of urbanization from the year 2016 to represent modern statistics and the year 2000 to represent time-lagged differences in recent history[94].

### 10. Pathogen Prevalence

As a measure of historical pathogens, we used human-transmitted disease rates from a 1973-1975 study[87]. Because this study was missing several provinces, we supplemented this data with disease rates from provincial statistical yearbooks from 2001 to 2008 for 14 provinces (we report more details on this method in an earlier study[24]).

### 11. Temperature

Some researchers have argued that temperature might drive differences between cultures (although the exact reason why temperature would cause differences is often unclear[65]). We collected data on the average temperature for the capital city of each province. In China, average temperature is highly correlated with rice $r(31) = 0.73$, $P < 0.001$.

### 12. Climatic Demands

One team of researchers argued that climatic demands can explain regional differences in China[29]. This theory is that humans bind together and rely on each other in response to difficult climates. We followed prior research and operationalized climatic demands as the sum deviation from 22 degrees Celsius for the average highs and lows in the coldest (January) and hottest month (July). In the supplemental materials, we test an alternative version of climatic demands theory that takes into account wealth.

### 13. Herding Cultures

Rice and wheat are just two types of subsistence styles. Studies comparing people in nearby farming and herding cultures have found evidence that herders tend to be more independent than farmers[30]. Areas of north and west China are home to cultures that traditionally herded, such as the Mongolians and Tibetans. As a measure of herding cultures, we calculated the percentage of the population in each province that belonged to 18 historically herding cultures from China's 2000 national census. Because this data was not normally distributed, we square root transformed the data.

### 14. Farming in General versus Rice Farming

Previous studies on subsistence theory compared farming cultures to herding, hunting, and fishing cultures, finding that farming cultures tend to be more interdependent[30,95]. Thus, we need to test whether what we are calling an effect of rice is not actually an effect of farming in general. As a measure of farming in general, we collected data on the percentage of cultivated land in each province. This is actually uncorrelated with rice farming $r(31) = 0.04$, $P = 0.823$.

### 15. Cognitive Process Words

Are the cognitive process words related to analytic cultural thought style? As a first pass, we calculated average cognitive process word use scores for each province. Then we compared these province scores to our earlier study measuring analytic thought among college students across China[24]. Provinces that used more cognitive process words scored marginally lower on holistic thought (or higher on analytic thought), $r(27) = -0.33$, $P = 0.083$. Aggregating scores to the province level limits the sample to 29 provinces, and the correlation was only marginally significant. However, the direction is consistent with the idea that regional differences in the use of cognitive process words reflect, to some extent, differences in holistic versus analytic thought.

### 16. Rice Statistics in Japan

One limitation of the rice statistics from Japan is that they do not separate dryland versus paddy rice. Theoretically, paddy rice should make cultures more interdependent than dryland rice because dryland rice grows without irrigation systems[17]. However, this may not present much of a problem in Japan because national statistics (which do separate paddy rice and dryland rice) show that dryland rice accounts for just 1.6% of rice land. Thus, the rice statistics overall overwhelmingly represent paddy rice.

In the main text, we ask the question of whether the data from 1975 represents historical patterns of rice farming. One way to test this is to use data from 10 larger regional blocks going back to 1950. Although there are fewer regional blocks than prefectures, the data showed very little variation from 1950 to 1975. Data from 1950 correlated highly with data from 1975, $r(8) = 0.90$, $P = 0.001$. Even nearly 60 years later (using data from 2009), the correlation remained high, $r(8) = 0.89$, $P = 0.001$. In sum, this data suggests that regional differences in rice percentage are highly stable over time. Data spanning roughly 100 years in China shows similar results[24].

To calculate the correlation between environmental suitability for rice and actual rice farming, we used data from the United Nations Food and Agriculture Organizations Global Agro-Ecological Zones database. This database estimates environmental suitability for different crops based on environmental variables like rain, sun, slope, and soil. We used data for high-input wetland rice. This calculates suitability scores for plots of land, regardless of whether people are actually farming rice there or not.

Examining the data, it was clear that large cities were farming less rice than predicted by their environmental suitability scores. This suggests that buildings are taking up land that would probably otherwise be growing rice (and probably grew rice historically). Thus, we ran a regression using environmental suitability scores and statistics on the percentage of land occupied by buildings per prefecture. Taking into account buildings, environmental suitability strongly predicted rice farming $β = 0.78$, $P < 0.001$.

## 17. Japan Regional Statistics

To measure economic development, we collected 2012 prefecture GDP per capita from the Cabinet Office of Japan's *2013 Annual Report on Prefectural-Level Accounts* (*Heisei 25 nen Kenminn keizai keisan*). To measure educational attainment, we collected 2010 data on the percentage of residents who had completed a college degree from a report of the Statistics Bureau of Japan (Table E).

## 18. R Squared for Rice vs. Modernization in Japan

With the China data, we compared the percentage of variance explained by rice, GDP, and urbanization. This comparison gets a bit trickier in Japan. In China, there is wide variation in rice farming. Many prefectures farm zero rice. Japan has a narrower range of variation in rice. In Japan, every prefecture farms at least some rice. Statistically, Japanese prefectures have less than half (39%) the variance in rice compared to Chinese prefectures. Thus, holding all else equal, we would expect rice to explain less variance in Japan because there is less variance in rice.

An analogy would be in testing for the effect of height on people's ability to dunk a basketball. In one study, we test 100 American 20-year-olds and find a large effect of height. Then we test 100 NBA players and find a smaller effect of height, since height varies less between NBA players than for the population at large. Thus, our expectations for the effect size need to take into account how much variation the sample has in the key variable.

For rice, the variation in China is a better representation of the world at large than Japan is. The world has many areas that farm no rice and many areas that farm a high percentage of rice. In contrast, Japan is a less representative sample of rice farming percentages, similar to how NBA players are less representative of the height of humans than regular adults .

To make matters more complicated, Japanese prefectures have more variation in GDP. We compared variance by converting GDP per capita to US dollars. Based on this, Japanese prefectures had roughly double (219%) the variance of China.

Using these differences in variance from China, we can calculate expected versus actual $R$ squared for Japan:

**Rice**
    Expected $R^2$:    0.12%
    Actual $R^2$:    0.17%

**GDP**
    Expected $R^2$:    0.26%
    Actual $R^2$:    0.20%

Thus, rice has lower variance in Japan and therefore is expected to account for less variance in word use. However, rice accounts for more word use than would be expected from the fact that there is less variation in rice in Japan. In contrast, GDP explains less variance than we would expect, given that GDP varies more in Japan than China.

## 19. Convergent Validity Tests for Word Categories

To test convergent validity, we calculated the mean word usage for each province and prefecture. We then calculated region-level correlations between word usage and other markers of collectivism and individualism. The notes below the tables list the data sources for the markers of individualism and collectivism. Table 2 shows the correlations for the word categories were different between rice and wheat regions. Table S20 shows the correlations for the word categories that were not different between rice and wheat regions.

In general, the word categories that did not show rice-wheat differences tended to fail the tests of convergent validity. For example, collectivistic areas tended to use *less* "we" than individualistic areas. And although collectivistic areas also used less "I," they used far less "we." Similarly, collectivistic areas actually used fewer words related to fashion and trends.

To calculate the prefecture-level correlations, we took into account the sample size for different prefectures. The main analyses throughout this paper use multi-level analyses, which take into account the different sample sizes across prefectures. But these basic validity correlations treat each prefecture as an equal data point. Therefore, we limited the convergent validity analyses to prefectures with over 200 users. That left 50 prefectures.